\documentclass{article}
\newcommand{\transfusiontitlelineI}{Transfusion: Understanding Transfer Learning}
\newcommand{\transfusiontitlelineII}{for Medical Imaging}
\newcommand{\transfusiontitle}{\transfusiontitlelineI{} \transfusiontitlelineII{}}

\ifdefined\formatforarxiv
\usepackage{amsmath}
\usepackage[table,svgnames,dvipsnames]{xcolor}
\usepackage{hyperref}
\usepackage{cleveref}

\newcommand\myshade{70}
\colorlet{mylinkcolor}{NavyBlue}
\colorlet{mycitecolor}{Aquamarine}
\colorlet{myurlcolor}{Aquamarine}

\hypersetup{
  linkcolor  = mylinkcolor!\myshade!black,
  citecolor  = mycitecolor!\myshade!black,
  urlcolor   = myurlcolor!\myshade!black,
  colorlinks = true,
}

\usepackage[margin=2.8cm]{geometry}
\usepackage{libertine}
\usepackage{libertinust1math}
\usepackage{inconsolata}
\usepackage[T1]{fontenc}    
\usepackage[sf,bf]{titlesec}
\usepackage{verbatimbox}

\usepackage{parskip}

\usepackage[]{natbib}
\title{\textsf{\textbf{\transfusiontitlelineI \\ \transfusiontitlelineII}}}

\usepackage[auth-lg,affil-sl]{authblk}

\newcommand*\samethanks[1][\value{footnote}]{\footnotemark[#1]}
\author[1,2]{Maithra Raghu\thanks{Equal Contribution}}
\author[2]{Chiyuan Zhang\samethanks[1]}
\author[1]{Jon Kleinberg\thanks{Equal Contribution}}
\author[2]{Samy Bengio\samethanks[2]}

\affil[1]{Cornell University}
\affil[2]{Google Brain}

\date{}

\else
     \PassOptionsToPackage{numbers, compress}{natbib}

\usepackage[final]{neurips_2019}

\usepackage[utf8]{inputenc} \usepackage[T1]{fontenc}    \usepackage{hyperref}       
\usepackage{libertine}
\usepackage{libertinust1math}
\usepackage{inconsolata}
\usepackage[T1]{fontenc}    
\title{\transfusiontitle}

\newcommand*\samethanks[1][\value{footnote}]{\footnotemark[#1]}
\author{  Maithra Raghu\thanks{Equal Contribution.} \\
  Cornell University and Google Brain\\
  \texttt{maithrar@gmail.com} \\
   \And
  Chiyuan Zhang\samethanks[1] \\
  Google Brain\\
  \texttt{chiyuan@google.com} \\
  \AND
  Jon Kleinberg\thanks{Equal Contribution.} \\
  Cornell University\\
  \texttt{kleinber@cs.cornell.edu} \\
  \And
  Samy Bengio\samethanks[2] \\
  Google Brain\\
  \texttt{bengio@google.com} \\
                              }
\fi
\usepackage{url}            \usepackage{booktabs}       \usepackage{amsfonts}       \usepackage{nicefrac}       \usepackage{microtype}      
\usepackage{graphicx}
\usepackage{url}
\usepackage{bm}
\usepackage[export]{adjustbox}
\usepackage{amsmath}
\usepackage{amsbsy}
\usepackage{calc}
\usepackage{amsthm}
\usepackage{amssymb}
\usepackage{floatrow}
\usepackage[percent]{overpic}

\usepackage{grffile}
\usepackage{subcaption}
\usepackage{booktabs} \usepackage{multirow}

\usepackage{sidecap}
\usepackage[export]{adjustbox}

\usepackage[table,svgnames,dvipsnames]{xcolor}

\usepackage{xspace}
\newcommand{\imagenet}{\textsc{ImageNet}\xspace}
\newcommand{\retina}{\textsc{Retina}\xspace}
\newcommand{\chexpert}{\textsc{CheXpert}\xspace}

\begin{document}

\maketitle

\begin{abstract}
Transfer learning from natural image datasets, particularly \imagenet, using standard large models and corresponding pretrained weights has become a de-facto method for deep learning applications to medical imaging.
However, there are fundamental differences in data sizes, features and task specifications between natural image classification and the target medical tasks, and there is little understanding of the effects of transfer. In this paper, we explore properties of transfer learning for medical imaging. A performance evaluation on two large scale medical imaging tasks shows that surprisingly, transfer offers little benefit to performance, and simple, lightweight models can perform comparably to \imagenet architectures. Investigating the learned representations and features, we find that some of the differences from transfer learning are due to the over-parametrization of standard models rather than sophisticated feature reuse. We isolate where useful feature reuse occurs, and outline the implications for more efficient model exploration. We also explore feature independent benefits of transfer arising from weight scalings.
\end{abstract}

\section{Introduction}
With the growth of deep learning, transfer learning has become integral to many applications --- especially in medical imaging, where the present standard is to take an existing architecture designed for natural image datasets such as \imagenet, together with corresponding pretrained weights (e.g. ResNet \citep{he2016deep}, Inception \citep{szegedy2015going}), and then fine-tune the model on the medical imaging data.

This basic formula has seen almost universal adoption across many different medical specialties. Two prominent lines of research have used this methodology for applications in radiology, training architectures like ResNet, DenseNet on chest x-rays \citep{wang2017chestx, rajpurkar2017chexnet} and ophthalmology, training Inception-v3, ResNet on retinal fundus images \citep{abramoff2016improved, Gulshan2016Retinal, raghu2018dup, de2018clinically}. The research on ophthalmology has also culminated in FDA approval \citep{topol2019high}, and full clinical deployment \citep{van2018validation}. Other applications include performing early detection of Alzheimer's Disease \citep{ding2018deep}, identifying skin cancer from dermatologist level photographs \citep{esteva2017dermatologist}, and even determining human embryo quality for IVF procedures \citep{khosravi2018robust}.

Despite the immense popularity of transfer learning in medical imaging, there has been little work studying its precise effects, even as recent work on transfer learning in the \textit{natural image} setting \citep{he2018rethinking, kornblith2018better, ngiam2018domain, huh2016makes, geirhos2018imagenet} has challenged many commonly held beliefs. For example in \citep{he2018rethinking}, it is shown that transfer (even between similar tasks) does not necessarily result in performance improvements, while \citep{kornblith2018better} illustrates that pretrained features may be less general than previously thought.

In the medical imaging setting, many such open questions remain. As described above, transfer learning is typically performed by taking a standard \imagenet architecture along with its pretrained weights, and then fine-tuning on the target task. However, \imagenet classification and medical image diagnosis have considerable differences.

First, many medical imaging tasks start with a large image of a bodily region of interest and use variations in local textures to identify pathologies. For example, in retinal fundus images, small red `dots' are an indication of microaneurysms and diabetic retinopathy \citep{ICDRStandards}, and in chest x-rays local white opaque patches are signs of consolidation and pneumonia. This is in contrast to natural image datasets like \imagenet, where there is often a clear global subject of the image (Fig.~\ref{fig:example-images}). There is thus an open question of how much \imagenet feature reuse is helpful for medical images.

Additionally, most datasets have larger images (to facilitate the search for local variations), but with many fewer images than \imagenet, which has roughly one million images. By contrast medical datasets range from several thousand images \citep{khosravi2018robust} to a couple hundred thousand \citep{Gulshan2016Retinal, rajpurkar2017chexnet}.

Finally, medical tasks often have significantly fewer classes ($5$ classes for Diabetic Retinopathy diagnosis \citep{Gulshan2016Retinal}, $5 - 14$ chest pathologies from x-rays \citep{rajpurkar2017chexnet}) than the standard \imagenet classification setup of $1000$ classes. As standard \imagenet architectures have a large number of parameters concentrated at the higher layers for precisely this reason, the design of these models is likely to be suboptimal for the medical setting.

In this paper, we perform a fine-grained study on transfer learning for medical images. Our main contributions are:

\textbf{[1]} We evaluate the performance of standard  architectures for natural images such as \imagenet, as well as a family of non-standard but smaller and simpler models, on two large scale medical imaging tasks, for which transfer learning is currently the norm. We find that (i) in all of these cases, transfer does not significantly help performance (ii) smaller, simpler convolutional architectures perform comparably to standard \imagenet models (iii) \imagenet performance is not predictive of medical performance. These conclusions also hold in the very small data regime.

\textbf{[2]} Given the comparable performance, we investigate whether using pretrained weights leads to different learned representations, by using (SV)CCA \citep{raghu2017svcca} to directly analyze the hidden representations. We find that pretraining does affect the hidden representations, but there is a confounding issue of model size, where the large, standard \imagenet models do not change significantly through the fine-tuning process, as evidenced through surprising correlations between representational similarity at initialization and after convergence.

\textbf{[3]} Using further analysis and weight transfusion experiments, where we partially reuse pretrained weights, we isolate locations where meaningful feature reuse does occur, and explore hybrid approaches to transfer learning where a subset of pretrained weights are used, and other parts of the network are redesigned and made more lightweight.

\textbf{[4]} We show there are also \textit{feature-independent} benefits to pretraining --- reusing only the \textit{scaling} of the pretrained weights but not the features can itself lead to large gains in convergence speed.

\section{Datasets}
\label{sec-data-setup}
\begin{figure*}
    \centering
    \includegraphics[width=0.115\linewidth]{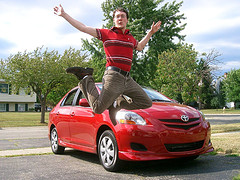}
    \includegraphics[width=.11\linewidth]{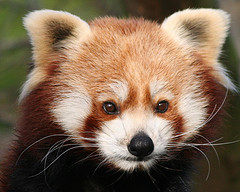}
    \includegraphics[width=.132\linewidth]{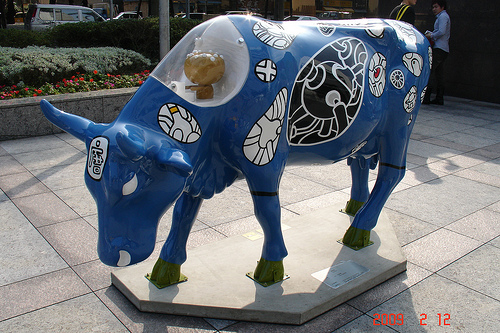}\hfill
    \includegraphics[width=.091\linewidth]{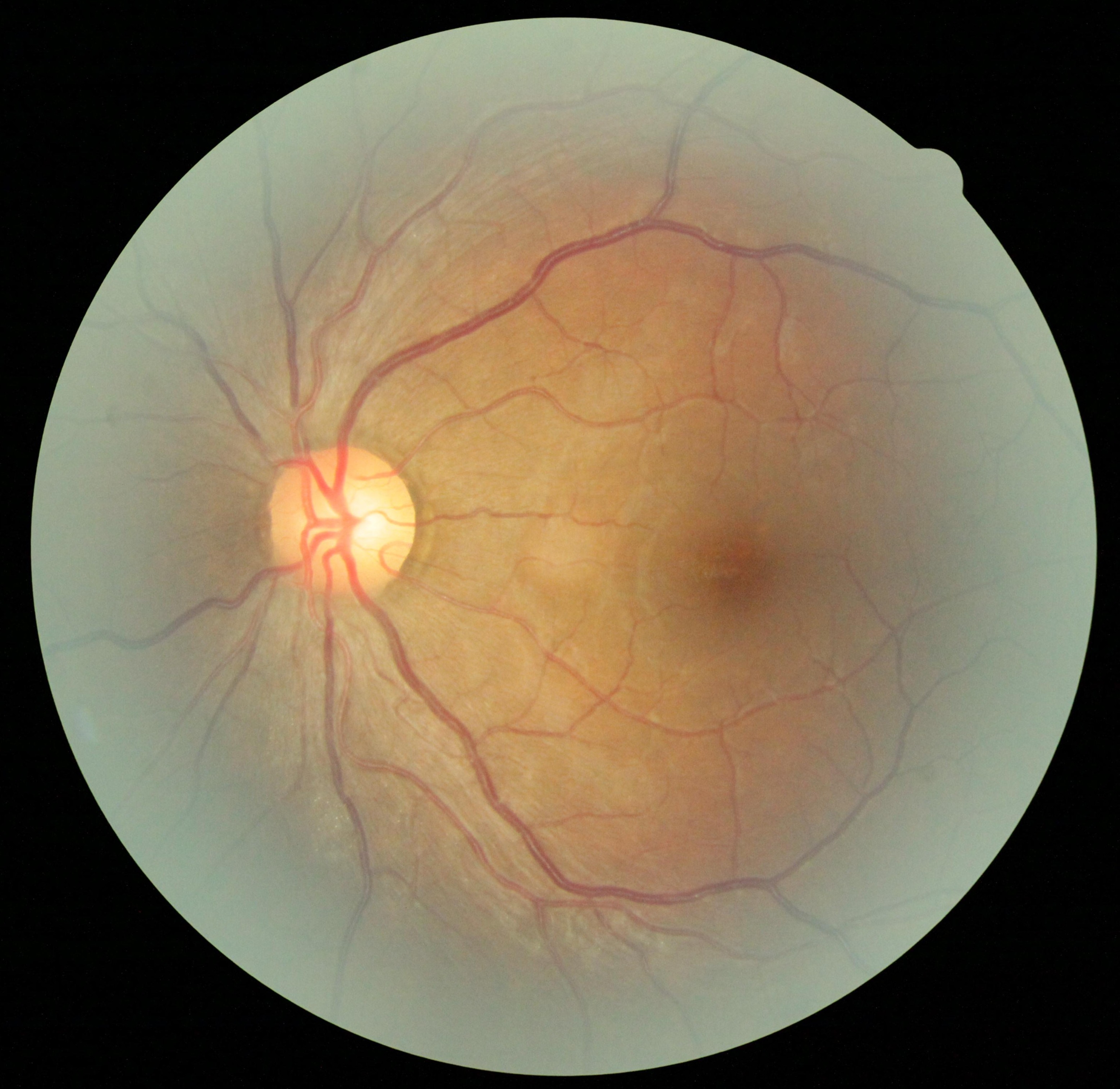}
    \includegraphics[width=.089\linewidth]{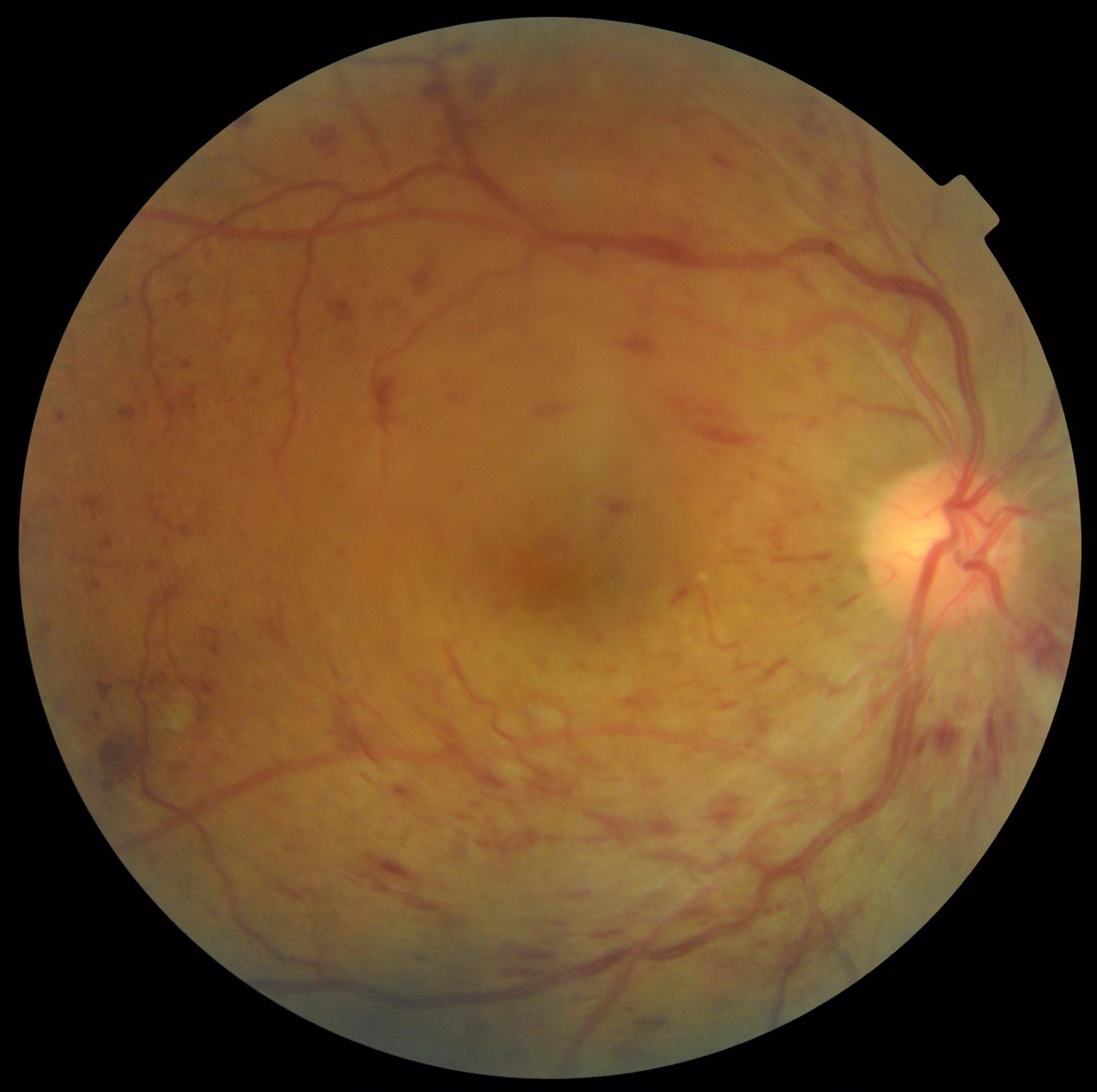}
    \includegraphics[width=.12\linewidth]{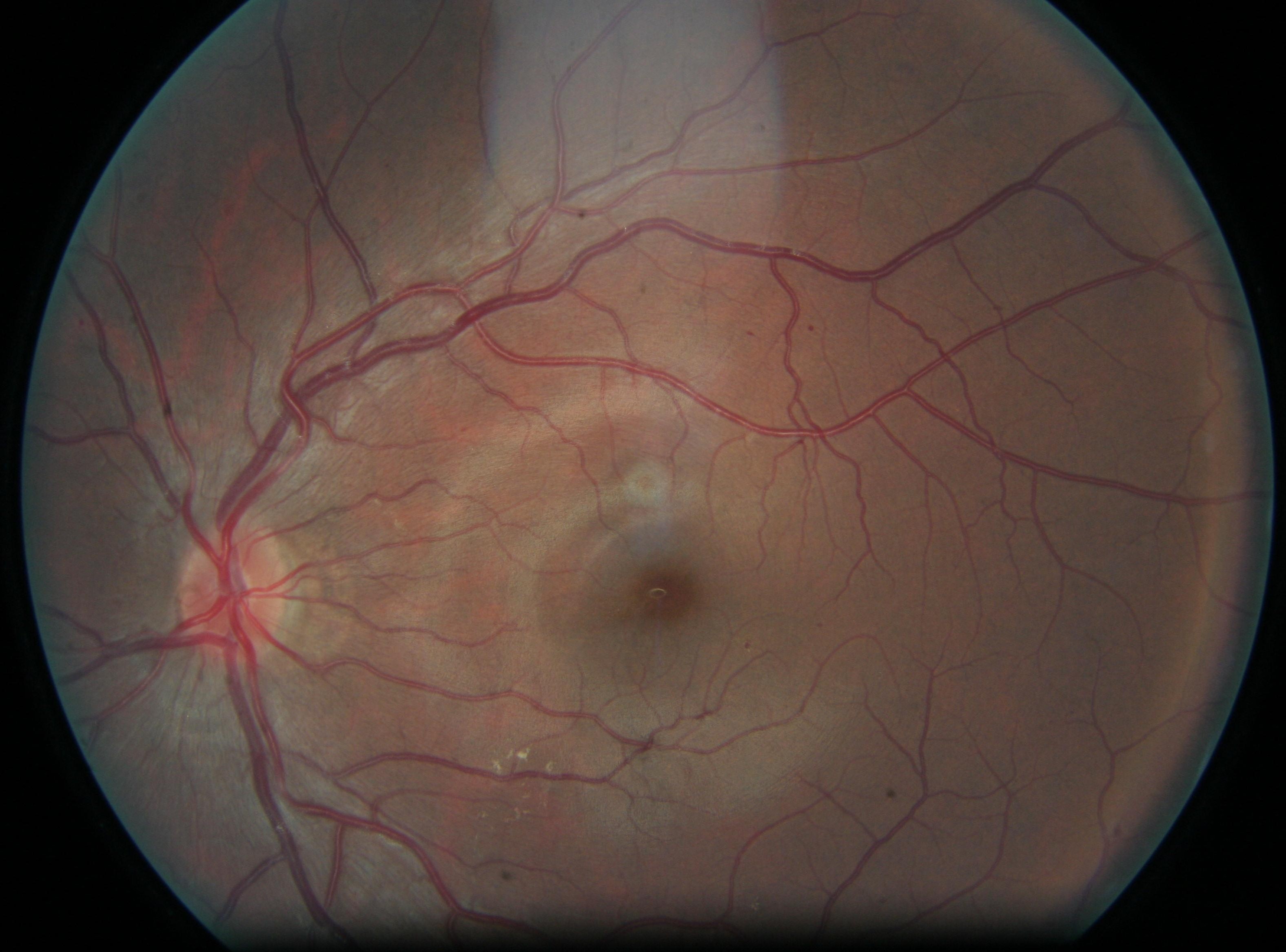}\hfill
    \includegraphics[width=.088\linewidth]{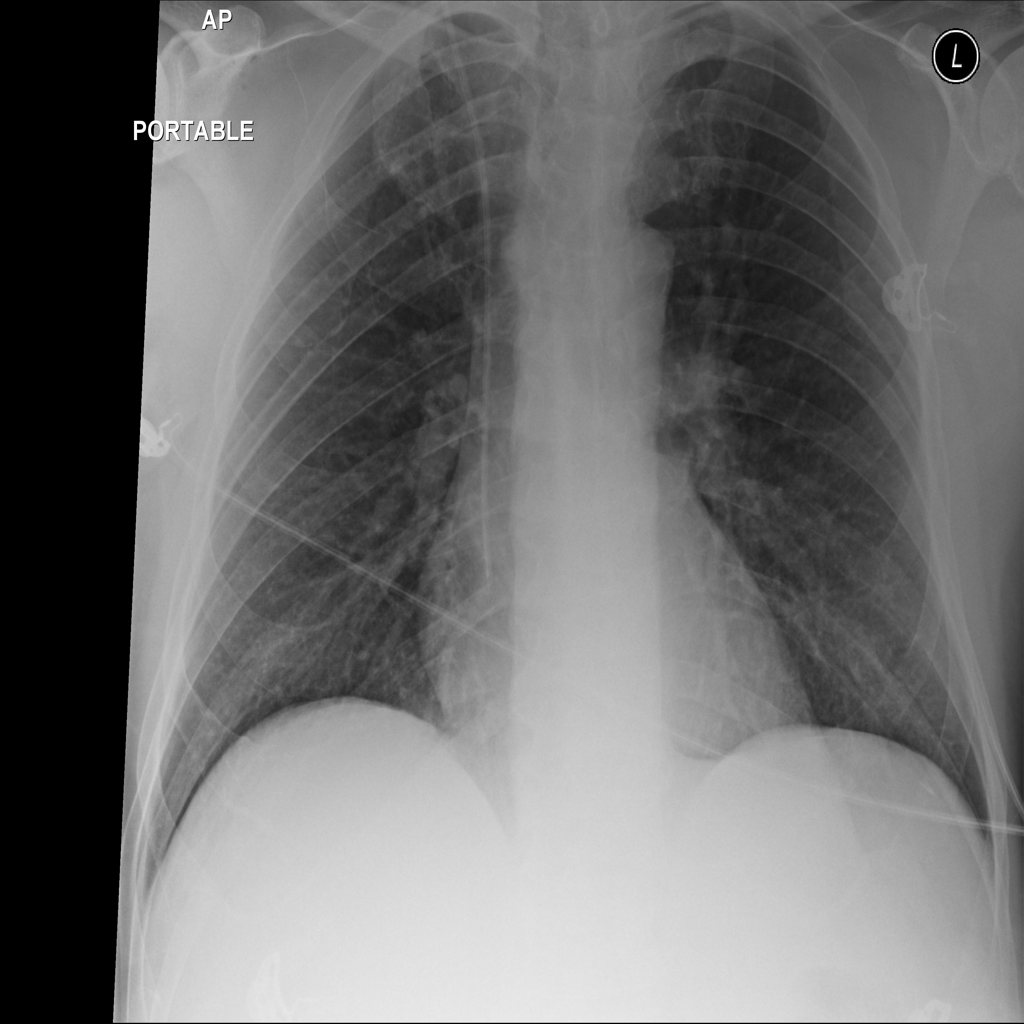}
    \includegraphics[width=.088\linewidth]{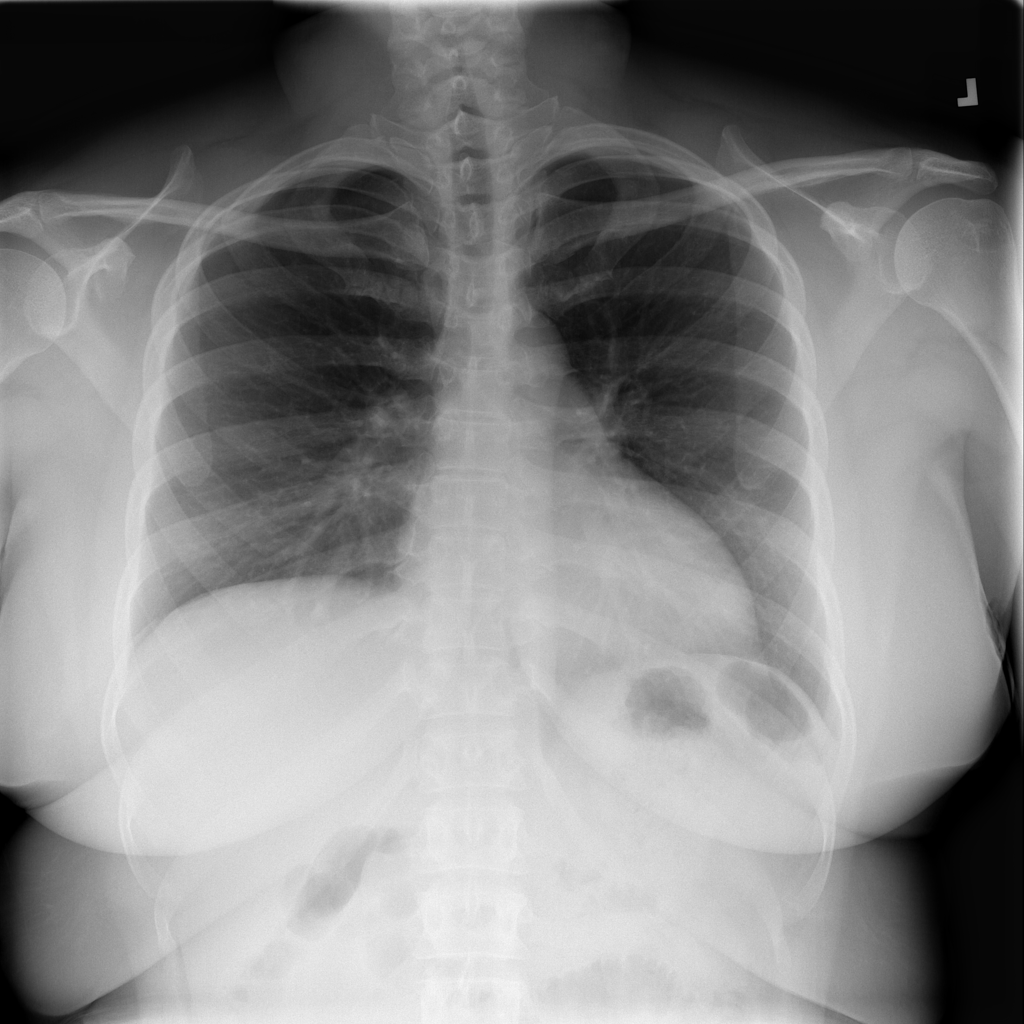}
    \includegraphics[width=.088\linewidth]{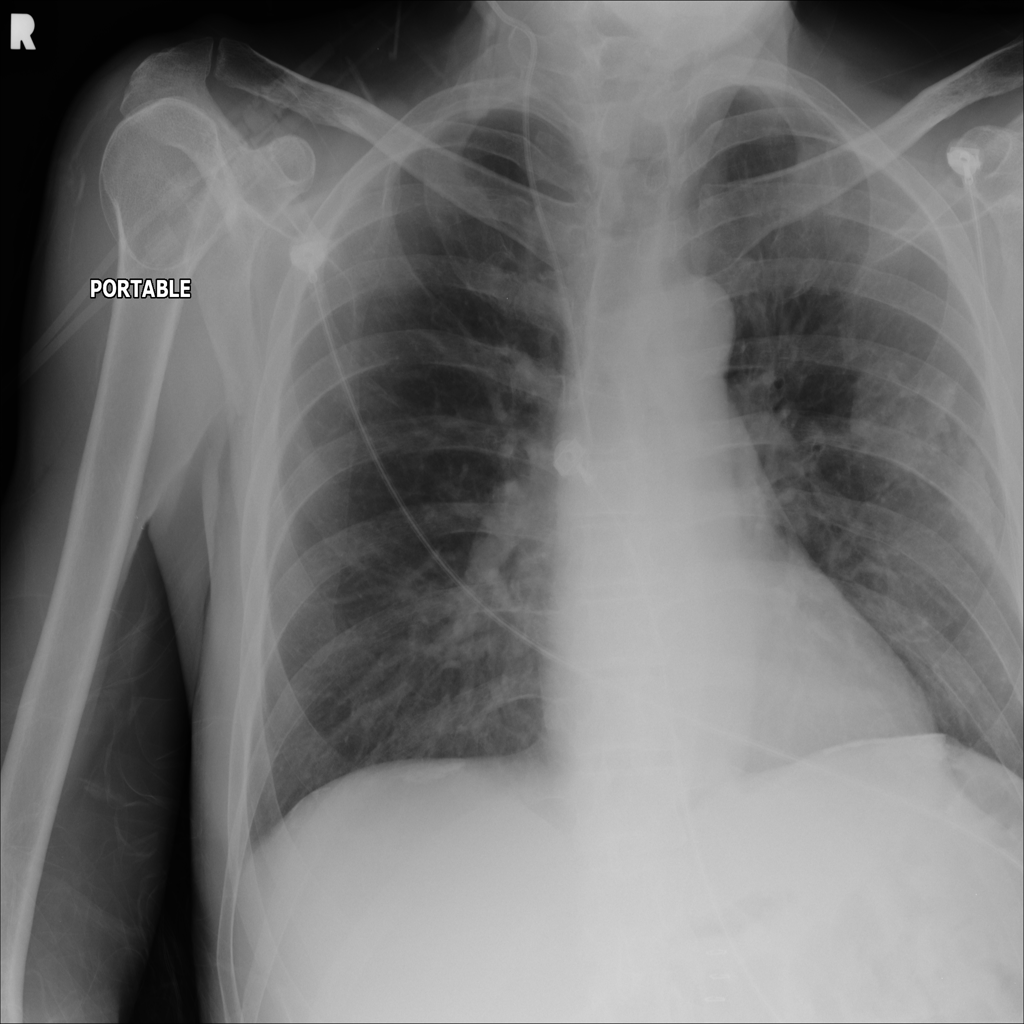}
    \caption{\small Example images from the \emph{\imagenet}, the \emph{retinal fundus photographs}, and the \chexpert datasets, respectively. The fundus photographs and chest x-rays have much higher resolution than the \imagenet images, and are classified by looking for small local variations in tissue.}
    \label{fig:example-images}
    \vspace*{-2mm}
\end{figure*}
Our primary dataset, the \retina data, consists of retinal \textit{fundus photographs} \citep{Gulshan2016Retinal}, large $587\times 587$ images of the back of the eye. These images are used to diagnose a variety of eye diseases including Diabetic Retinopathy (DR) \citep{ahsan2015dr}. DR is graded on a five-class scale of increasing severity \citep{ICDRStandards}. Grades 3 and up are \textit{referable DR} (requiring immediate specialist attention), while grades 1 and 2 correspond to
\textit{non-referable DR}. As in prior work \citep{Gulshan2016Retinal, abramoff2016improved} we evaluate via AUC-ROC on identifying referable DR.

We also study a second medical imaging dataset, \chexpert \citep{irvin2019chexpert}, which consists of chest x-ray images (resized to $224 \times 224$), which can be used to diagnose 5 different thoracic pathologies: atelectasis, cardiomegaly, consolidation, edema and pleural effusion. We evaluate our models on the AUC of diagnosing each of these pathologies. Figure~\ref{fig:example-images} shows some example images from both datasets and \imagenet, demonstrating drastic differences in visual features among those datasets.

\section{Models and Performance Evaluation of Transfer Learning}
\label{sec-models-performance}
To lay the groundwork for our study, we select multiple neural network architectures and evaluate their performance when (1) training from random initialization and (2) doing transfer learning from \imagenet. We train both standard, high performing \imagenet architectures that have been popular for transfer learning, as well as a family of significantly smaller convolutional neural networks, which achieve comparable performance on the medical tasks.

As far as we are aware, there has been little work studying the effects of transfer learning from \imagenet on smaller, non-standard \imagenet architectures. (For example, \citep{pasa2019efficient} studies a different model, but does not evaluate the effect of transfer learning.) This line of investigation is especially important in the medical setting, where large, computationally expensive models might significantly impede mobile and on-device applications. Furthermore, in standard \imagenet models, most of the parameters are concentrated at the top, to perform the $1000$-class classification. However, medical diagnosis often has considerably fewer classes -- both the retinal fundus images and chest x-rays have just $5$ classes -- likely meaning that \imagenet models are highly overparametrized.

We find that across both datasets and all models, transfer learning does not significantly affect performance. Additionally, the family of smaller lightweight convolutional networks performs comparably to standard \imagenet models, despite having significantly worse accuracy on \imagenet -- the \imagenet task is not necessarily a good indication of success on medical datasets. Finally, we observe that these conclusions also hold in the setting of very limited data.

\subsection{Description of Models}
For the standard \imagenet architectures, we evaluate ResNet50 \citep{he2018rethinking} and Inception-v3 \citep{szegedy2015going}, which have both been used extensively in medical transfer learning applications \citep{abramoff2016improved, Gulshan2016Retinal, wang2017chestx}. We also design a family of simple, smaller convolutional architectures. The basic building block for this family is the popular sequence of a (2d) convolution, followed by batch normalization \citep{ioffe2015batch} and a relu activation. Each architecture has four to five repetitions of this basic layer. We call this model family CBR. Depending on the choice of the convolutional filter size (fixed for the entire architecture), the number of channels and layers, we get a family of architectures with size ranging from a third of the standard \imagenet model size (CBR-LargeT, CBR-LargeW) to one twentieth the size (CBR-Tiny). Full architecture details are in the Appendix.

\subsection{Results}
We evaluate three repetitions of the different models and  initializations (random initialization vs pretrained weights) on the two medical tasks, with the result shown in Tables \ref{table-performances}, \ref{table-chexpert-performances}. There are two possibilities for repetitions of transfer learning: we can have a fixed set of pretrained weights and multiple training runs from that initialization, or for each repetition, first train from scratch on \imagenet and then fine-tune on the medical task. We opt for evaluating the former, as that is the standard method used in practice. For all models except for Inceptionv3, we first train on \imagenet to get the pretrained weights. For Inceptionv3, we used the pretrained weights provided by \citep{inceptionv3}.

\begin{table}\small
  \caption{\label{table-performances}\small \textbf{Transfer learning and random initialization perform comparably across both standard \imagenet architectures and simple, lightweight CNNs for AUCs from diagnosing moderate DR. Both sets of models also have similar AUCs, despite significant differences in size and complexity.} Model performance on DR diagnosis is also not closely correlated with \imagenet performance, with the small models performing poorly on \imagenet but very comparably on the medical task.}
  \centering
  \begin{tabular}{llllll}
    \toprule            \textbf{Dataset} &  \textbf{Model Architecture} & \textbf{Random Init} & \textbf{Transfer} & \textbf{Parameters} & \textbf{\imagenet Top5}\\
    \midrule
    \retina & Resnet-50 & $96.4\% \pm 0.05$ & $96.7\% \pm 0.04$ & $23570408$ &  $92.\% \pm 0.06$ \\
    \retina & Inception-v3 & $96.6\% \pm 0.13$ & $96.7\% \pm 0.05$ & $22881424$ & $93.9\%$ \\
    \retina & CBR-LargeT & $96.2\% \pm 0.04$ & $96.2\% \pm 0.04$ & $8532480$  & $77.5\% \pm 0.03$ \\
    \retina & CBR-LargeW &  $95.8\% \pm 0.04$ & $95.8\% \pm 0.05$ & $8432128$ & $75.1\% \pm 0.3$ \\
    \retina & CBR-Small &  $95.7\% \pm 0.04$ & $95.8\% \pm 0.01$ & $2108672$ & $67.6\% \pm 0.3$ \\
    \retina & CBR-Tiny &  $95.8\% \pm 0.03$ & $95.8\% \pm 0.01$ & $1076480$ & $73.5\% \pm 0.05$ \\
    \bottomrule
  \end{tabular}
  \vspace{-2mm}
\end{table}

\begin{table}[]\small
    \caption{\label{table-chexpert-performances} \small \textbf{Transfer learning provides mixed performance gains on chest x-rays.} Performances (AUC\%) of diagnosing different pathologies on the \chexpert dataset. Again we see that transfer learning does not help significantly, and much smaller models performing comparably.}
    \centering
    \begin{tabular}{lccccc}
    \toprule
    \textbf{Model Architecture} & \textbf{Atelectasis} & \textbf{Cardiomegaly} & \textbf{Consolidation} & \textbf{Edema}      & \textbf{Pleural Effusion} \\
    \midrule
    Resnet-50        & 79.52$\pm$0.31 & 75.23$\pm$0.35 & 85.49$\pm$1.32 & 88.34$\pm$1.17 & 88.70$\pm$0.13 \\
    \rowcolor{AliceBlue}
    Resnet-50 (trans) & 79.76$\pm$0.47 & 74.93$\pm$1.41 & 84.42$\pm$0.65 & 88.89$\pm$1.66 & 88.07$\pm$1.23 \\
    CBR-LargeT &        81.52$\pm$0.25 & 74.83$\pm$1.66 & 88.12$\pm$0.25 & 87.97$\pm$1.40 & 88.37$\pm$0.01 \\
    \rowcolor{AliceBlue}
    CBR-LargeT (trans) & 80.89$\pm$1.68 & 76.84$\pm$0.87 & 86.15$\pm$0.71 & 89.03$\pm$0.74 & 88.44$\pm$0.84 \\
    CBR-LargeW &         79.79$\pm$0.79 & 74.63$\pm$0.69 & 86.71$\pm$1.45 & 84.80$\pm$0.77 & 86.53$\pm$0.54 \\
    \rowcolor{AliceBlue}
    CBR-LargeW (trans) & 80.70$\pm$0.31 & 77.23$\pm$0.84 & 86.87$\pm$0.33 & 89.57$\pm$0.34 & 87.29$\pm$0.69 \\
    CBR-Small &          80.43$\pm$0.72 & 74.36$\pm$1.06 & 88.07$\pm$0.60 & 86.20$\pm$1.35 & 86.14$\pm$1.78 \\
    \rowcolor{AliceBlue}
    CBR-Small (trans) &  80.18$\pm$0.85 & 75.24$\pm$1.43 & 86.48$\pm$1.13 & 89.09$\pm$1.04 & 87.88$\pm$1.01 \\
    CBR-Tiny &           80.81$\pm$0.55 & 75.17$\pm$0.73 & 85.31$\pm$0.82 & 84.87$\pm$1.13 & 85.56$\pm$0.89 \\
    \rowcolor{AliceBlue}
    CBR-Tiny (trans)  &  80.02$\pm$1.06 & 75.74$\pm$0.71 & 84.28$\pm$0.82 & 89.81$\pm$1.08 & 87.69$\pm$0.75 \\
    \bottomrule
    \end{tabular}
\end{table}
Table \ref{table-performances} shows the model performances on the \retina data (AUC of identifying moderate Diabetic Retinopathy (DR), described in Section \ref{sec-data-setup}), along with \imagenet top 5 accuracy. Firstly, we see that transfer learning has minimal effect on performance, not helping the smaller CBR architectures at all, and only providing a fraction of a percent gain for Resnet and Inception. Next, we see that despite the significantly lower performance of the CBR architectures on \imagenet, they perform very comparably to Resnet and Inception on the \retina task. These same conclusions are seen on the chest x-ray results, Table \ref{table-chexpert-performances}. Here we show the performance AUC for the five different pathologies (Section \ref{sec-data-setup}). We again observe mixed gains from transfer learning. For Atelectasis, Cardiomegaly and Consolidation, transfer learning performs slightly worse, but helps with Edema and Pleural Effusion.

\subsection{The Very Small Data Regime}
We conducted additional experiments to study the effect of transfer learning in the very small data regime. Most medical datasets are significantly smaller than \imagenet, which is also the case for our two datasets. However, our datasets still have around two hundred thousand examples, and other settings many only have a few thousand. To study the effects in this very small data regime, we trained models on only $5000$ datapoints on the \retina dataset, and examined the effect of transfer learning.
\begin{table*}[h]
  \centering
   \begin{tabular}{llll}
    \toprule            \textbf{Model} & \textbf{Rand Init} & \textbf{Pretrained}   \\
    \midrule
   Resnet50 & \hspace*{3mm} $92.2 \%$ & \hspace*{3mm} $94.6 \%$  \\
   CBR-LargeT & \hspace*{3mm} $93.6 \%$ & \hspace*{3mm} $93.9 \%$  \\
   CBR-LargeW & \hspace*{3mm} $93.6 \%$ & \hspace*{3mm} $93.7 \%$  \\
    \bottomrule
  \end{tabular}
  \caption{\small \textbf{Benefits of transfer learning in the small data regime are largely due to architecture size.} AUCs when training on the \retina task with only $5000$ datapoints. We see a bigger gap between random initialization and transfer learning for Resnet (a large model), but not for the smaller CBR models.}
  \label{table-small-data-regime}
\end{table*}
The results, in Table \ref{table-small-data-regime}, suggest that while transfer learning has a bigger effect with very small amounts of data, there is a confounding effect of model size -- transfer primarily helps the large models (which are designed to be trained with a million examples) and smaller models again show little difference between transfer and random initialization.

\section{Representational Analysis of the Effects of Transfer}
\label{sec-analysis}
In Section \ref{sec-models-performance} we saw that transfer learning and training from random initialization result in very similar performance across different neural architectures and tasks. This gives rise to some natural questions about the effect of transfer learning on the kinds of \textit{representations} learned by the neural networks. Most fundamentally, does transfer learning in fact result in any representational differences compared to training from random initialization? Or are the effects of the initialization lost? Does feature reuse take place, and if so, where exactly? In this section, we provide some answers to these basic questions. Our approach directly analyzes and compares the hidden representations learned by different populations of neural networks, using (SV)CCA \citep{raghu2017svcca, morcos2018insights}, revealing an important dependence on \textit{model size}, and differences in behavior between lower and higher layers. These insights, combined with results Section \ref{sec-convergence} suggest new, hybrid approaches to transfer learning.

\textbf{Quantitatively Studying Hidden Representations with (SV)CCA}
To understand how pretraining affects the features and representations learned by the models, we would like to (quantitatively) study the learned intermediate functions (latent layers). Analyzing latent representations is challenging due to their complexity and the lack of any simple mapping to inputs, outputs or other layers. A recent tool that effectively overcomes these challenges is (Singular Vector) Canonical Correlation Analysis, (SV)CCA \citep{raghu2017svcca, morcos2018insights}, which has been used to study latent representations through training, across different models, alternate training objectives, and other properties \citep{raghu2017svcca, morcos2018insights, saphra2018understanding, magill2018neural, gotmare2018closer, kudugunta2019investigating, voita2019bottom}. Rather than working directly with the model parameters or neurons, CCA works with {\em neuron activation vectors} --- the ordered collection of outputs of the neuron on a sequence of inputs.  Given the activation vectors for two sets of neurons (say, corresponding to distinct layers), CCA seeks linear combinations of each that are as correlated as possible.  We adapt existing CCA methods to prevent the size of the activation sets from overwhelming the computation in large models (details in Appendix \ref{sec-app-cca-details}), and apply them to compare the latent representations of corresponding hidden layers of different pairs of neural networks, giving a CCA similarity score of the learned intermediate functions.

\textbf{Transfer Learning and Random Initialization Learn Different Representations}
\begin{figure}
\centering
\begin{tabular}{c}
\hspace*{-10mm}\includegraphics[width=0.8\columnwidth]{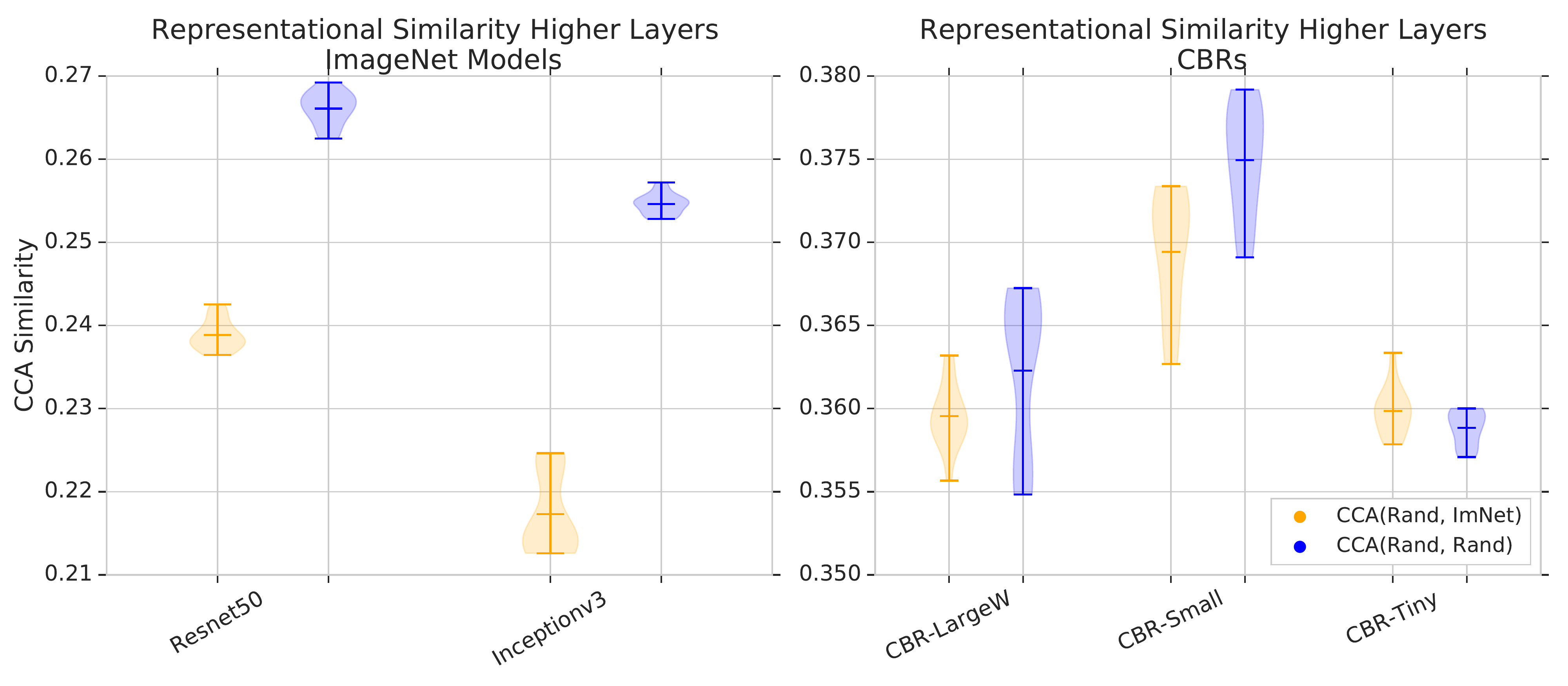}
\end{tabular}
\caption{\small \textbf{Pretrained weights give rise to different hidden representations than training from random initialization for large models.} We compute CCA similarity scores between representations learned using pretrained weights and those from random initialization. We do this for the top two layers (or stages for Resnet, Inception) and average the scores, plotting the results in orange. In blue is a baseline similarity score, for representations trained from different random initializations. We see that representations learned from random initialization are more similar to each other than those learned from pretrained weights for larger models, with less of a distinction for smaller models.}
\label{fig:ccas-of-populations}
\end{figure}
Our first experiment uses CCA to compare the similarity of the hidden representations learned when training from pretrained weights to those learned when training from random initialization. We use the representations learned at the top two layers (for CBRs) or stages (for Resnet, Inception) before the output layer, averaging their similarity scores. As a baseline to compare to, we also look at CCA similarity scores for the same representations when training from random initialization with two different seeds (different initializations and gradient updates). The results are shown in Figure \ref{fig:ccas-of-populations}. For larger models (Resnet, Inception), there is a clear difference between representations, with the similarity of representations between training from random initialization and pretrained weights (orange) noticeably lower than representations learned independently from different random initializations (blue). However for smaller models (CBRs), the functions learned are more similar.

\textbf{Larger Models Change Less Through Training}
The reasons underlying this difference between larger and smaller models becomes apparent as we further study the hidden representations of all the layers. We find that \textit{larger models change much less during training}, especially in the lowest layers. This is true \textit{even when they are randomly initialized}, ruling out feature reuse as the sole cause, and implying their overparametrization for the task. This is in line with other recent findings \citep{zhang2019all}.
\begin{figure}
\centering
\begin{tabular}{ccc}
\hspace*{-22mm} \includegraphics[width=0.42\columnwidth]{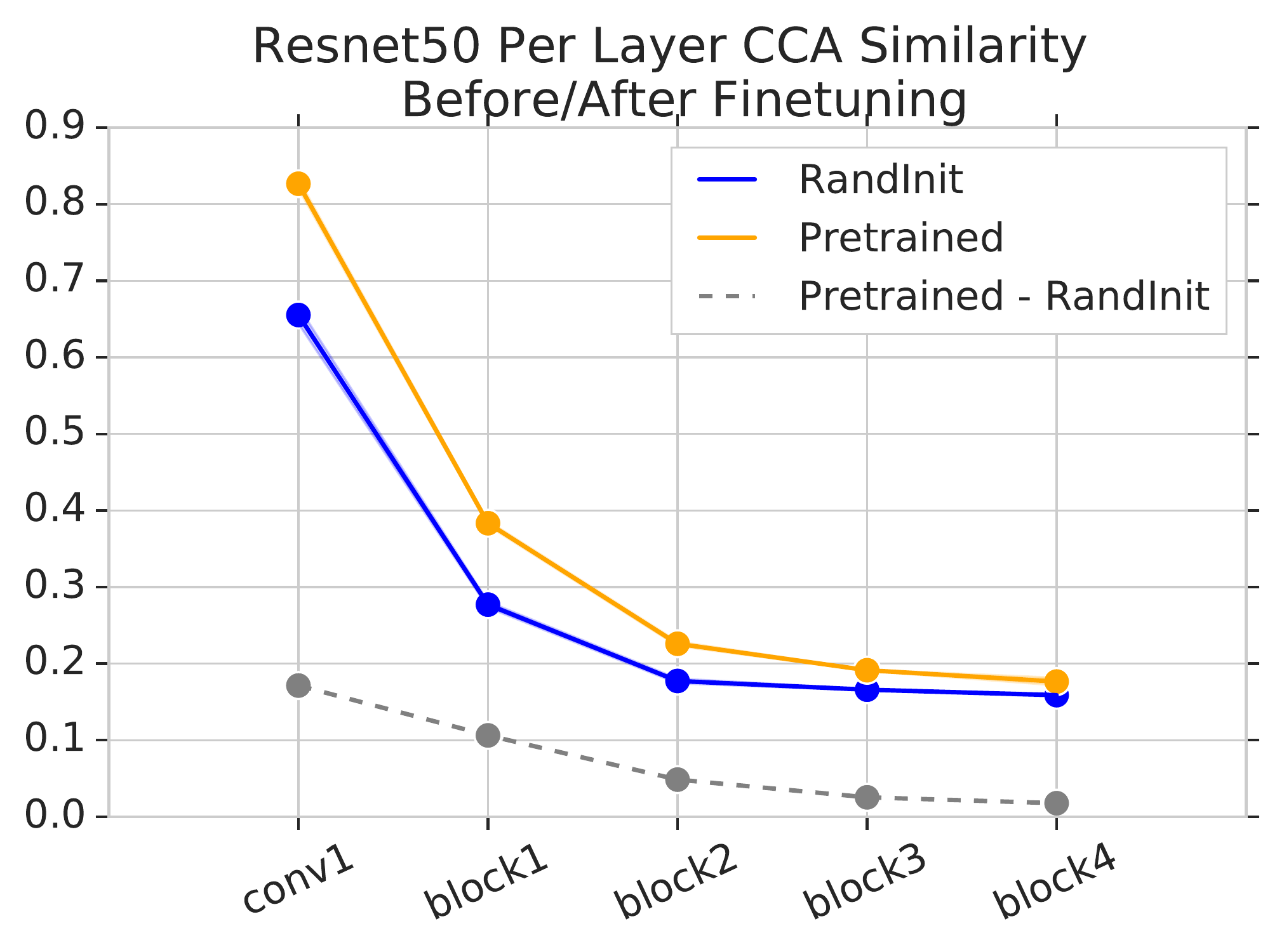} &
\hspace*{-5mm} \includegraphics[width=0.42\columnwidth]{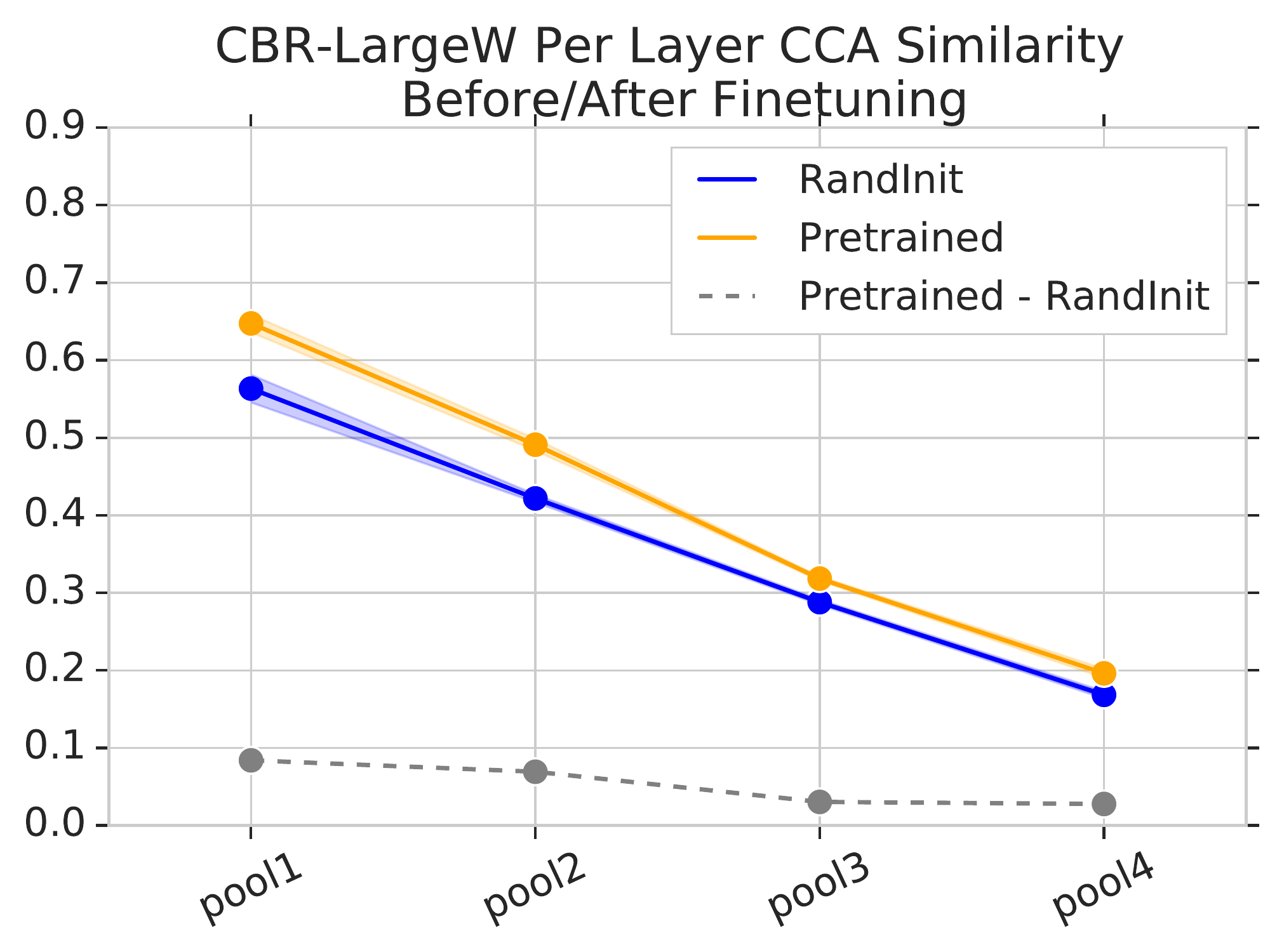}  &
\hspace*{-5mm} \includegraphics[width=0.42\columnwidth]{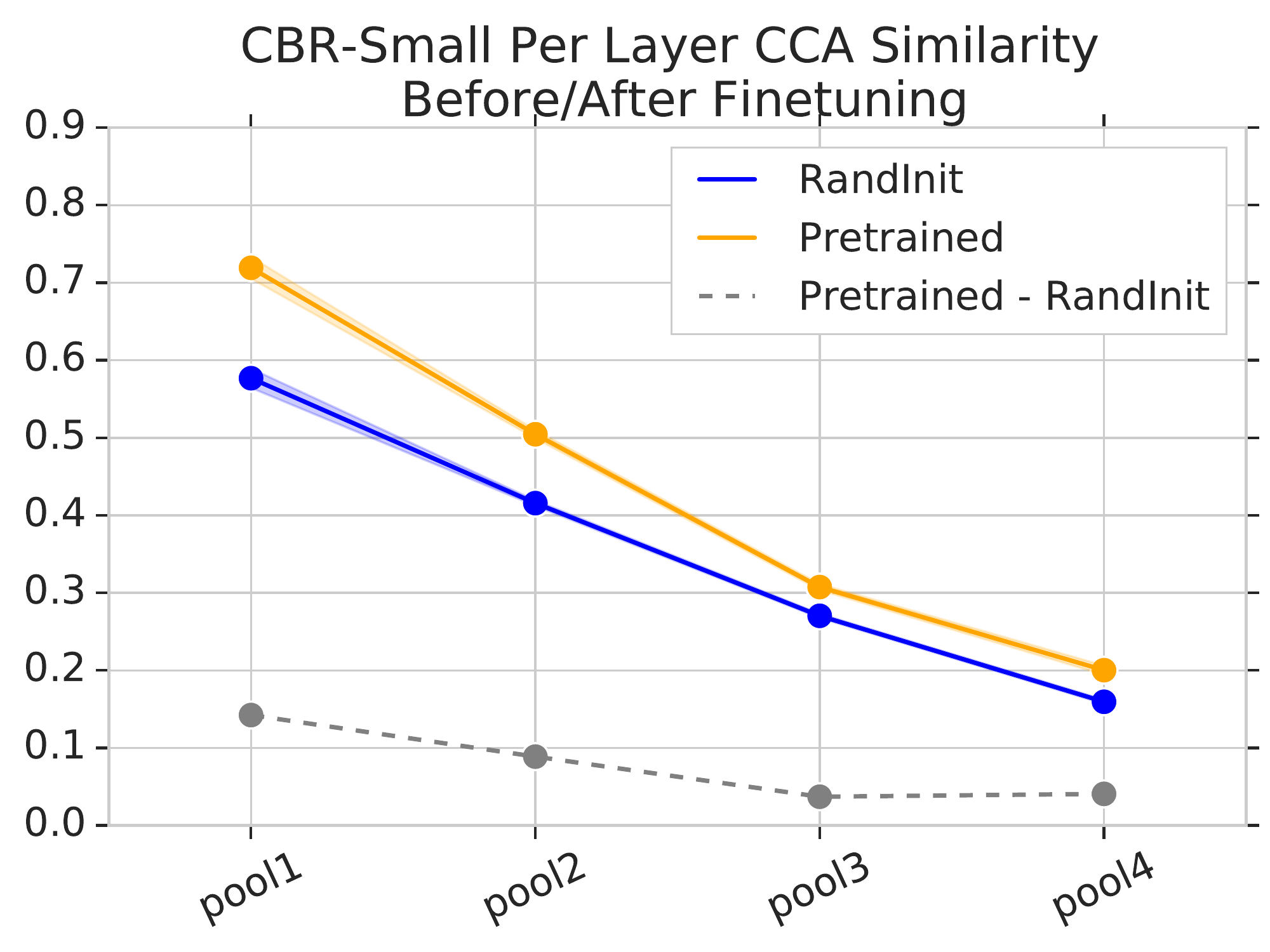}
\end{tabular}
\caption{\small \textbf{Per-layer CCA similarities before and after training on medical task.} For all models, we see that the lowest layers are most similar to their initializations, and this is especially evident for Resnet50 (a large model). We also see that feature reuse is mostly restricted to the bottom two layers (stages for Resnet) --- the only place where similarity with initialization is significantly higher for pretrained weights (grey dotted lines shows the difference in similarity scores between pretrained and random initialization).}
\label{fig:cca-before-after-finetuning}
\end{figure}

\begin{figure}
\centering
\begin{tabular}{cccc}
\hspace*{-30mm} \includegraphics[width=0.35\columnwidth]{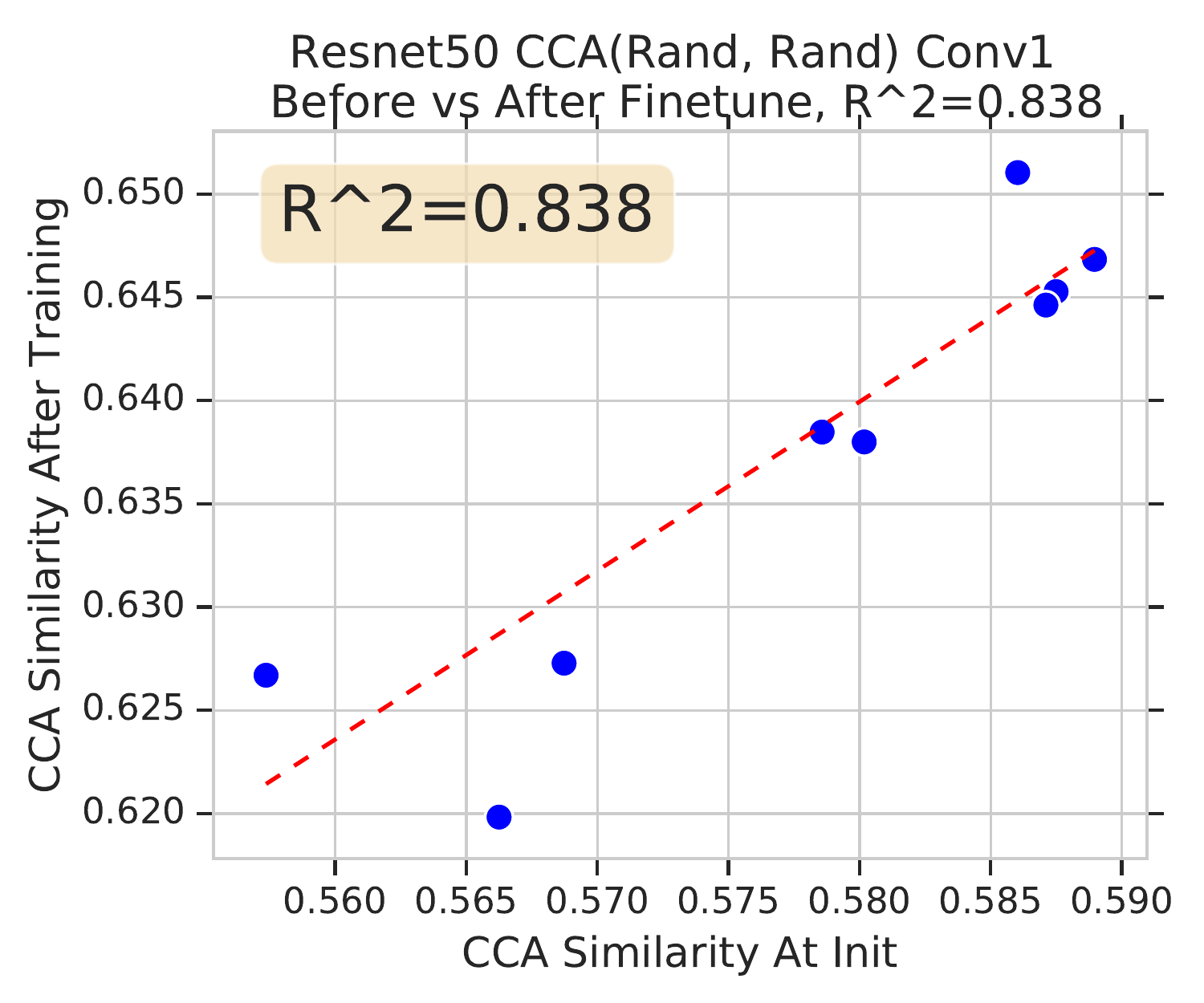} &
\hspace*{-5mm} \includegraphics[width=0.35\columnwidth]{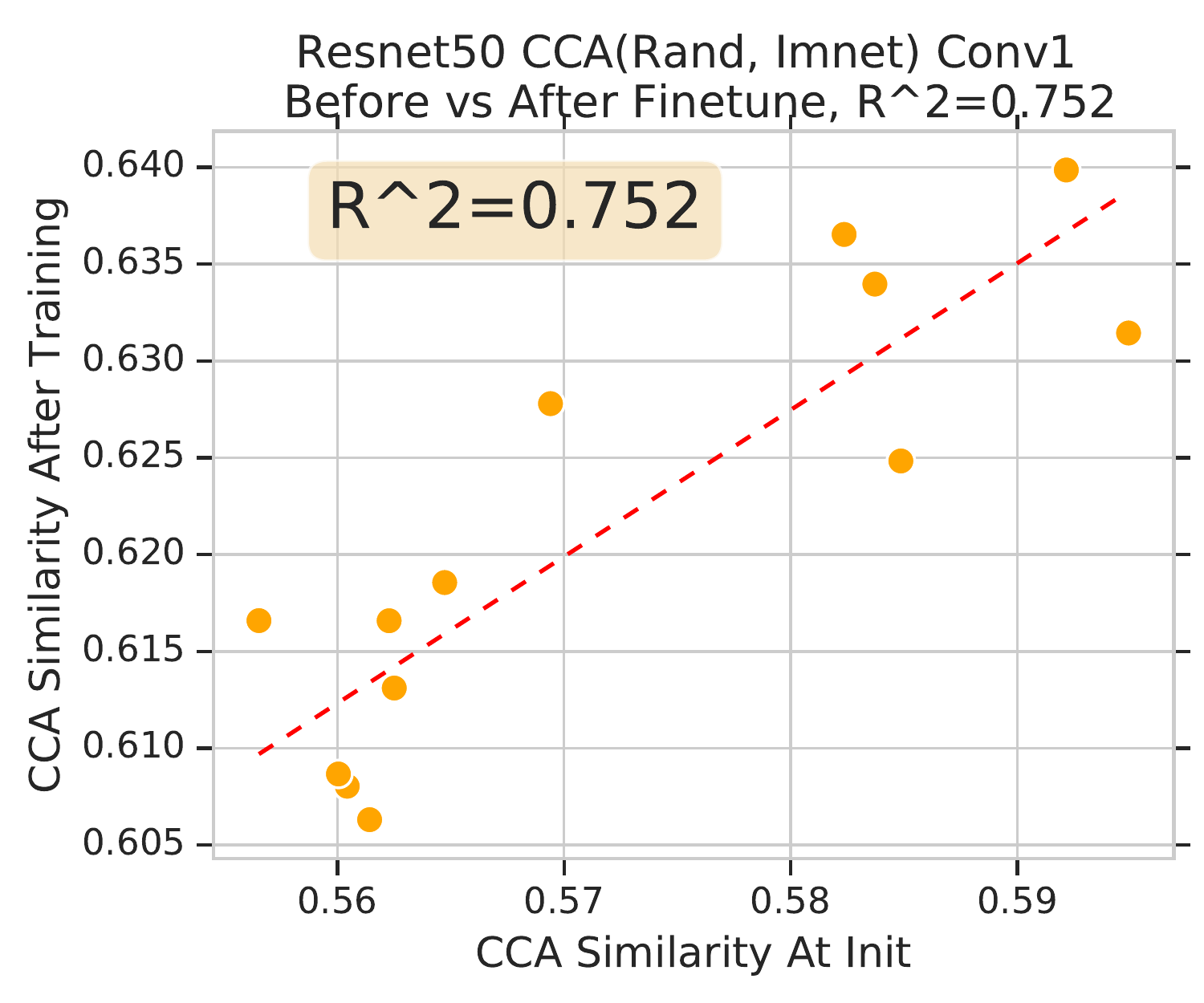}  &
\hspace*{-5mm} \includegraphics[width=0.35\columnwidth]{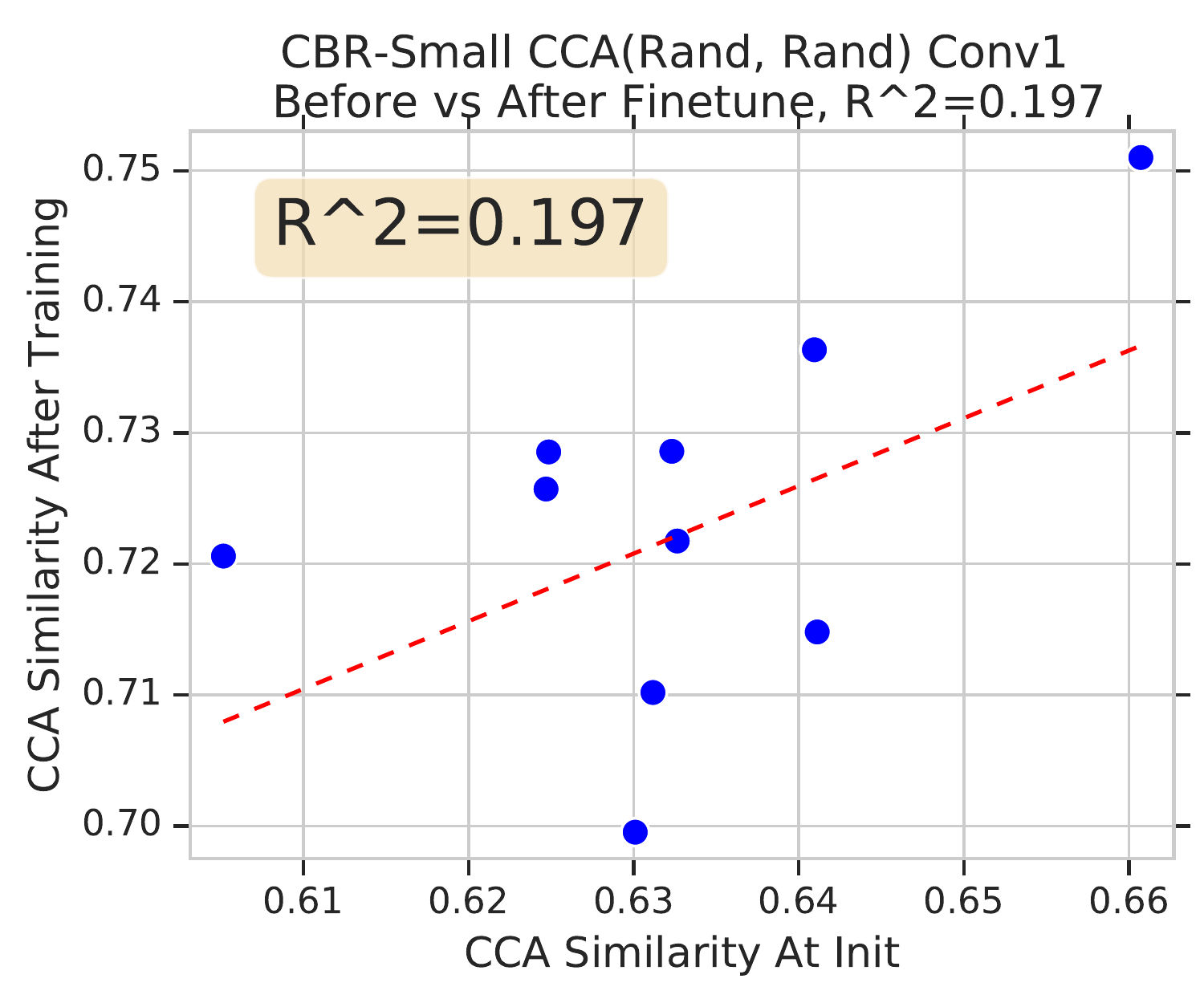} &
\hspace*{-5mm} \includegraphics[width=0.35\columnwidth]{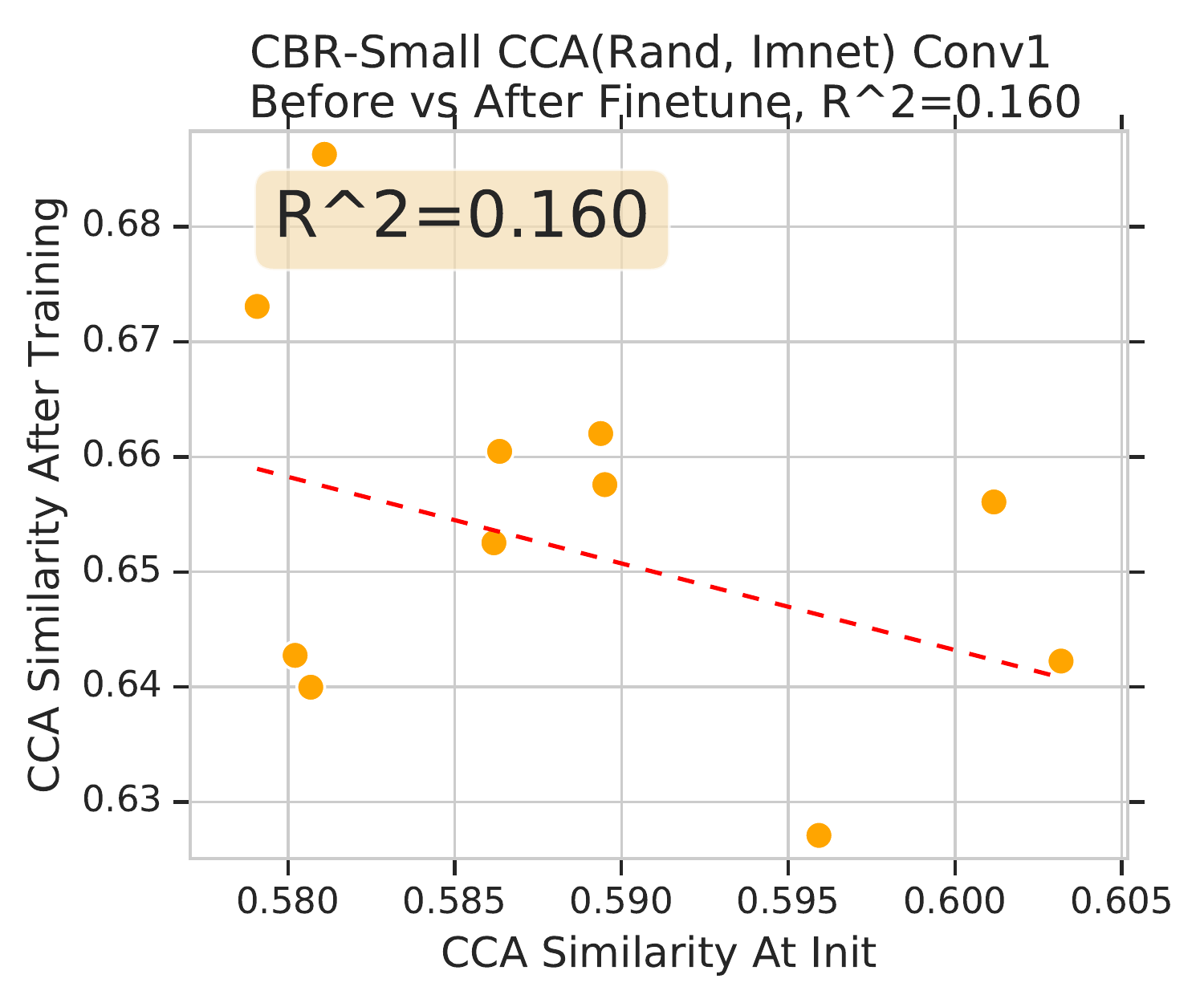} \\
\hspace*{-30mm} \includegraphics[width=0.35\columnwidth]{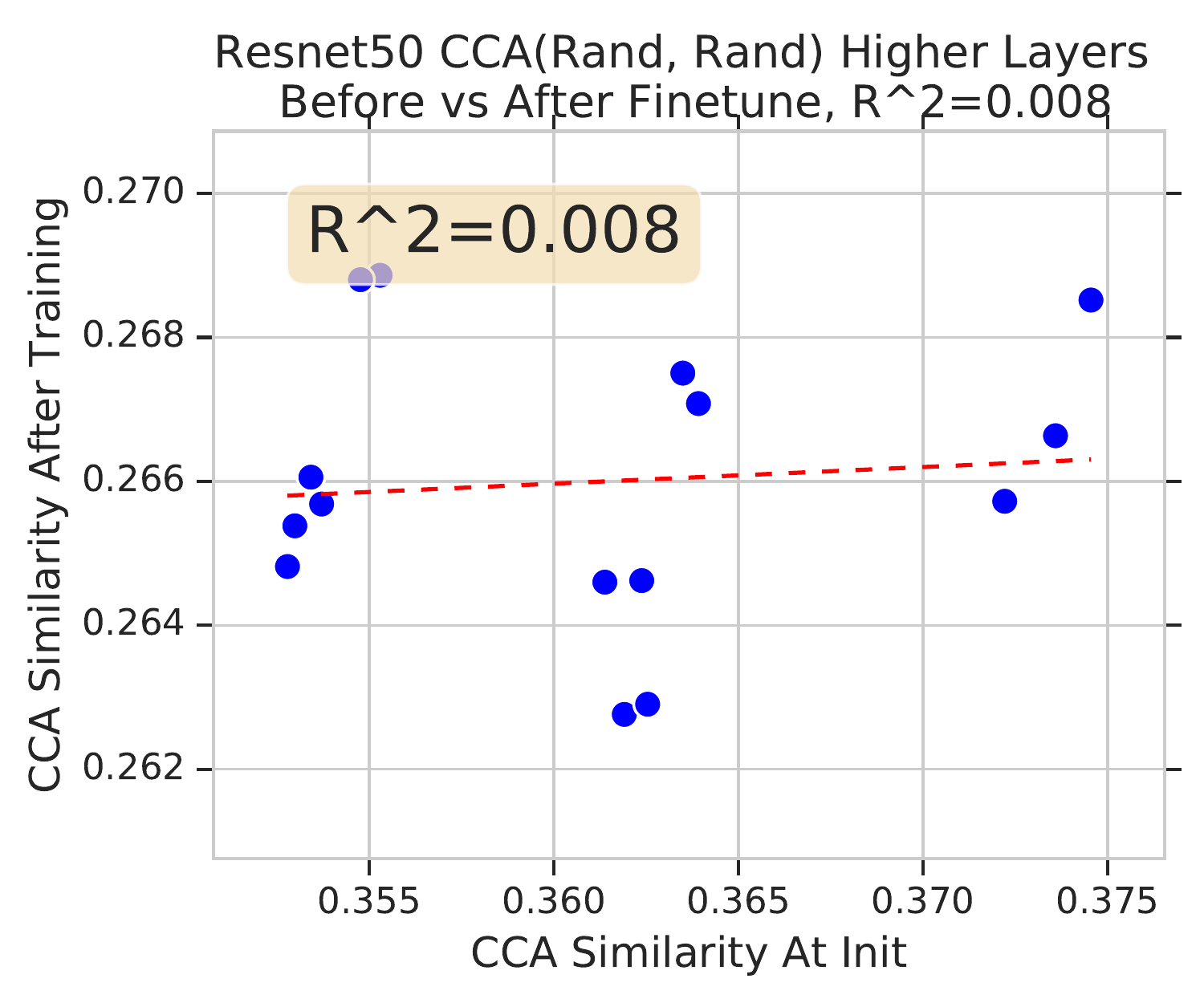} &
\hspace*{-5mm} \includegraphics[width=0.35\columnwidth]{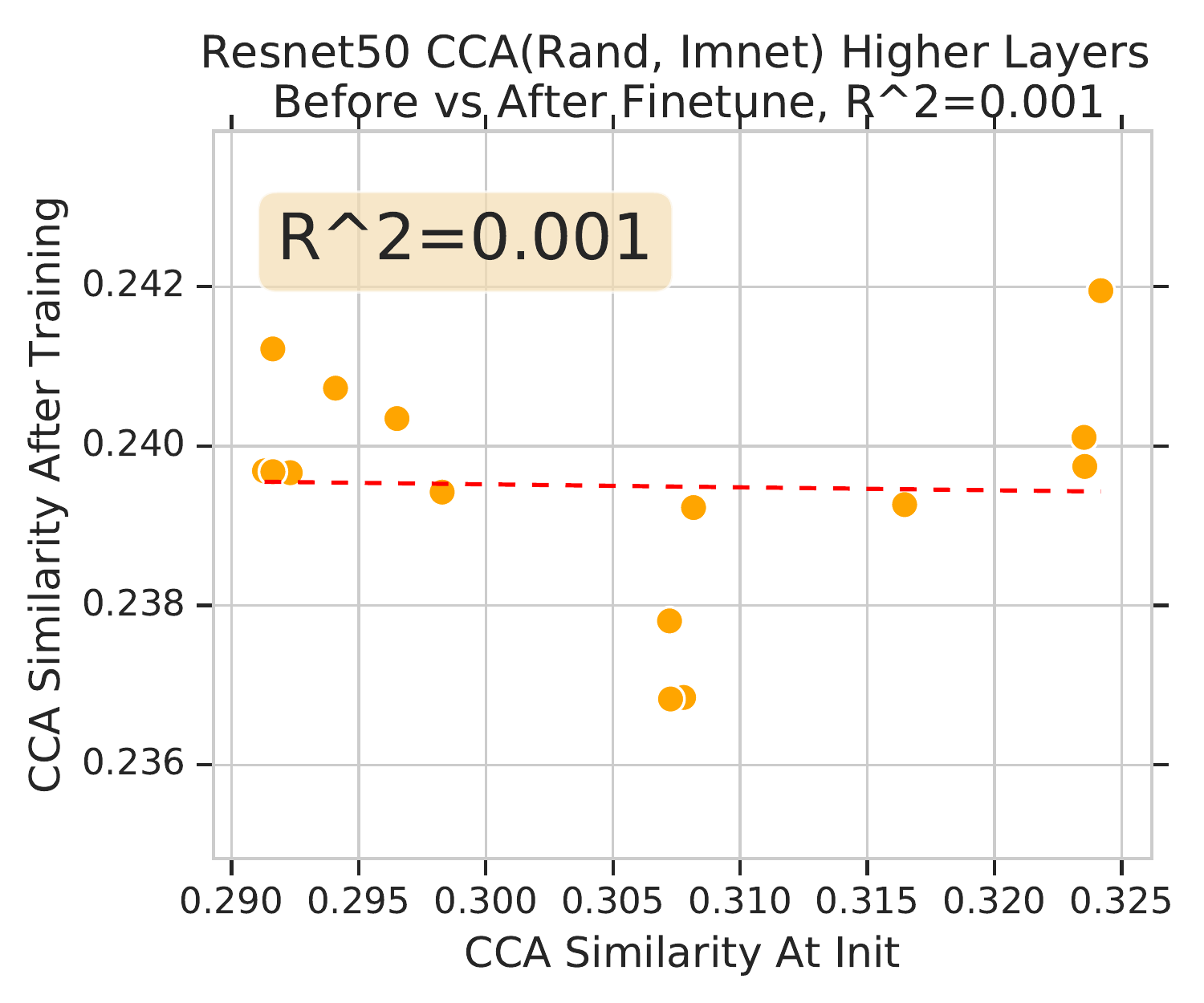}  &
\hspace*{-5mm} \includegraphics[width=0.35\columnwidth]{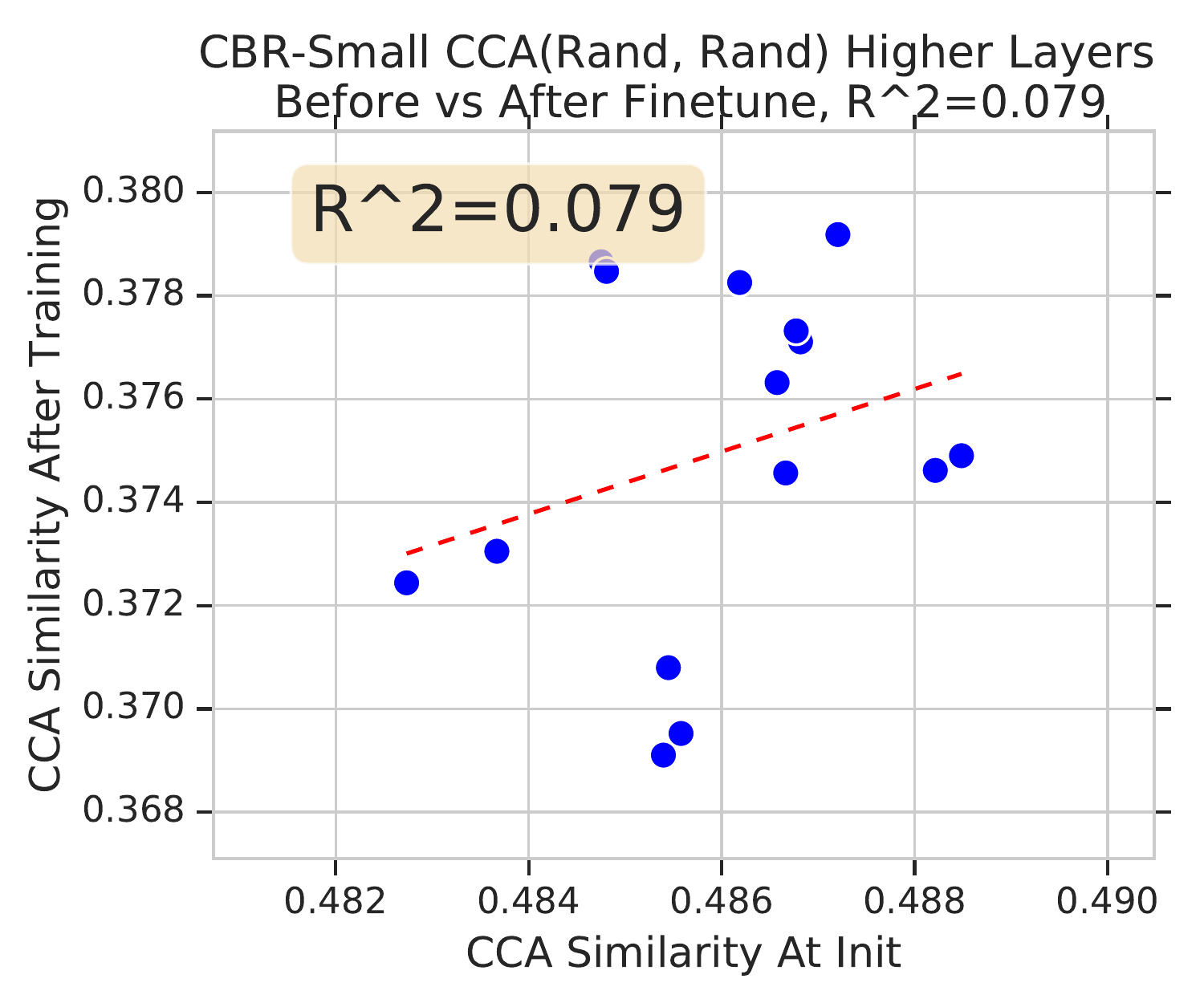} &
\hspace*{-5mm} \includegraphics[width=0.35\columnwidth]{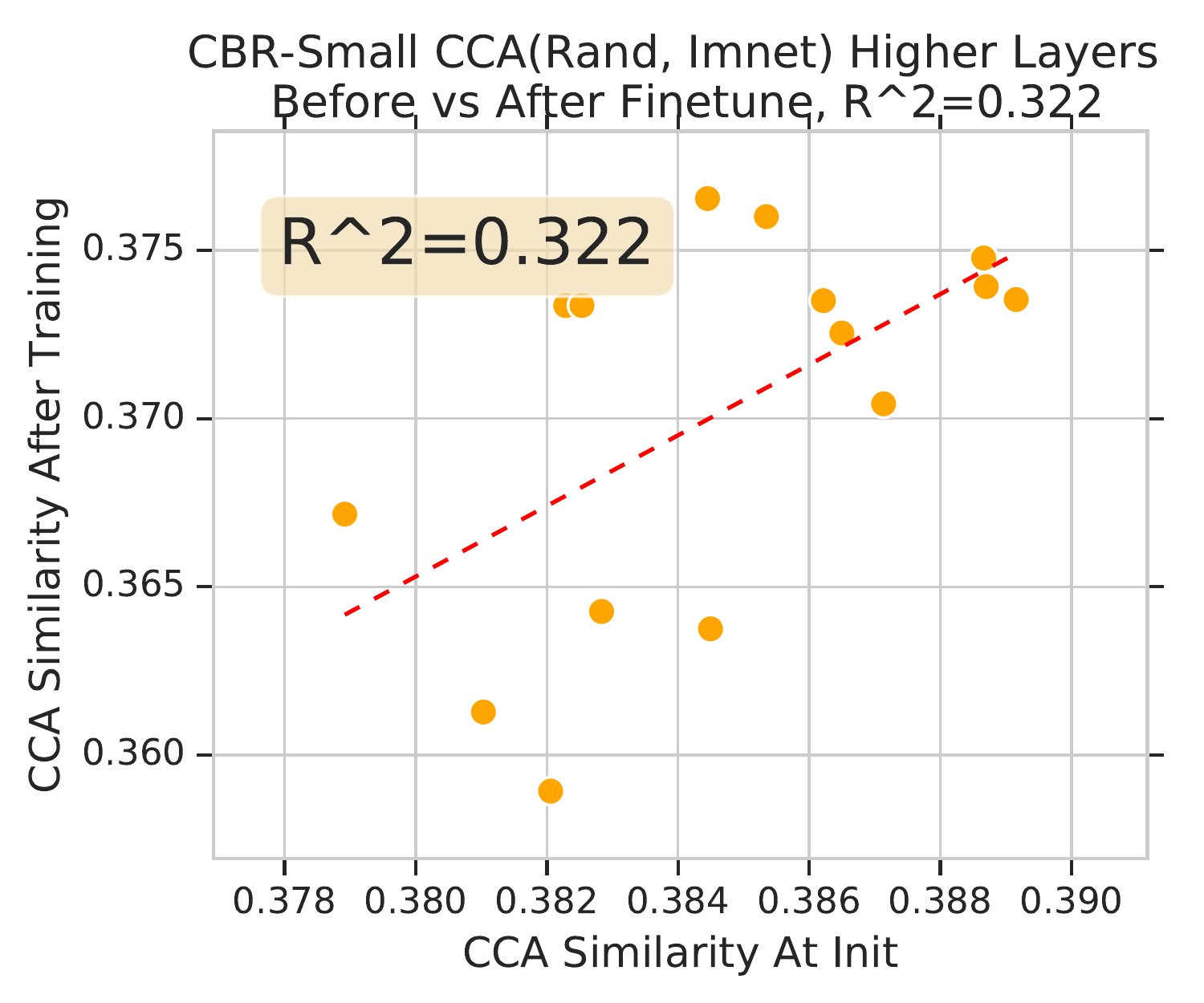}
\end{tabular}
\caption{\small  \textbf{Large models move less through training at lower layers: similarity at initialization is highly correlated with similarity at convergence for large models.} We plot CCA similarity of Resnet (conv1) initialized randomly and with pretrained weights at (i) initialization, against (ii) CCA similarity of the converged representations (top row second from left.) We also do this for two different random initializations (top row, left). In \textit{both} cases (even for random initialization), we see a surprising, strong correlation between similarity at initialization and similarity after convergence ($R^2 = 0.75, 0.84$). This is not the case for the smaller CBR-Small model, illustrating the overparametrization of Resnet for the task. Higher must likely change much more for good task performance.}
\label{fig:cca-start-end-conv1}
\end{figure}
In Figure \ref{fig:cca-before-after-finetuning}, we look at per-layer representational similarity before/after finetuning, which shows that the lowest layer in Resnet (a large model), is significantly more similar to its initialization than in the smaller models. This plot also suggests that any serious feature reuse is restricted to the lowest couple of layers, which is where similarity before/after training is clearly higher for pretrained weights vs random initialization. In Figure \ref{fig:cca-start-end-conv1}, we plot the CCA similarity scores between representations using pretrained weights and random initialization \textit{at initialization} vs after training, for the lowest layer (conv1) as well as higher layers, for Resnet and CBR-Small. Large models changing less through training is evidenced by a surprising correlation between the CCA similarities for Resnet conv1, which is not true for higher layers or the smaller CBR-Small model.

\textbf{Filter Visualizations and the Absence of Gabors}
\label{sec-feature-viz}
As a final study of how pretraining affects the model representations, we visualize some of the filters of conv1 for Resnet and CBR-Small (both 7x7 kernels), before and after training on the \retina task. The filters are shown in Figure \ref{fig:retina-weight-vis}, with visualizations for chest x-rays in the Appendix. These add evidence to the aformentioned observation: the Resnet filters change much less than those of CBR-Small.  In contrast, CBR-Small moves more from its initialization, and has more similar learned filters in random and pretrained initialization. Interestingly, CBR-Small does not appear to learn Gabor filters when trained from scratch (bottom row second column). Comparing the third and fourth columns of the bottom row, we see that CBR-Small even erases some of the Gabor filters that it is initialized with in the pretrained weights.

\begin{figure}
    \centering
    \begin{subfigure}{.24\linewidth}
        \includegraphics[width=\linewidth]{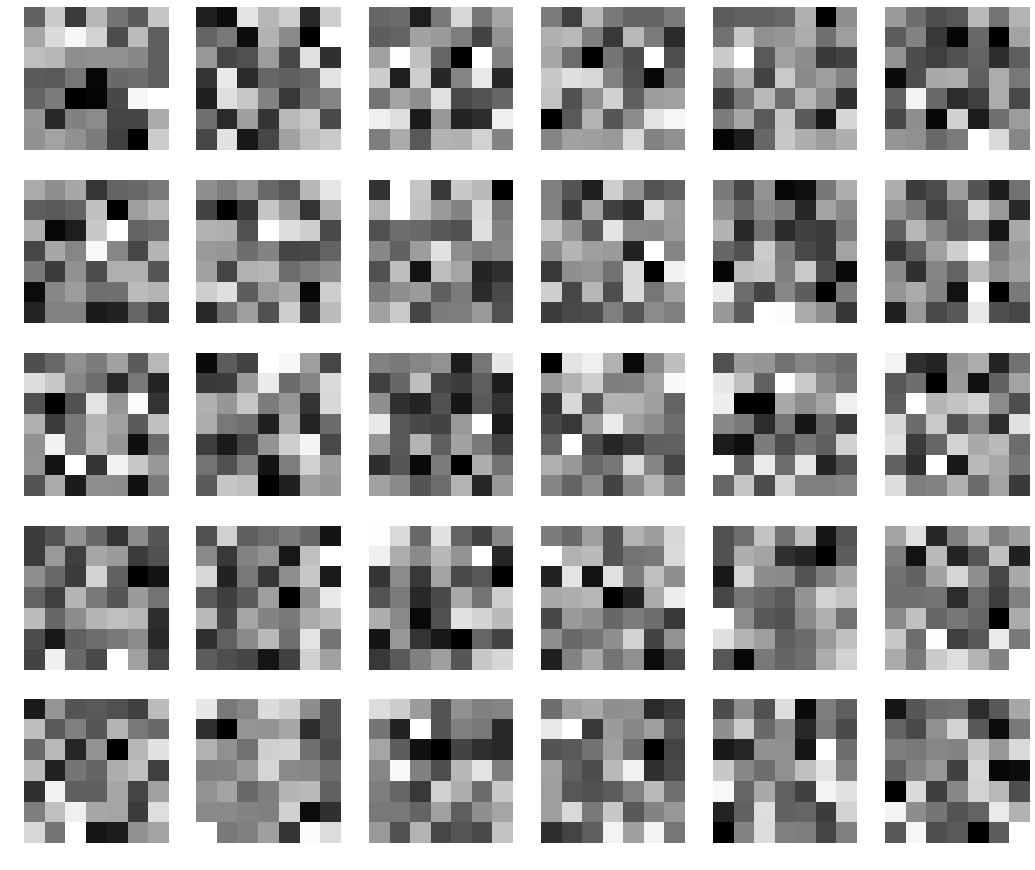}
        \caption{Resnet Init}
    \end{subfigure}
    \begin{subfigure}{.24\linewidth}
        \includegraphics[width=\linewidth]{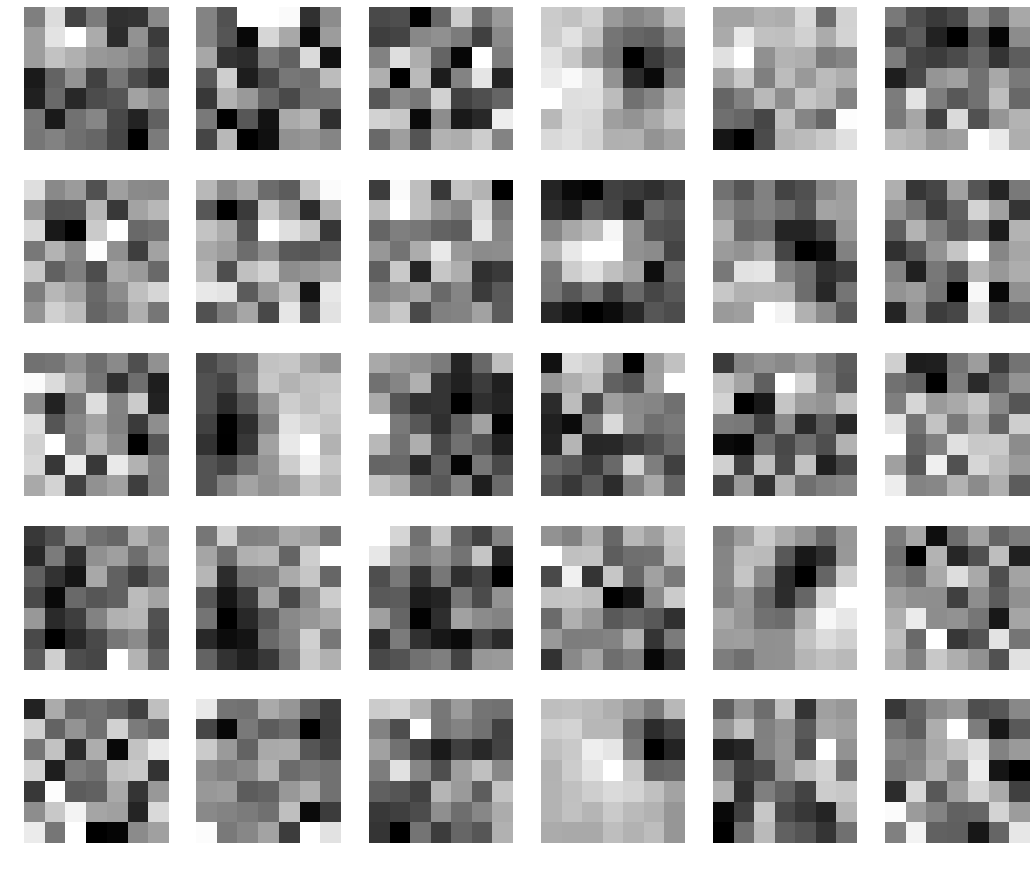}
        \caption{Resnet Final}
    \end{subfigure}
    \hspace{2mm}
    \begin{subfigure}{.24\linewidth}
        \includegraphics[width=\linewidth]{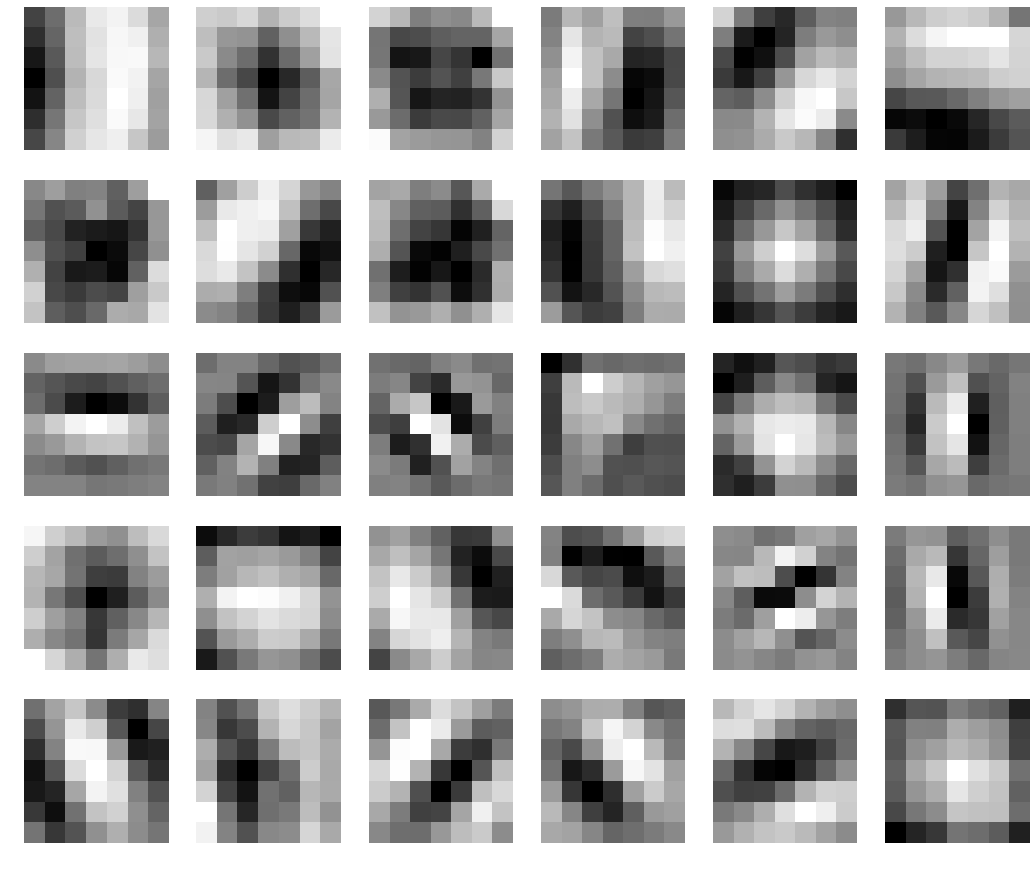}
        \caption{Res-trans Init}
    \end{subfigure}
    \begin{subfigure}{.24\linewidth}
        \includegraphics[width=\linewidth]{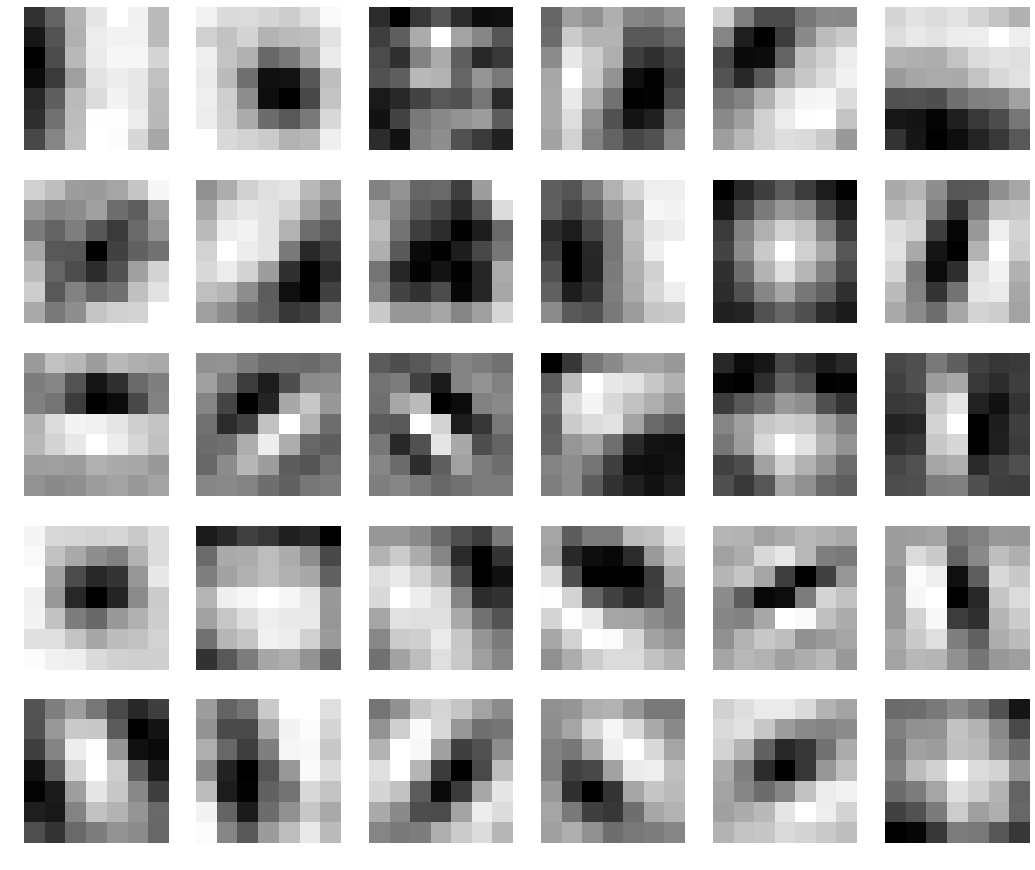}
        \caption{Res-trans final}
    \end{subfigure}
    \begin{subfigure}{.24\linewidth}
        \includegraphics[width=\linewidth]{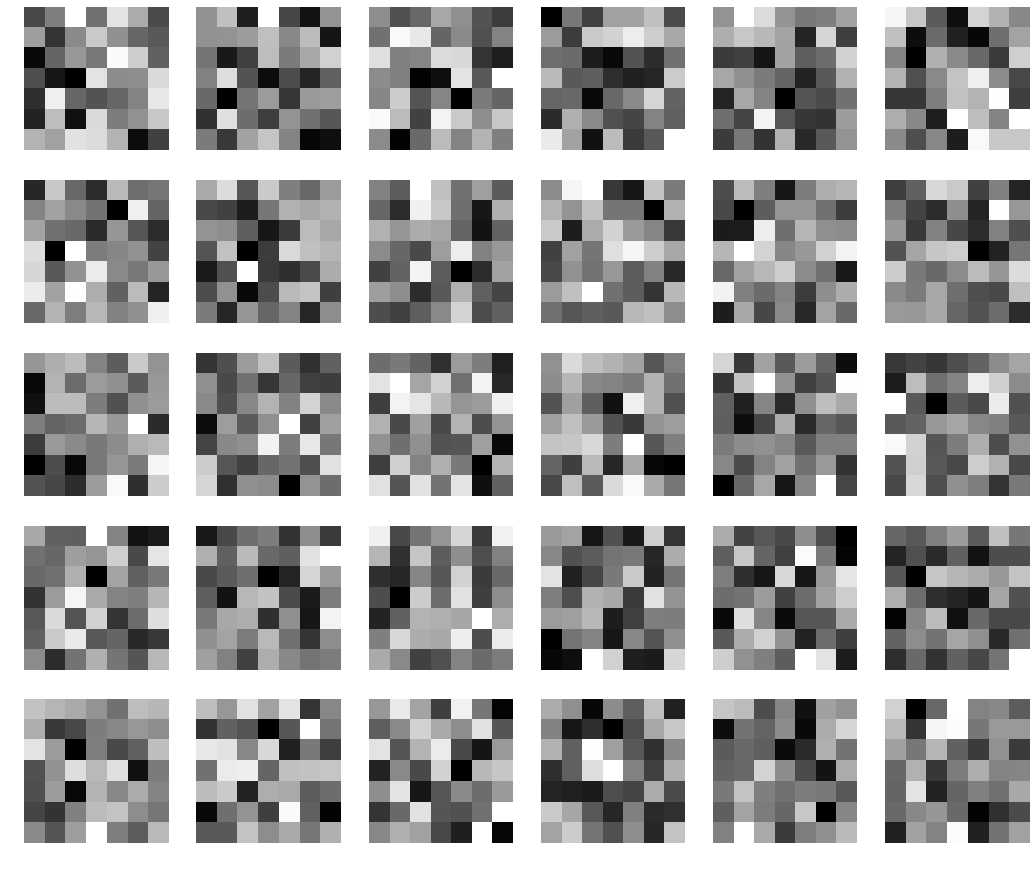}
        \caption{CBR-Small Init}
    \end{subfigure}
    \begin{subfigure}{.24\linewidth}
        \includegraphics[width=\linewidth]{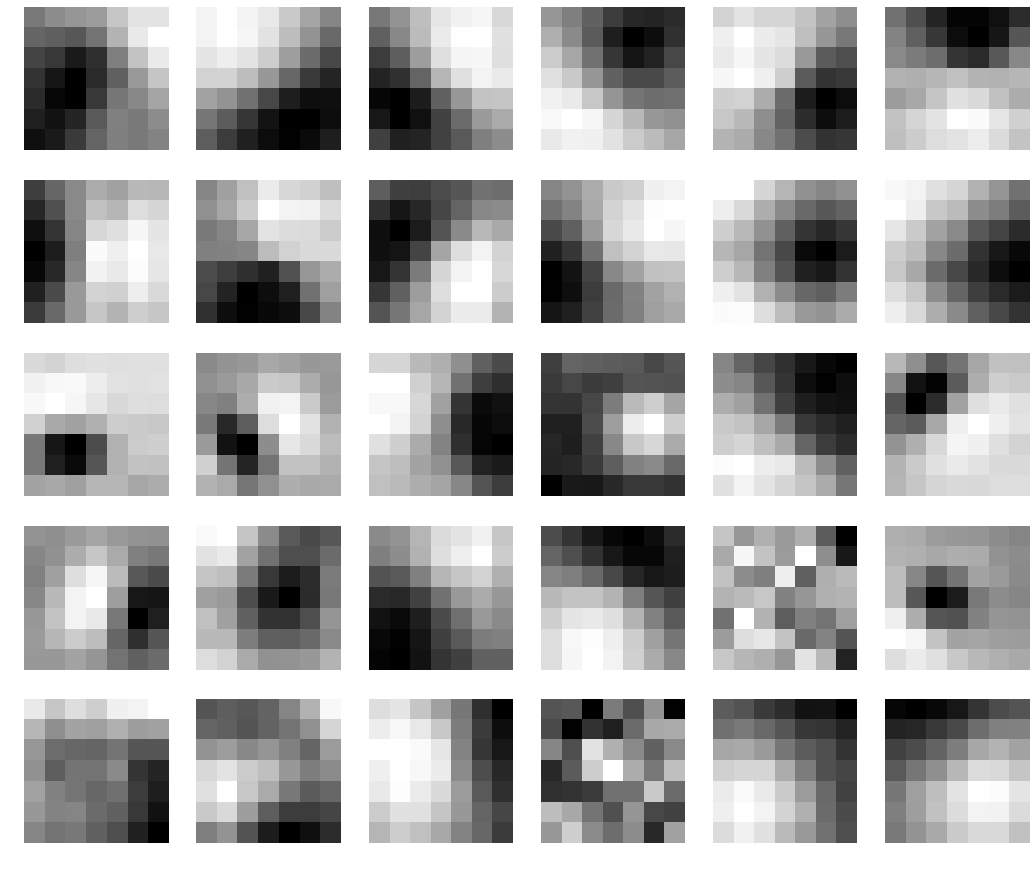}
        \caption{CBR-Small Final}
    \end{subfigure}
    \hspace{2mm}
    \begin{subfigure}{.24\linewidth}
        \includegraphics[width=\linewidth]{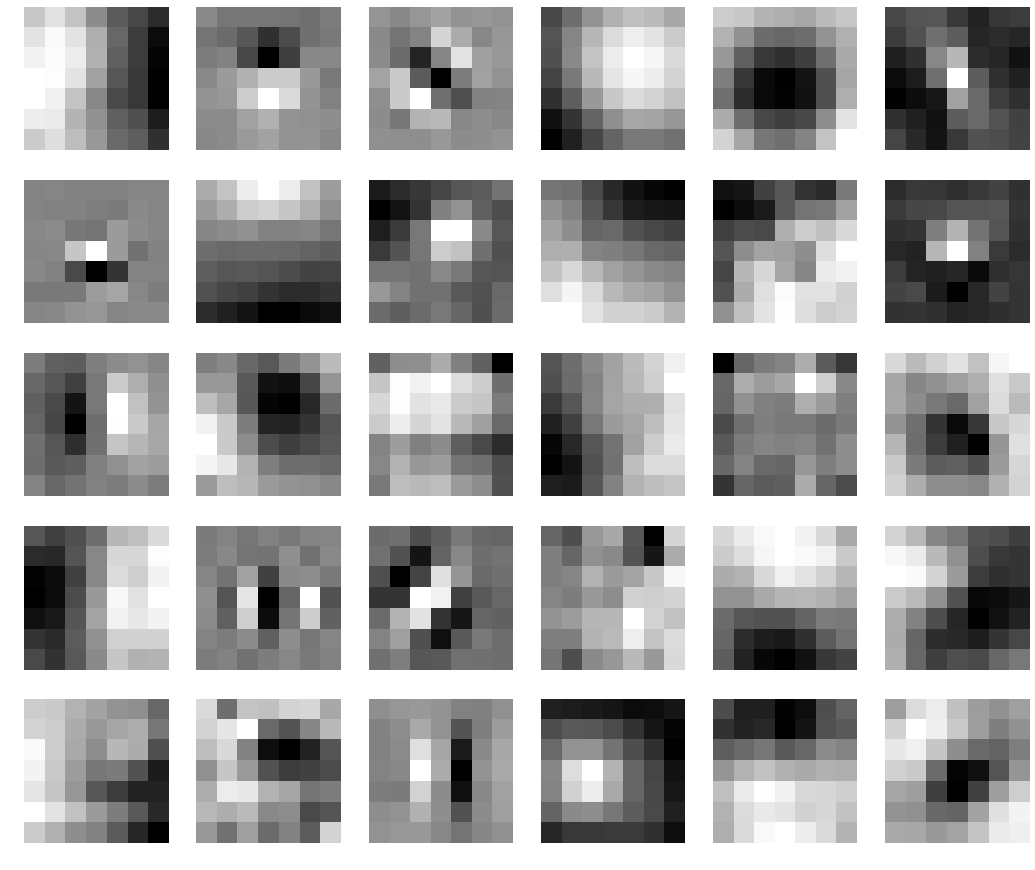}
        \caption{CBR Trans}
    \end{subfigure}
    \begin{subfigure}{.24\linewidth}
        \includegraphics[width=\linewidth]{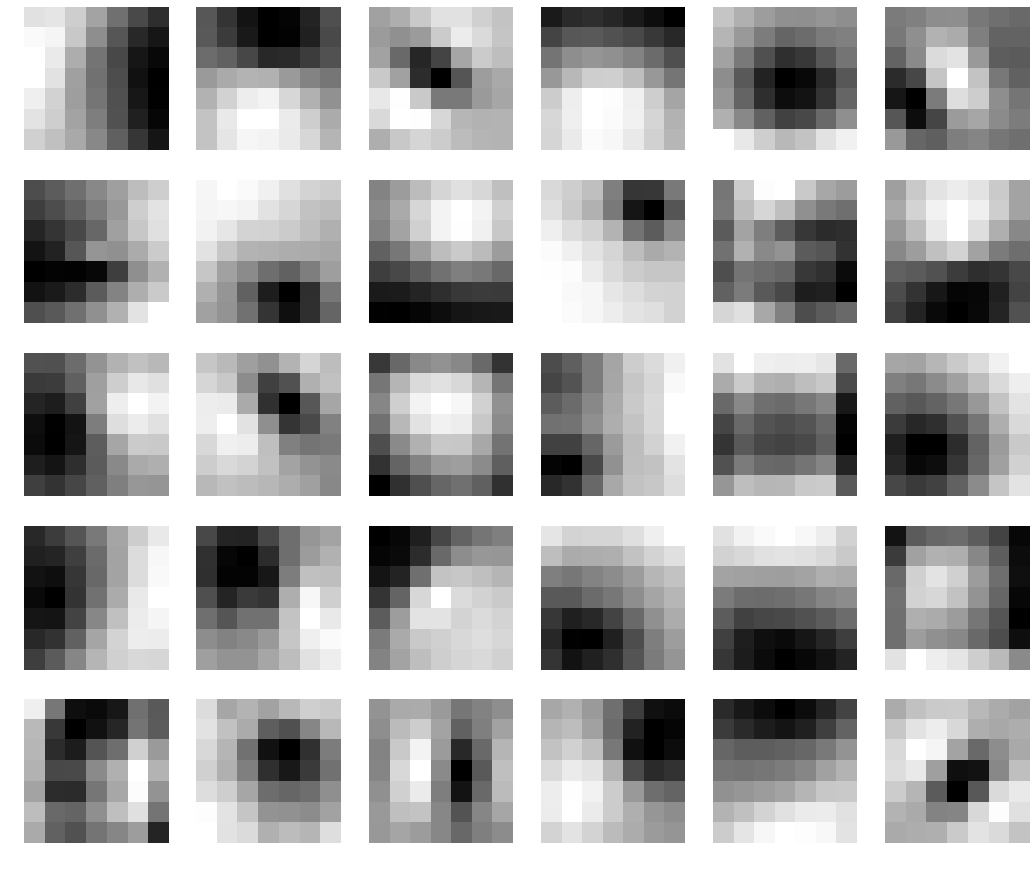}
        \caption{CBR Trans Final}
    \end{subfigure}
    \caption{\small\textbf{Visualization of conv1 filters shows the remains of initialization after training in Resnet, and the lack of and erasing of Gabor filters in CBR-Small.} We visualize the filters before and after training from random initialization and pretrained weights for Resnet (top row) and CBR-Small (bottom row). Comparing the similarity of (e) to (f) and (g) to (h) shows the limited movement of Resnet through training, while CBR-Small changes much more. We see that CBR does not learn Gabor filters when trained from scratch (f), and also erases some of the pretrained Gabors (compare (g) to (h).)}
    \label{fig:retina-weight-vis}
\end{figure}

\section{Convergence: Feature Independent Benefits and Weight Transfusion}
\label{sec-convergence}
In this section, we investigate the effects of transfer learning on convergence speed, finding that: (i) surprisingly, transfer offers \textit{feature independent} benefits to convergence simply through better weight scaling (ii) using pretrained weights from the lowest two layers/stages has the biggest effect on convergence --- further supporting the findings in the previous section that any meaningful feature reuse is concentrated in these lowest two layers (Figure \ref{fig:cca-before-after-finetuning}.) These results suggest some hybrid approaches to transfer learning, where only a subset of the pretrained weights (lowest layers) are used, with a lightweight redesign to the top of the network, and even using entirely \textit{synthetic} features, such as synthetic Gabor filters (Appendix~\ref{sec:app-syn-gabor}). We show these hybrid approaches capture most of the benefits of transfer and enable greater flexibility in its application.
\begin{figure}
\centering
\begin{tabular}{ccc}
\hspace*{-20mm} \includegraphics[width=0.4\columnwidth]{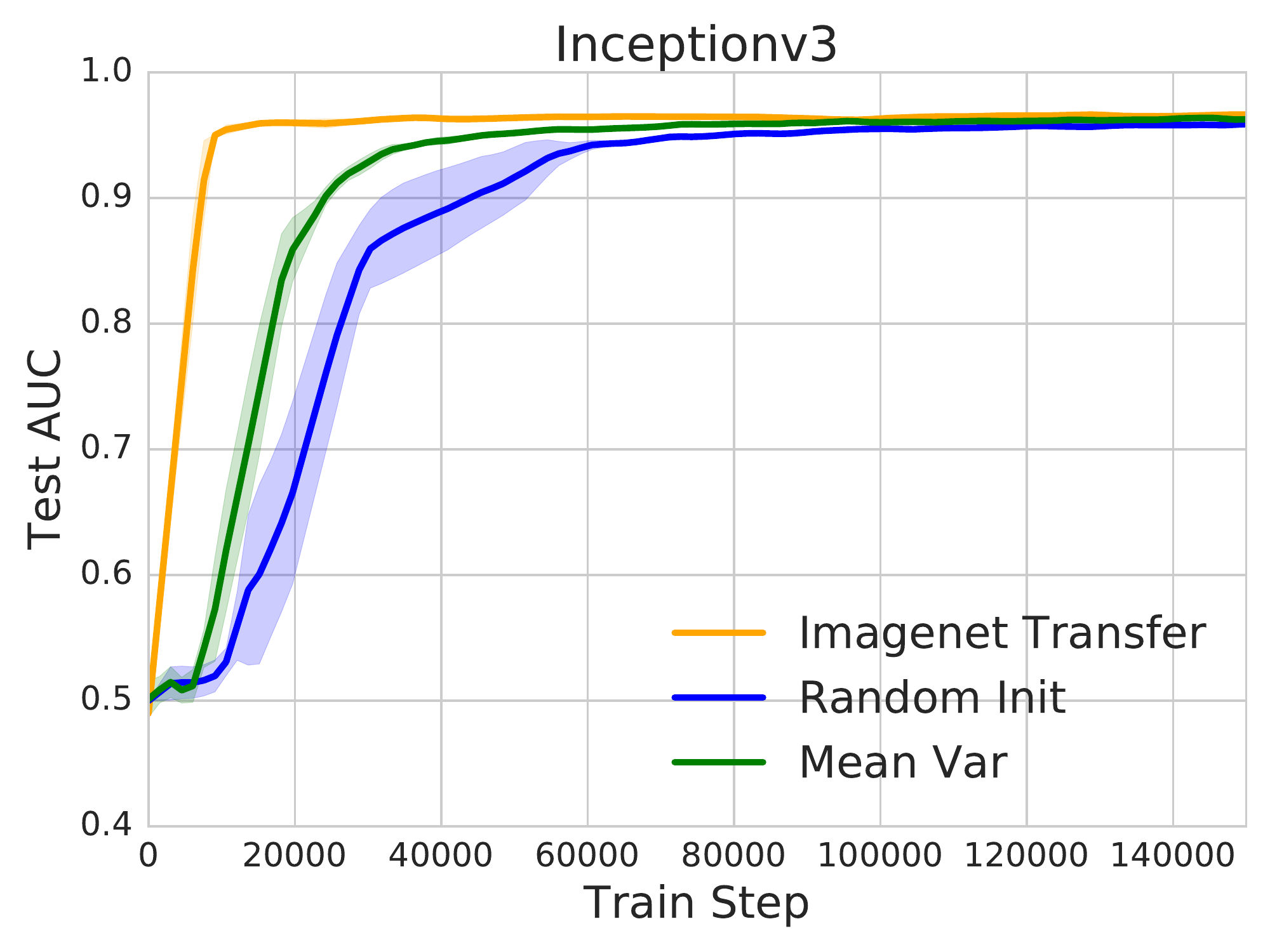} &
\hspace*{-5mm} \includegraphics[width=0.4\columnwidth]{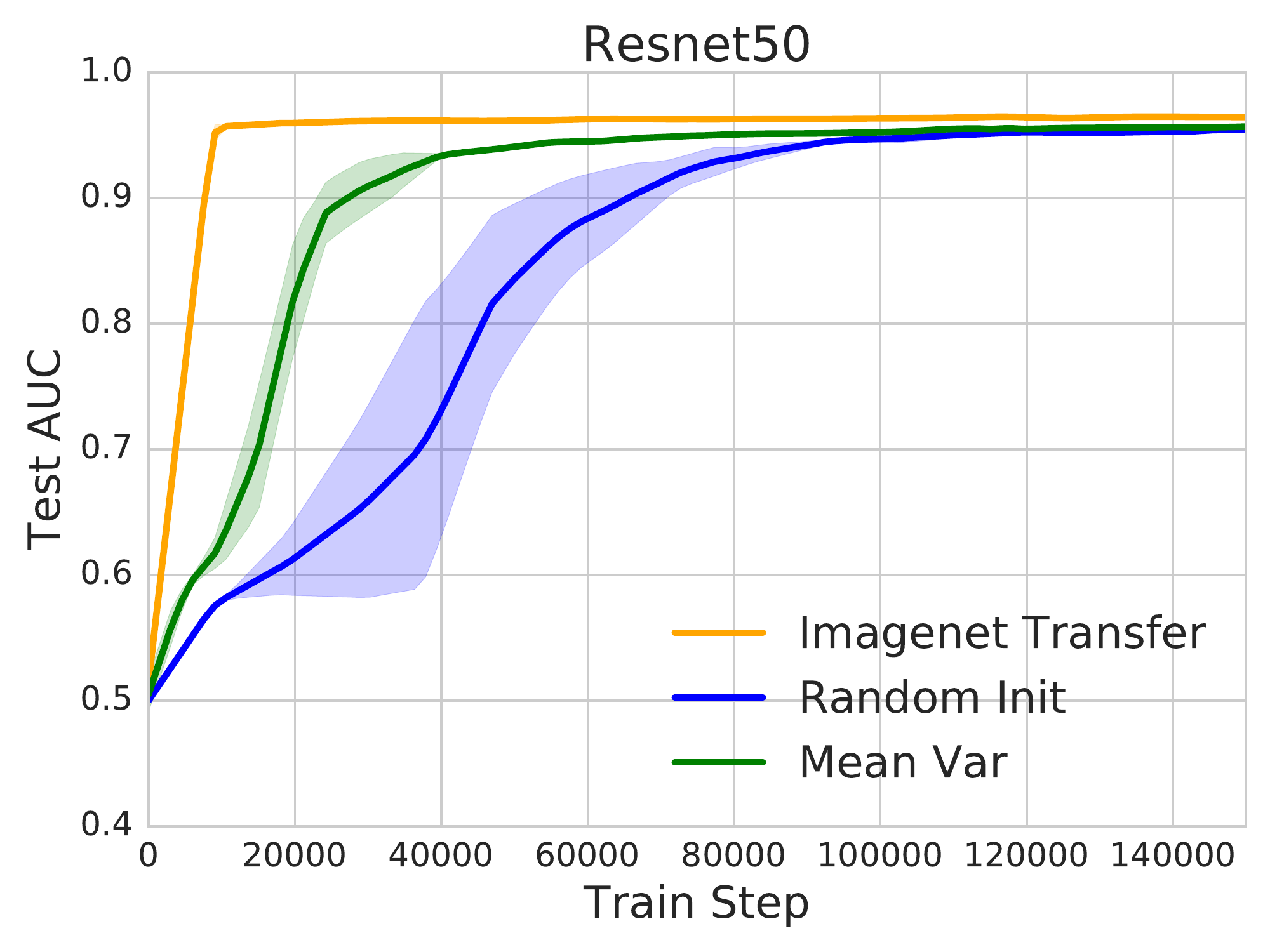}  &
\hspace*{-5mm} \includegraphics[width=0.44\columnwidth]{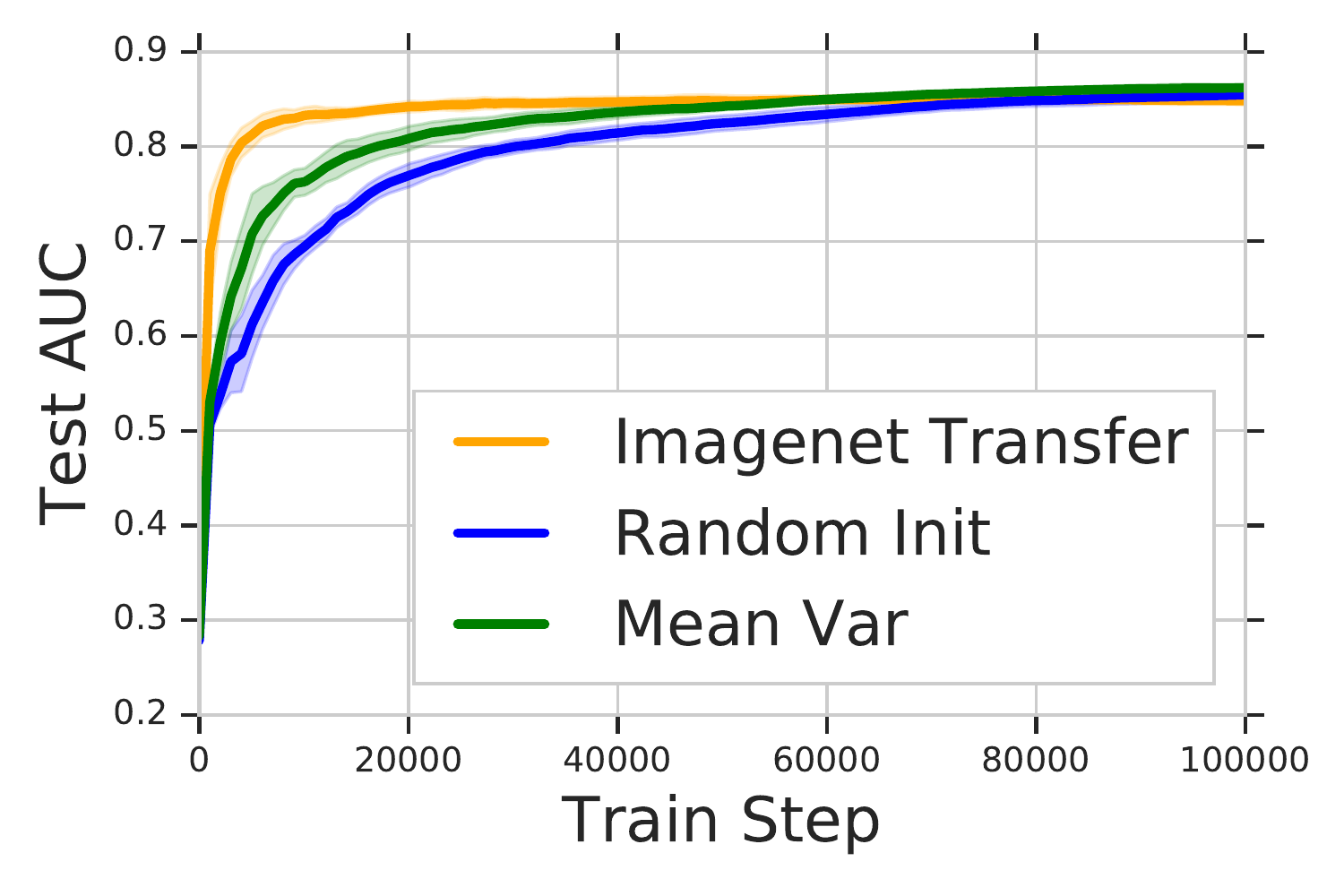}
\end{tabular}
\vspace*{-5mm}
\caption{\small \textbf{Using only the scaling of the pretrained weights (Mean Var Init) helps with convergence speed}.
    The figures compare the standard transfer learning and the \emph{Mean Var}
    initialization scheme to training from scratch. On both the \retina data (a-b) and the \chexpert data (c) (with Resnet50 on  the \emph{Consolidation} disease), we see convergence speedups.}
\label{fig:meanvar-in-body}
\end{figure}

\textbf{Feature Independent Benefits of Transfer: Weight Scalings}
We consistently observe that using pretrained weights results in faster convergence. One explanation for this speedup is that there is significant feature reuse. However, the results of Section \ref{sec-analysis} illustrate that there are many confounding factors, such as model size, and feature reuse is likely limited to the lowest layers. We thus tested to see whether there were \textit{feature independent} benefits of the pretrained weights, such as \textit{better scaling}. In particular, we initialized a \textit{iid weights} from $\mathcal{N}(\tilde{\mu}, \tilde{\sigma}^2)$, where $\tilde{\mu}$ and $\tilde{\sigma}^2$ are the mean and variance of $\tilde{W}$, the pretrained weights. Doing this for each layer separately inherits the scaling of the pretrained weights, but destroys all of the features. We called this the \textit{Mean Var} init, and found that it significantly helps speed up convergence (Figure \ref{fig:meanvar-in-body}.) Several additional experiments studying batch normalization, weight sampling, etc are in the Appendix.
\begin{figure}
\centering
\begin{tabular}{cc}
\hspace*{-10mm} \includegraphics[width=0.51\columnwidth]{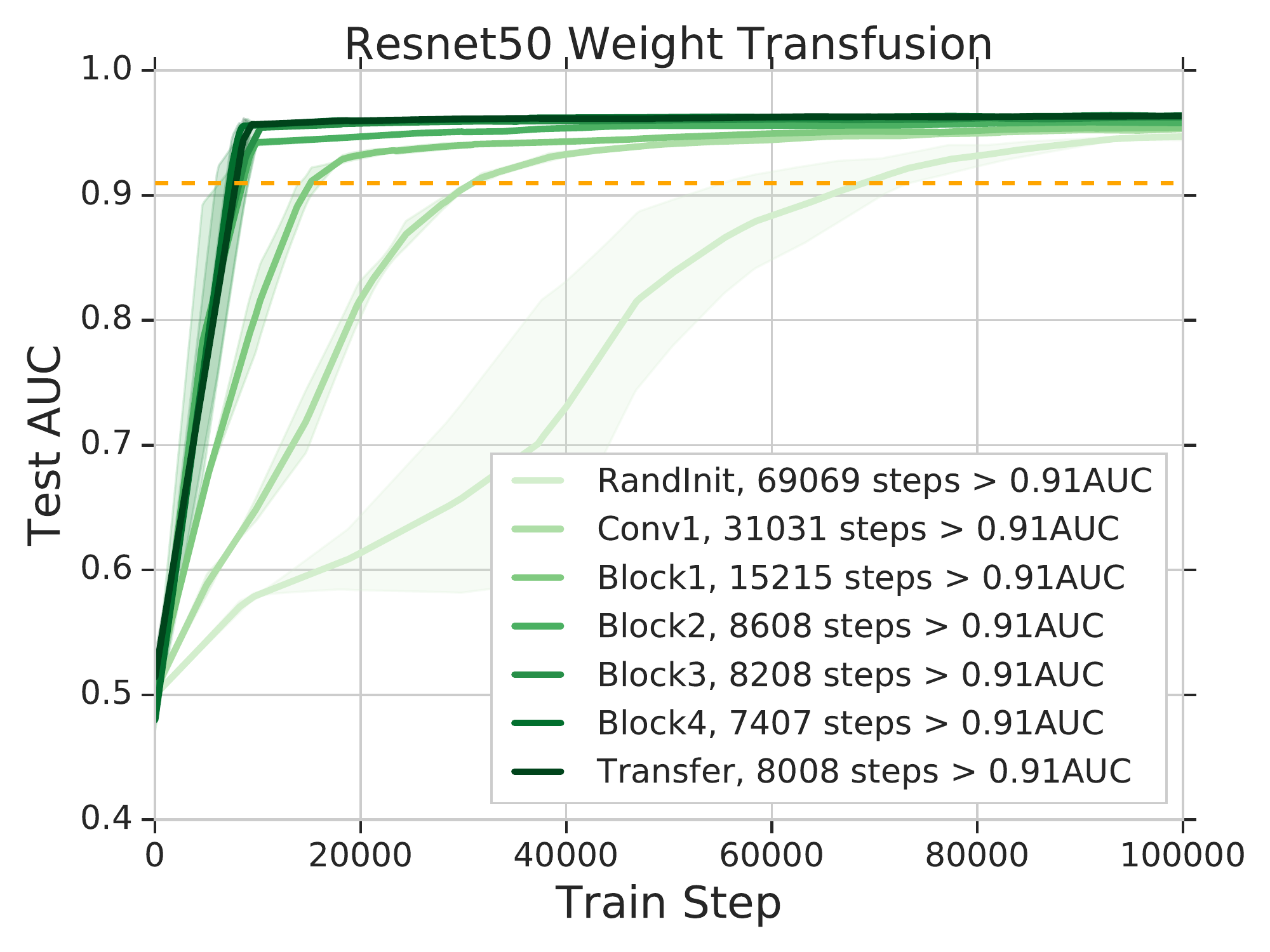} &
 \includegraphics[width=0.51\columnwidth]{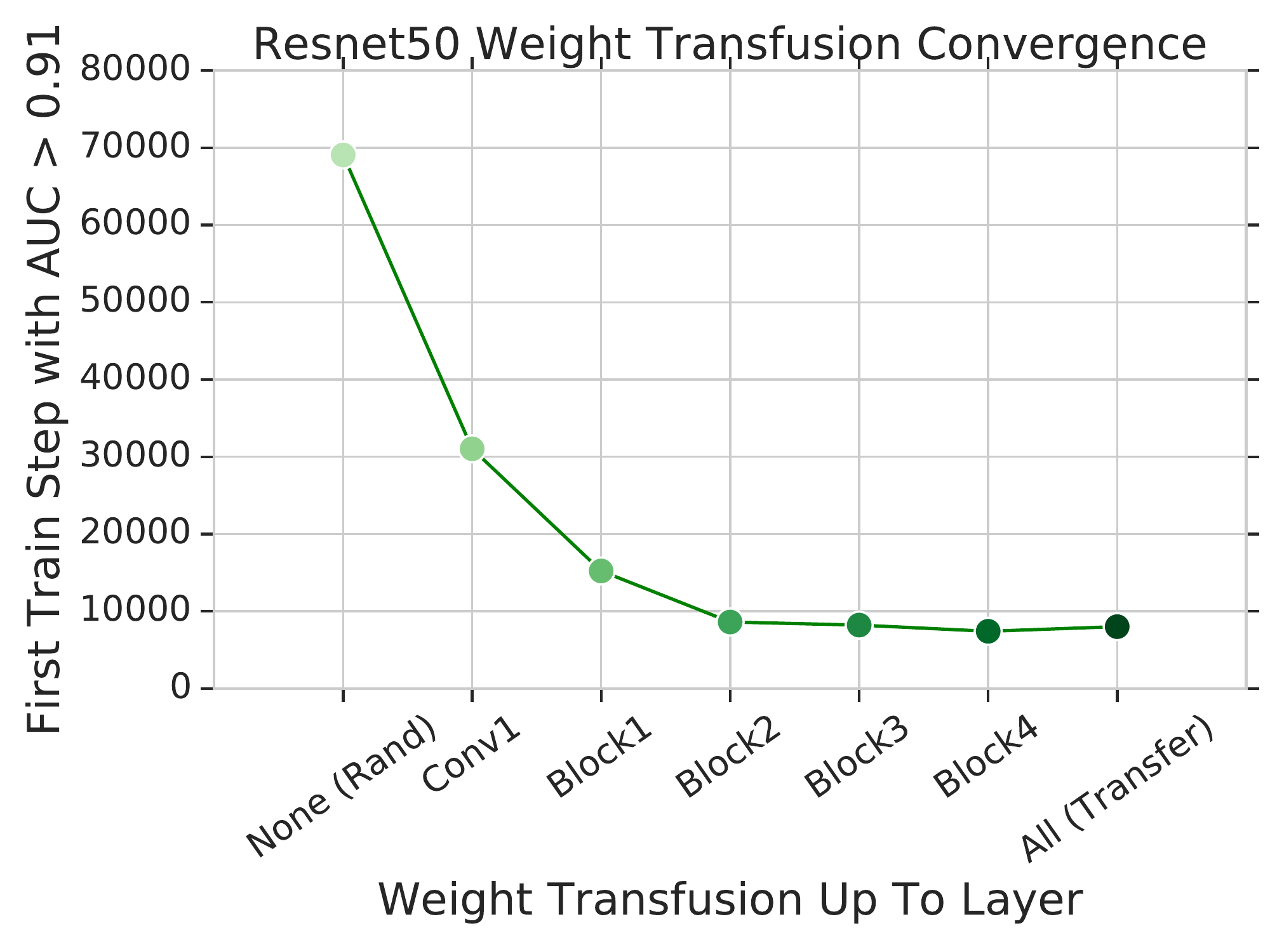}
\end{tabular}
\caption{\small \textbf{Reusing a subset of the pretrained weights (weight transfusion), further supports only the lowest couple of layers performing meaningful feature reuse.} We initialize a Resnet with a contiguous subset of the layers using pretrained weights (weight transfusion), and the rest randomly, and train on the \retina task. On the left, we show the convergene plots when transfusing up to conv1 (just one layer), up to block 1 (conv1 and all the layers in block1), etc up to full transfer. On the right, we plot the number of train steps taken to reach $91\%$ AUC for different numbers of transfused weights. Consistent with findings in Section \ref{sec-analysis}, we observe that reusing the lowest layers leads to the greatest gain in convergence speed. Perhaps surprisingly, just reusing conv1 gives the greatest marginal convergence speedup, even though transfusing weights for a block means several new layers are using pretrained weights.}
\label{fig:transfusion-new}
\end{figure}

\begin{figure}
\centering
\begin{tabular}{cc}
\hspace*{-12mm} \includegraphics[width=0.55\columnwidth]{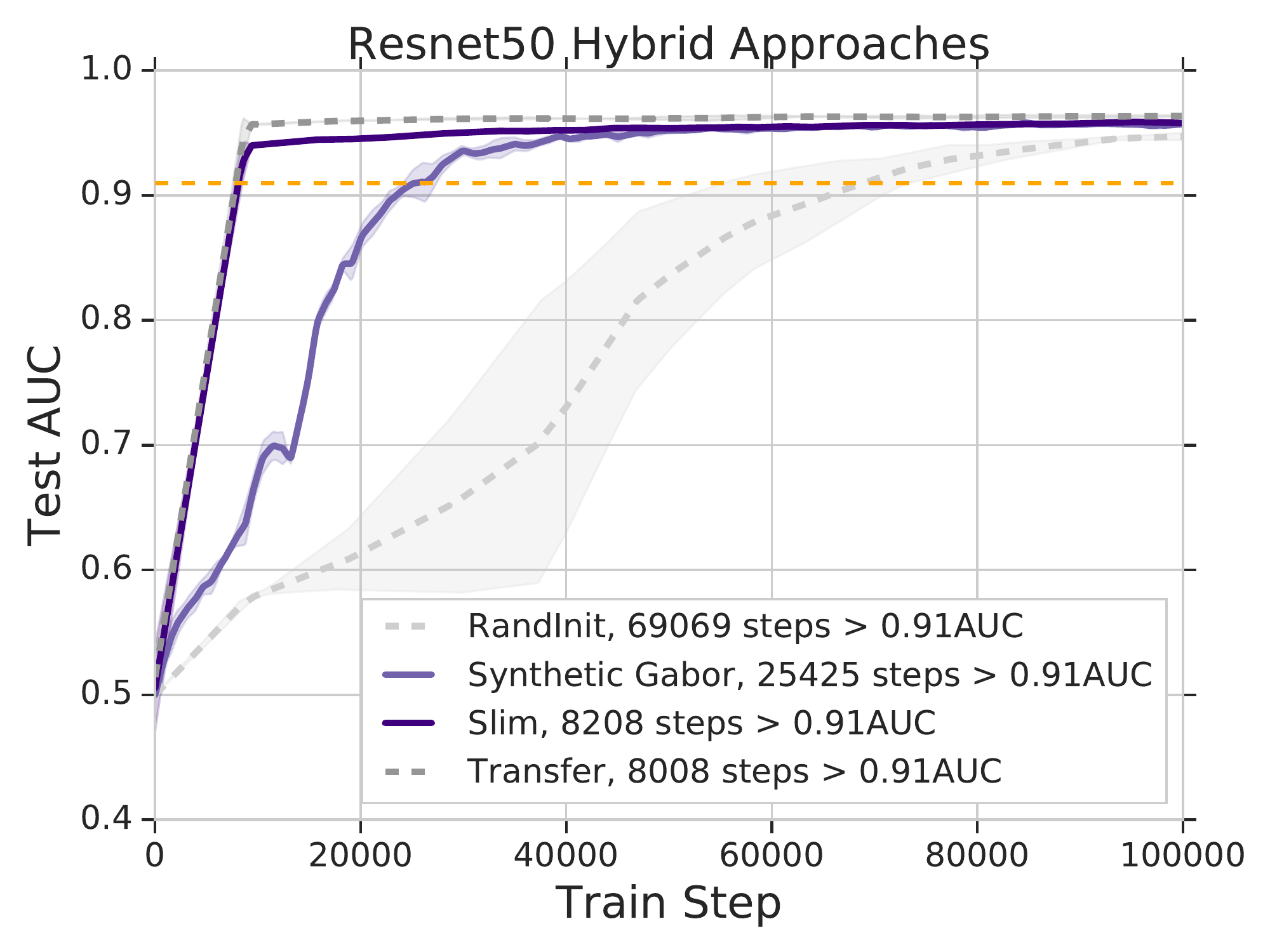} &
\hspace*{-5mm} \includegraphics[width=0.55\columnwidth]{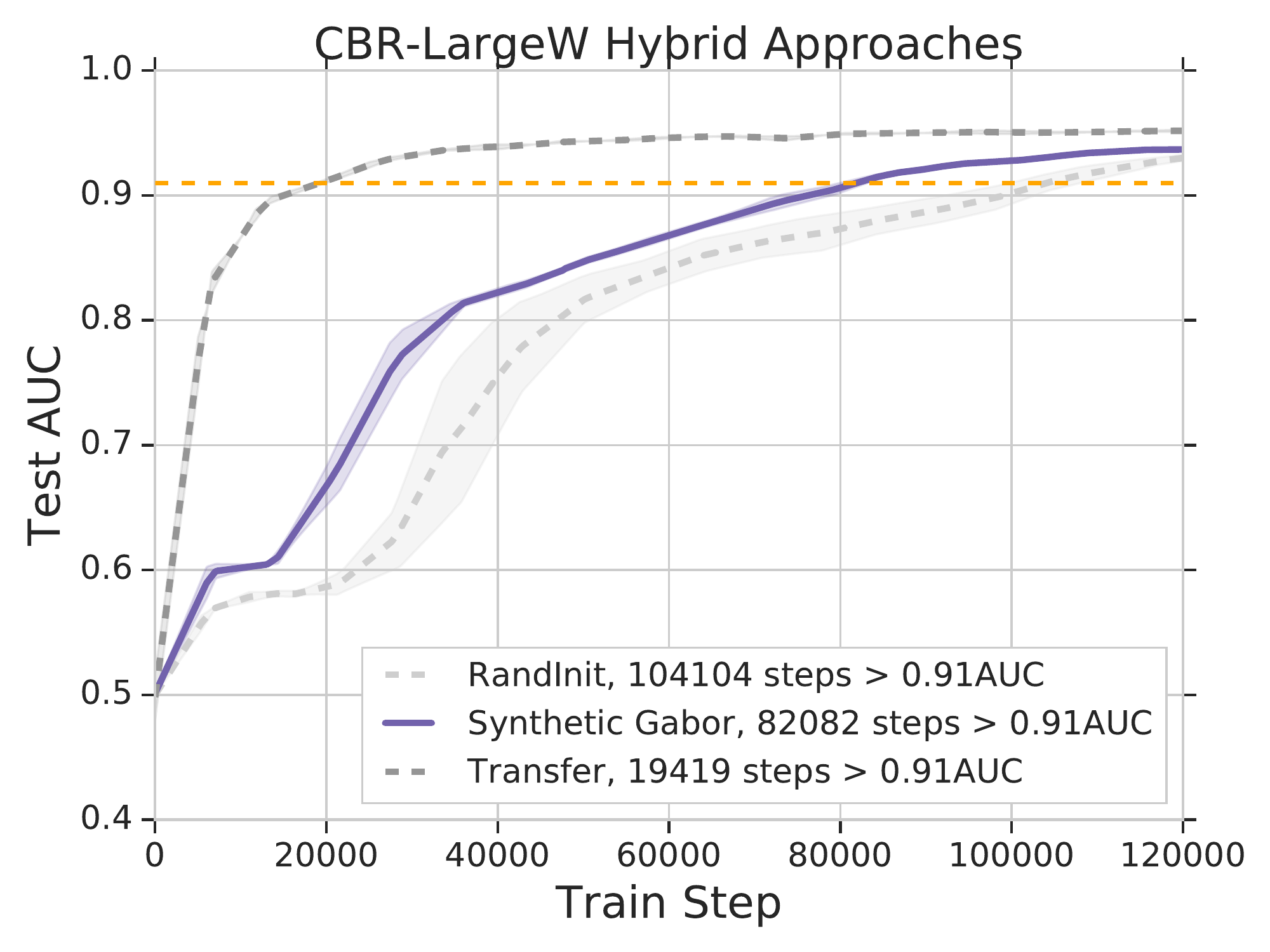}
\end{tabular}
\caption{\small \textbf{Hybrid approaches to transfer learning: reusing a subset of the weights and slimming the remainder of the network, and using synthetic Gabors for conv1.} For Resnet, we look at the effect of reusing pretrained weights up to Block2, and slimming the remainder of the network (halving the number of channels), randomly initializing those layers, and training end to end. This matches performance and convergence of full transfer learning. We also look at initializing conv1 with \textit{synthetic Gabor filters} (so \textit{no} use of pretrained weights), and the rest of the network randomly, which performs equivalently to reusing conv1 pretrained weights. This result generalizes to different architectures, e.g. CBR-LargeW on the right.}
\label{fig:transfer-hybrids}
\end{figure}
\textbf{Weight Transfusions and Feature Reuse}
We next study whether the results suggested by Section \ref{sec-analysis}, that meaningful feature reuse is restricted to the lowest two layers/stages of the network is supported by the effect on convergence speed. We do this via a \textit{weight transfusion} experiment, transfering a contiguous set of some of the pretrained weights, randomly initializing the rest of the network, and training on the medical task. Plotting the training curves and steps taken to reach a threshold AUC in Figure \ref{fig:transfusion-new} indeed shows that using pretrained weights for lowest few layers has the biggest training speedup. Interestingly, just using pretrained weights for conv1 for Resnet results in the largest gain, despite transfusion for a Resnet block meaning multiple layers are now reusing pretrained weights.

\textbf{Takeaways: Hybrid Approaches to Transfer Learning} The transfusion results suggest some hybrid, more flexible approaches to transfer learning. Firstly, for larger models such as Resnet, we could consider reusing pretrained weights up to e.g. Block2, redesiging the top of the network (which has the bulk of the parameters) to be more lightweight, initializing these layers randomly, and training this new Slim model end to end. Seeing the disproportionate importance of conv1, we might also look at the effect of initializing conv1 with \textit{synthetic Gabor filters} (see Appendix~\ref{sec:app-syn-gabor} for details) and the rest of the network randomly. In Figure \ref{fig:transfer-hybrids} we illustrate these hybrid approaches. Slimming the top of the network in this way offers the same convergence and performance as transfer learning, and using synthetic Gabors for conv1 has \textit{the same effect} as pretrained weights for conv1. These variants highlight many new, rich and flexible ways to use transfer learning.

\section{Conclusion}
In this paper, we have investigated many central questions on transfer learning for medical imaging applications. Having benchmarked both standard \imagenet architectures and non-standard lightweight models (itself an underexplored question) on two large scale medical tasks, we find that transfer learning offers limited performance gains and much smaller architectures can perform comparably to the standard \imagenet models. Our exploration of representational similarity and feature reuse reveals surprising correlations between similarities at initialization and after training for standard \imagenet models, providing evidence of their overparametrization for the task. We also find that meaningful feature reuse is concentrated at the lowest layers and explore more flexible, hybrid approaches to transfer suggested by these results, finding that such approaches maintain all the benefits of transfer and open up rich new possibilities. We also demonstrate feature-independent benefits of transfer learning for better weight scaling and convergence speedups.

\bibliography{refs}

\appendix\newpage\clearpage
{\Large\textbf{Appendix to ``\transfusiontitle''}}

\section{Details on Datasets, Models and Hyperparameters}
The \retina dataset consisted of around 250k training images, and 70k test images. The train test split was done by patient id (as is standard for medical datasets) to ensure no accidental similarity between the train/test dataset. The chest x-ray dataset is open sourced and available from \citep{irvin2019chexpert}, which has all of the details. Briefly, they have 223k training images and binary indicator for multiple diseases assiciated with each image extracted automatically from the meta data. The standard \imagenet (ILSVRC 2012) dataset was also used to pretrain models.

For dataset preprocessing we used mild random cropping, as well as standard normalization by the mean and standard deviation for \imagenet. We augmented the data with hue and contrast augmentations. For the \retina data, we used random horizontal and vertical flips, and for the chest x-ray data, we did not do random flip. We did not do model specific hyperparameter tuning on each target data, and used fixed standard hyperparameters.

For experiments on the \retina data, we trained the standard \imagenet models, Resnet50 and Inception-v3, by replacing the final 1000 class \imagenet classification head with a five class head for DR diagnosis, or five classes for the five different chest x-ray diseases. We use the sigmoid activation at the top instead of the multiclass softmax activation, and the train the models in the multi-label binary classification framework.

The CBR family of small convolutional neural networks consists of multiple conv2d-batchnorm-relu layers followed by a maxpool. Each maxpool has spatial window (3x3) and stride (2x2). For each CBR architecture, there is one filter size for all the convolutions (which all have stride 1). Below, conv-n denotes a 2d convolutionl with n output channels.
\begin{itemize}
\item \textbf{CBR-LargeT}(all) has 7x7 conv filters: (conv32-bn-relu) maxpool (conv64-bn-relu) maxpool (conv128-bn-relu) maxpool (conv256-bn-relu) maxpool (conv512-bn-relu) global avgpool, classification
\item \textbf{CBR-LargeW}(ide) has 7x7 conv filters: (conv64-bn-relu) maxpool (conv128-bn-relu) maxpool (conv256-bn-relu) maxpool (conv512-bn-relu) maxpool, global avgpool, classification.
\item \textbf{CBR-Small} has 7x7 conv filters: (conv32-bn-relu) maxpool (conv64-bn-relu) maxpool (conv128-bn-relu) maxpool (conv256-bn-relu) maxpool global avgpool, classification
\item \textbf{CBR-Tiny} has 5x5 conv filters: (conv64-bn-relu) maxpool (conv128-bn-relu) maxpool (conv256-bn-relu) maxpool (conv512-bn-relu) maxpool, global avgpool, classification.
\end{itemize}

The models on \retina are trained on $587\times 587$ images, with learning rate 0.001 and a batch size of 8 (for memory considerations.) The Adam optimizer is used. The models on the chest x-ray are trained on $224\times 224$ images, with a batch size of 32, and vanilla SGD with momentum (coefficient 0.9). The learning rate scheduling is inherited from the \imagenet training pipeline, which warms up from 0 to $0.1\times \frac{32}{256}$ in 5 epochs, and then decay with a factor of 10 on epoch 30, 60, and 90, respectively.
\section{Additional Dataset Size Results}
Complementing the data varying experiments in the main text, we additional experiments on varying the amount of training data, fidning that for around $50$k datapoints, we return to only seeing a fractional improvement of transfer learning. Future work could study how hybrid approaches perform when less data is available.
\begin{table*}[t]
  \centering
   \begin{tabular}{llllll}
    \toprule            \textbf{Model} & \textbf{Init Method} & \textbf{5k} & \textbf{10k} & \textbf{50k} & \textbf{100k} \\
    \midrule
   Resnet50 & \imagenet Pretrained & $94.6 \%$ & $94.8\%$  & $95.7\%$ & $96.0$   \\
   Resnet50 &  Random Init & $92.2 \%$ & $93.3 \%$  & $95.3 \%$ & $95.9 \%$ \\
   CBR-LargeT & Random Init & $93.6 \%$ & - & - & - \\
   CBR-LargeT & Pretrained & $93.9 \%$ & - & - & - \\
   CBR-LargeW & Random Init & $93.6 \%$ & - & - & - \\
   CBR-LargeW & Pretrained & $93.7 \%$ & - & - & - \\
    \midrule
   Resnet50 & Conv1 Pretrained & $92.9 \%$ & - & - & - \\
    \midrule
    Resnet50 & Mean Var Init & - & $94.4\%$  & $95.5\%$ & $95.8\%$ \\
    \bottomrule
  \end{tabular}
  \caption{\small \textbf{Additional performance results when varying initialization and the dataset size on the \retina task.} For Resnet50, we show performances when training on very small amounts of data. We see that even finetuning (with early stopping) on 5k datapoints beats the results from performing fixed feature extraction, Figure \ref{fig:freezing-results}, suggesting finetuning should always be preferred. For 5k, 10k datapoints, we see a larger gap between transfer learning and random init (closed by 50k datapoints) but this is likely due to the enormous size of the model (typically trained on 1 million datapoints) compared to the dataset size. This is supported by evaluating the effect of transfer on CBR-LargeT and CBR-LargeW, where transfer again does not help much. (These are one third the size of Resnet50, and we expect the gains of transfer to be even more minimal for CBR-Small and CBR-Tiny.) We also show results for using the MeanVar init, and see some gains in performance for the very small data setting. We also see a small gain on 5k datapoints when just reusing the conv1 weights for Resnet50.}
  \label{table-vary-data}
\end{table*}

\section{CCA Details}
\label{sec-app-cca-details}
For full details on the implementation of CCA, we reference prior work \citep{raghu2017svcca, morcos2018insights}, as well as the open sourced code (the source of our implementation): https://github.com/google/svcca

One challenge we face when implementing CCA is the large size of the convolutional activations. These activations have shape $(n, h, w, c)$, where $n$ is the number of datapoints, $c$ the number of channels, and $h, w$ the spatial dimensions. These values all vary significantly across the network, e.g. conv1 has shape $(n, 294, 294, 64)$, while activations at the end of block 3 have shape $(n, 19, 19, 1024)$. Because CCA is sensitive to both the number of datapoints $n$ (actually $hwn$ for convolutional layers) and the number of neurons -- $c$ for large convolutional layers -- there is large variations in scaling across different layers in the model. To address this, we do the following: let $L$ and $L'$ be the layers we want to compare, with shape (height, width, channels),  $(h_L, w_L, c_L)$. We apply CCA as follows:
\begin{itemize}
\item Pick $p$, the total number of image patches to compute activation vectors and CCA over, and $d$, the maximum number of neuron activation vectors to correlate with
\item Pick the number of datapoints $n$ so that $nh_Lw_L = p$.
\item Sample $d$ of the $c_L$ channels, and apply CCA to the resulting $d$ x $nh_Lw_L$ activation matrices.
\item Repeat over samples of $d$ and $n$.
\end{itemize}
This works much better than prior approaches of averaging over all of the spatial dimensions \citep{morcos2018insights}, or flattening across all of the neurons \citep{raghu2017svcca} (too computationally expensive in this setting.)

\section{Additional Results from Representation Analysis}

\begin{table*}[h]
  \centering
   \begin{tabular}{llllll}
    \toprule
   \textbf{Description} & \textbf{Conv1} & \textbf{Block1} & \textbf{Block2} & \textbf{Block3} &  \textbf{Block4}   \\
    \midrule
   Resnet50 CCA(ImNet1, ImNet2) & $0.865$ & $0.559$ & $0.421$ & $0.343$ & $0.313$   \\
   Resnet50 CCA(Rand1, Rand2) & $0.647$ & $0.369$ & $0.277$ & $0.256$ & $0.276$   \\
   Resnet50 Diff & $\textbf{0.218}$ & $\textbf{0.191}$ & $\textbf{0.144}$ & $\textbf{0.086}$ & $\textbf{0.037}$   \\
   \midrule
  \end{tabular}
     \begin{tabular}{lllll}
    \toprule
   \textbf{Description} & \textbf{Pool1} & \textbf{Pool2} & \textbf{Pool3} & \textbf{Pool4} \\
    \midrule
   CBR-Small CCA(ImNet1, ImNet2) & $0.825$ & $0.709$ & $0.477$ & $0.395$ \\
   CBR-Small CCA(Rand1, Rand2) & $0.723$ & $0.541$ & $0.401$ & $0.349$ \\
   CBR-Small Diff & $\textbf{0.102}$ & $\textbf{0.168}$ & $\textbf{0.076}$ & $\textbf{0.046}$   \\
   \bottomrule
  \end{tabular}
  \caption{\small \textbf{Representational comparisons between trained ImageNet models with different seeds highlight the variation of behavior in higher and lower layers, and differences between larger and smaller models.} We compute CCA similarity between representations at different layers when training from different random seeds with (i) (the same) pretrained weights (ii) different random inits, for Resnet and CBR-Small. The results support the conclusions of the main text. For Resnet50, in the lowest layers such as Conv1 and Block1, we see that representations learned when using (the same) pretrained weights are much more similar to each other (diff $0.2$ in CCA score) than representations learned from different random initializations. This $\sim 0.2$ difference is also much higher than (somewhat) corresponding differences in CBR-Small, for Pool1, Pool2. Actually, as Resnet50 is much deeper, the large difference in Block1 is very striking. (Block 1 alone contains much more layers than all of CBR-Small.) By Block3 and Block4 however, the CCA similarity difference between pretrained representations and those from random initialization is much smaller, and slightly lower than the differences for Pool3, Pool4 in CBR-Small, suggesting that pretrained weights are not having much of a difference on the kinds of functions learned. For CBR-Small, we also see that pretrained weights result in larger differences between the representations in the lower layers, but these become much smaller in the higher layers. We also observe that representations in CBR-Small trained from random initialization (especially in the lower layers e.g. Pool1) are more similar to each other than in Resnet50, suggesting things move more.}
  \label{table-imnet-imnet}
\end{table*}
Here, we include some additional results studying the representations of these models.
We perform more representational similarity comparisons between networks trained from (the same) pretrained weights (as is standard), but different random seeds. We do this for Resnet50 (a large model) and CBR-Small (a small model), and Table \ref{table-imnet-imnet} includes these results as well as similarity comparisons for networks trained with different random seeds and \textit{different} random initializations as a baseline. The comparisons across layers and models is slightly involved, but as we detail below, the evidence further supports the conclusions in the main text:
\begin{itemize}
\item \textit{Larger models change less through training.} Comparing CCA similarity scores across models is a little challenging, due to different scalings, so we look at the difference in CCA similarity between two models trained with pretrained weights, and two models trained from random initialization, for Resnet50 and CBR-Small. Comparing this value for Conv1 (in Resnet50) to Pool1 (in CBR-Small), we see that pretraining results in much more similar representations compared to random initialization in the large model over the small model.
\item \textit{The effect of pretraining is mostly limited to the lowest layers} For higher layers, the CCA similarities between representations using pretrained weights and those trained from random initializations are closer, and the difference between CBR-Small and Resnet-50 is non-existent, suggesting that the effects of pretraining mostly affect the lowest layers across models, with finetuning changing representations at the top.
\end{itemize}

\begin{figure}
    \centering
    \begin{subfigure}{.24\linewidth}
        \includegraphics[width=\linewidth]{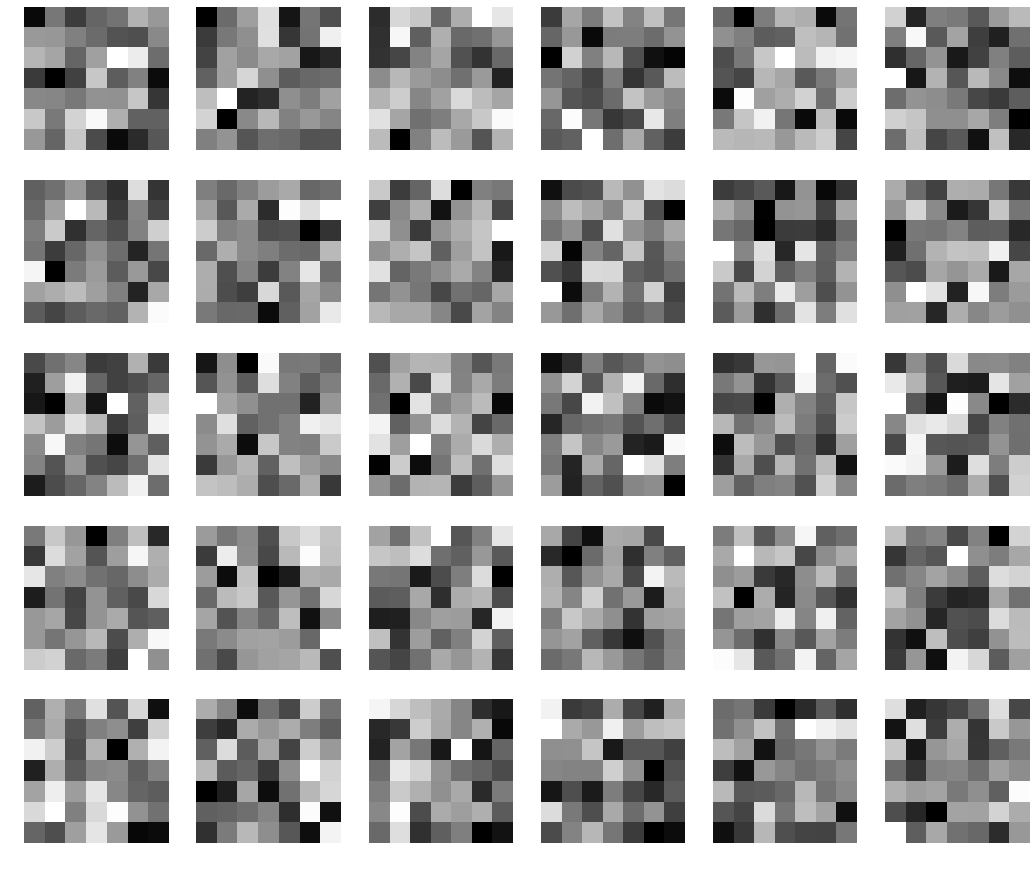}
        \caption{rand init}
    \end{subfigure}
    \begin{subfigure}{.24\linewidth}
        \includegraphics[width=\linewidth]{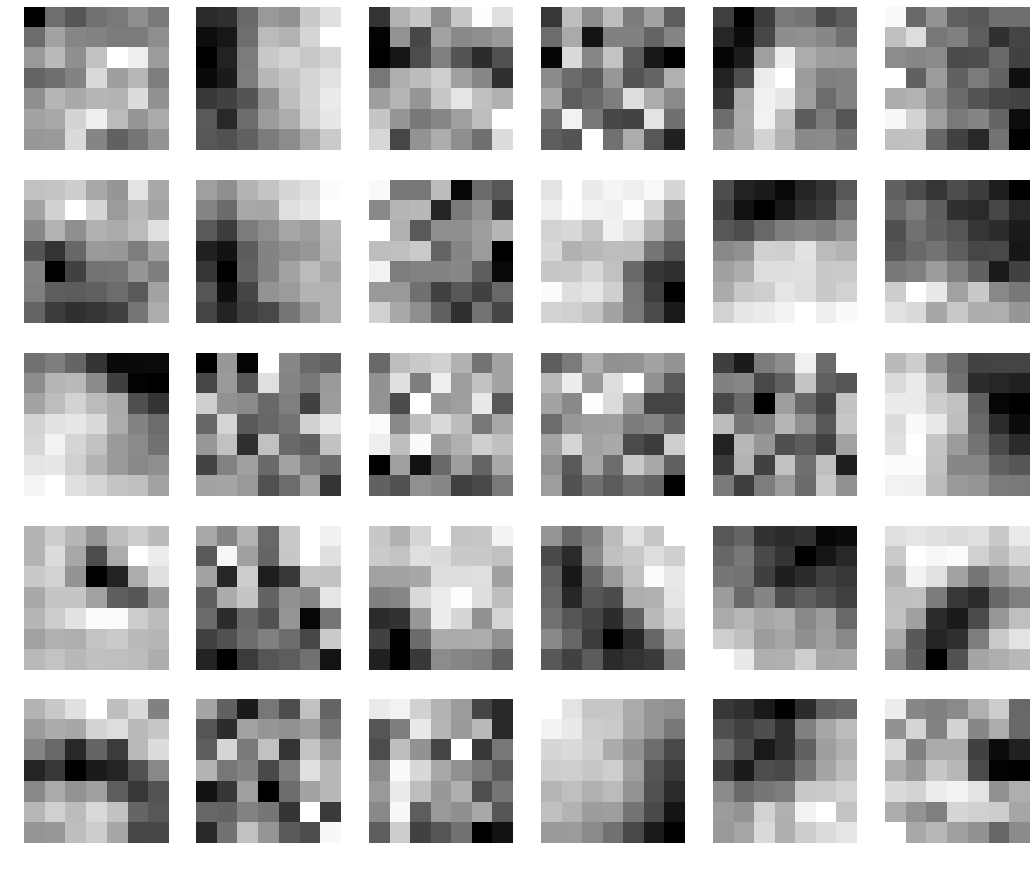}
        \caption{final (rand init)}
    \end{subfigure}
    \begin{subfigure}{.24\linewidth}
        \includegraphics[width=\linewidth]{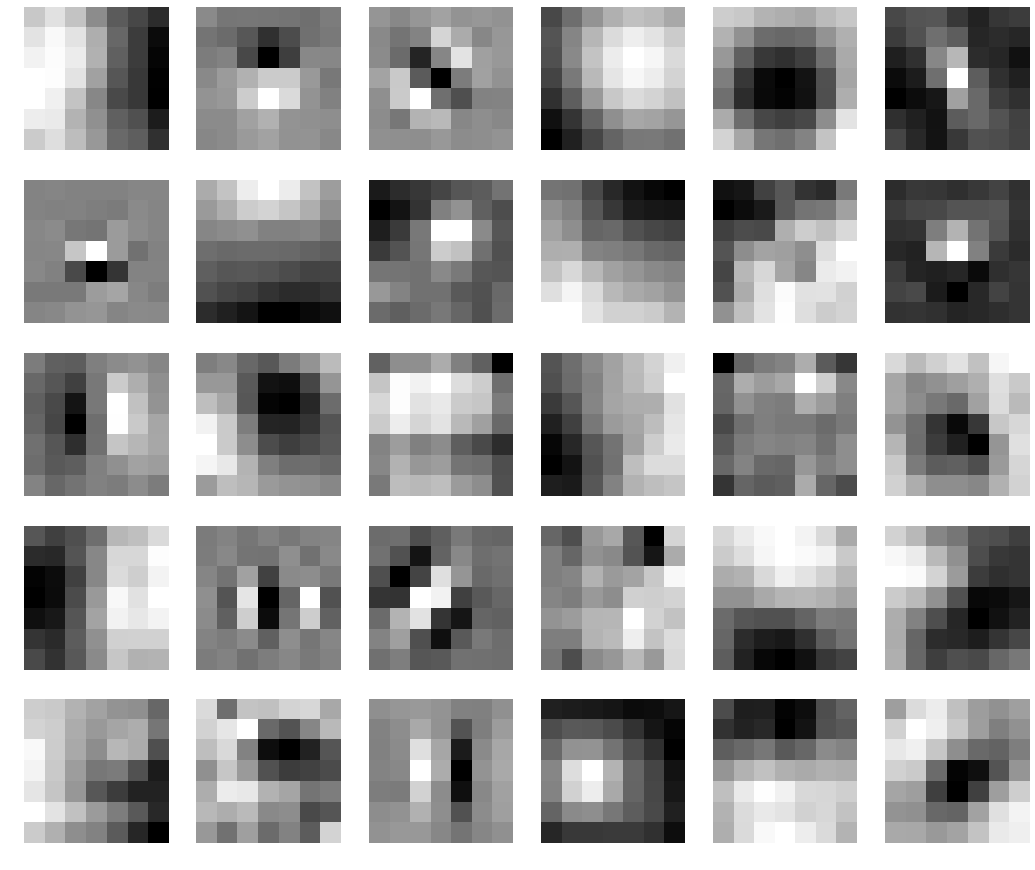}
        \caption{transfer init}
    \end{subfigure}
    \begin{subfigure}{.24\linewidth}
        \includegraphics[width=\linewidth]{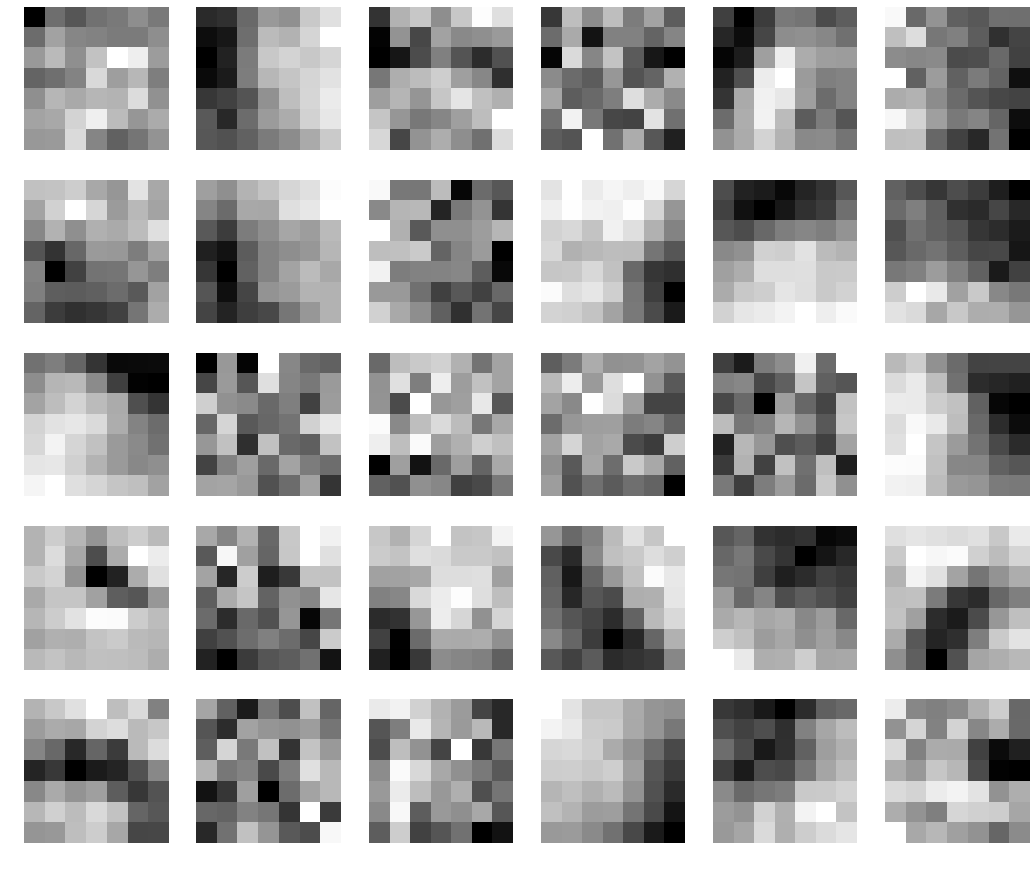}
        \caption{final (transfer init)}
    \end{subfigure}
    \caption{\small\textbf{First layer filters of CBR-Small on the \chexpert data.} (a) and (c) show
    the randomly initialized filters and filters initialized from a model (the same architecture)
    pre-trained on \imagenet. (b) and (d) shows the final converged filters from the two different initializations,
    respectively.}
    \label{fig:chexpert-vcsh-weight-vis}
\end{figure}

\begin{figure}
    \centering
    \begin{subfigure}{.24\linewidth}
        \includegraphics[width=\linewidth]{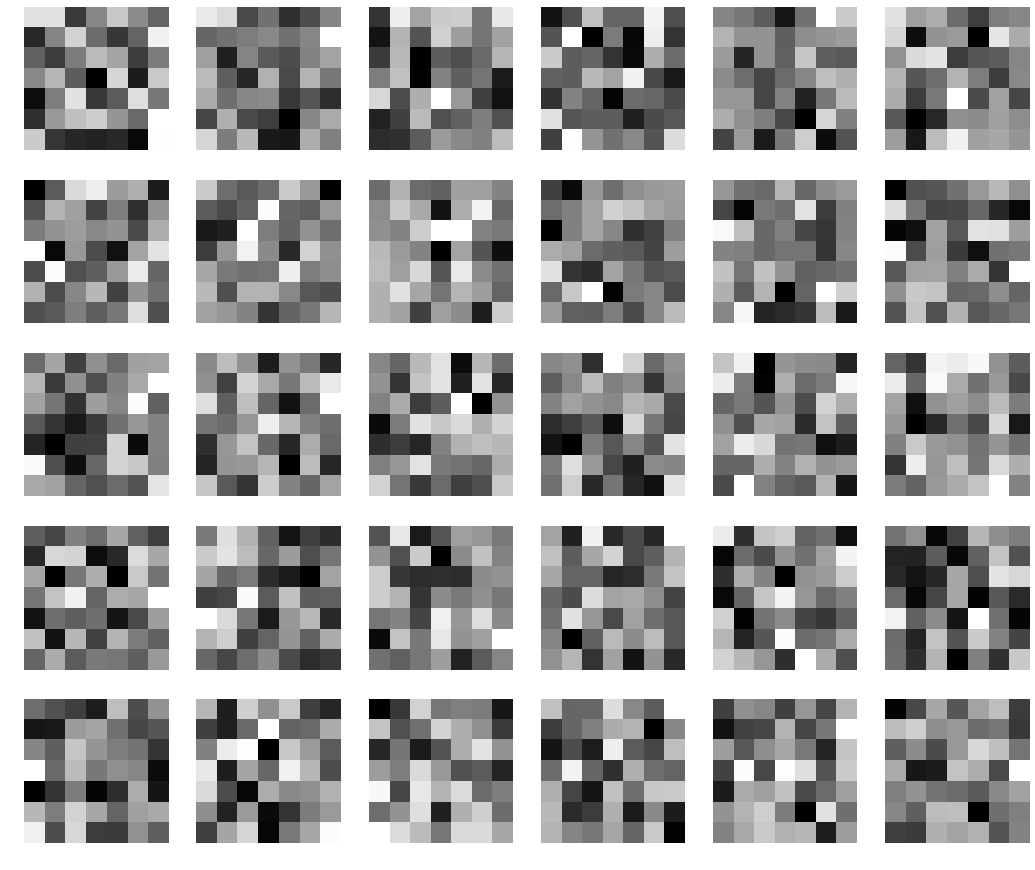}
        \caption{rand init}
    \end{subfigure}
    \begin{subfigure}{.24\linewidth}
        \includegraphics[width=\linewidth]{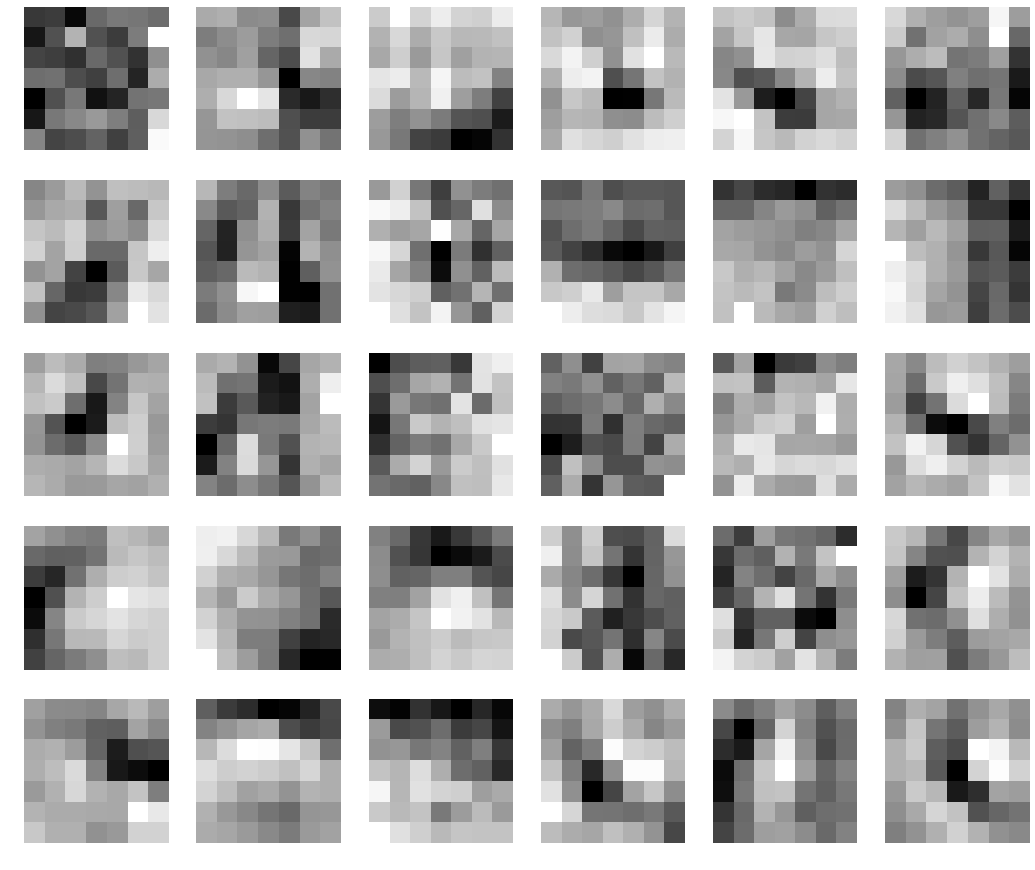}
        \caption{final (rand init)}
    \end{subfigure}
    \begin{subfigure}{.24\linewidth}
        \includegraphics[width=\linewidth]{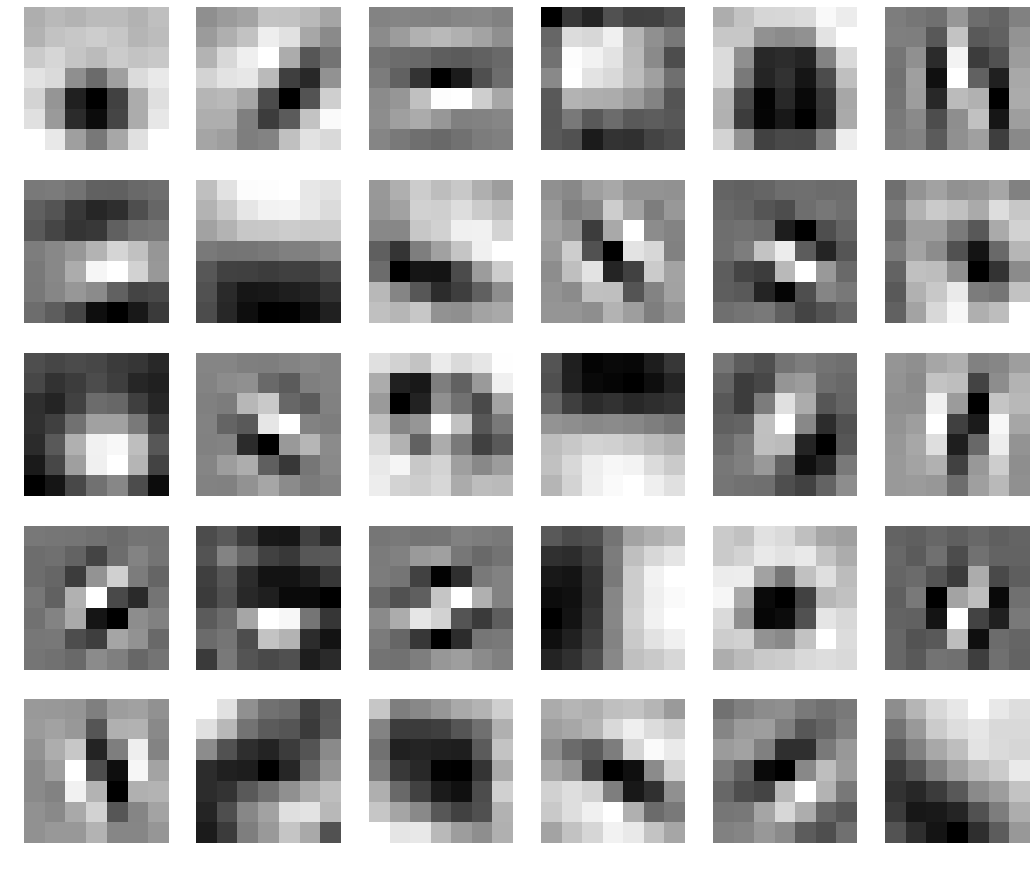}
        \caption{transfer init}
    \end{subfigure}
    \begin{subfigure}{.24\linewidth}
        \includegraphics[width=\linewidth]{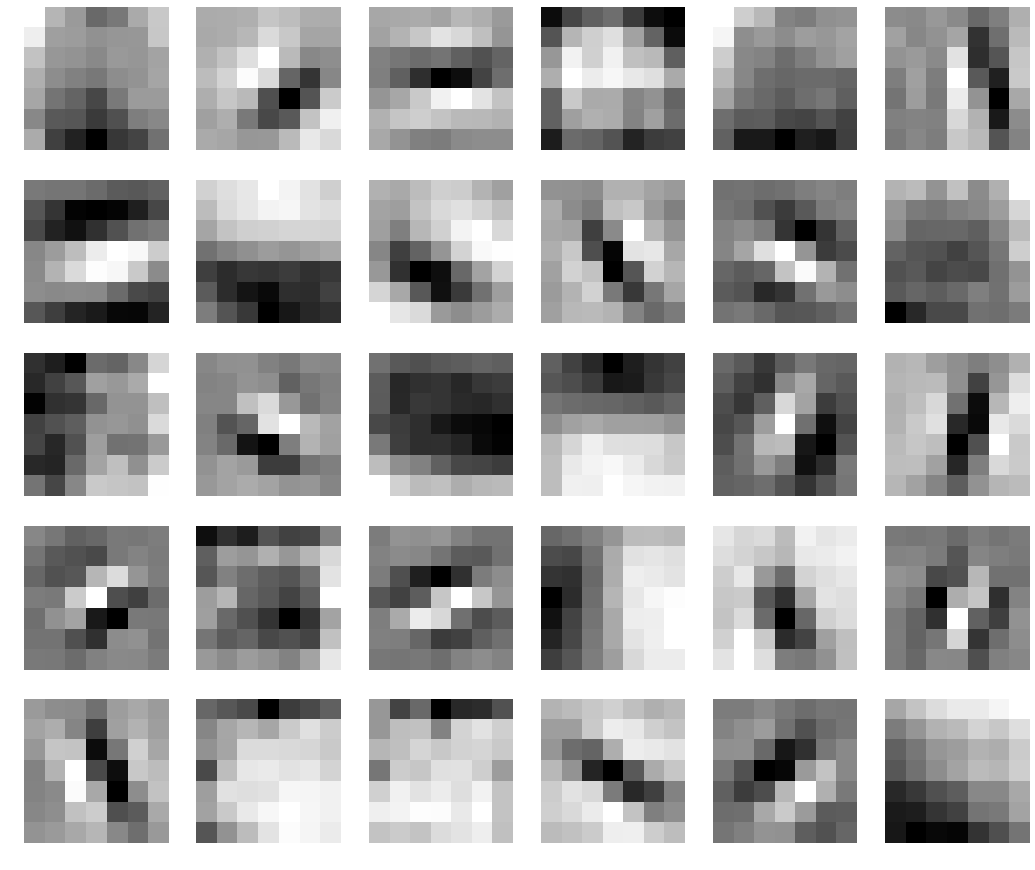}
        \caption{final (transfer init)}
    \end{subfigure}
    \caption{\small\textbf{First layer filters of Resnet-50 on the \chexpert data.} (a) and (c) show
    the randomly initialized filters and filters initialized from a model (the same architecture)
    pre-trained on \imagenet. (b) and (d) shows the final converged filters from the two different initializations,
    respectively.}
    \label{fig:chexpert-resnet-weight-vis}
\end{figure}

Figure~\ref{fig:chexpert-vcsh-weight-vis} and Figure~\ref{fig:chexpert-resnet-weight-vis} compare the first layer filters between transfer learning and training from random initialization on the \chexpert data for the CBR-Small and Resnet-50 architectures, respectively. Those results complement Figure~\ref{fig:retina-weight-vis} in the main text.

\begin{figure}
\centering
\begin{tabular}{c}
\hspace{-10mm}\includegraphics[width=.8\columnwidth]{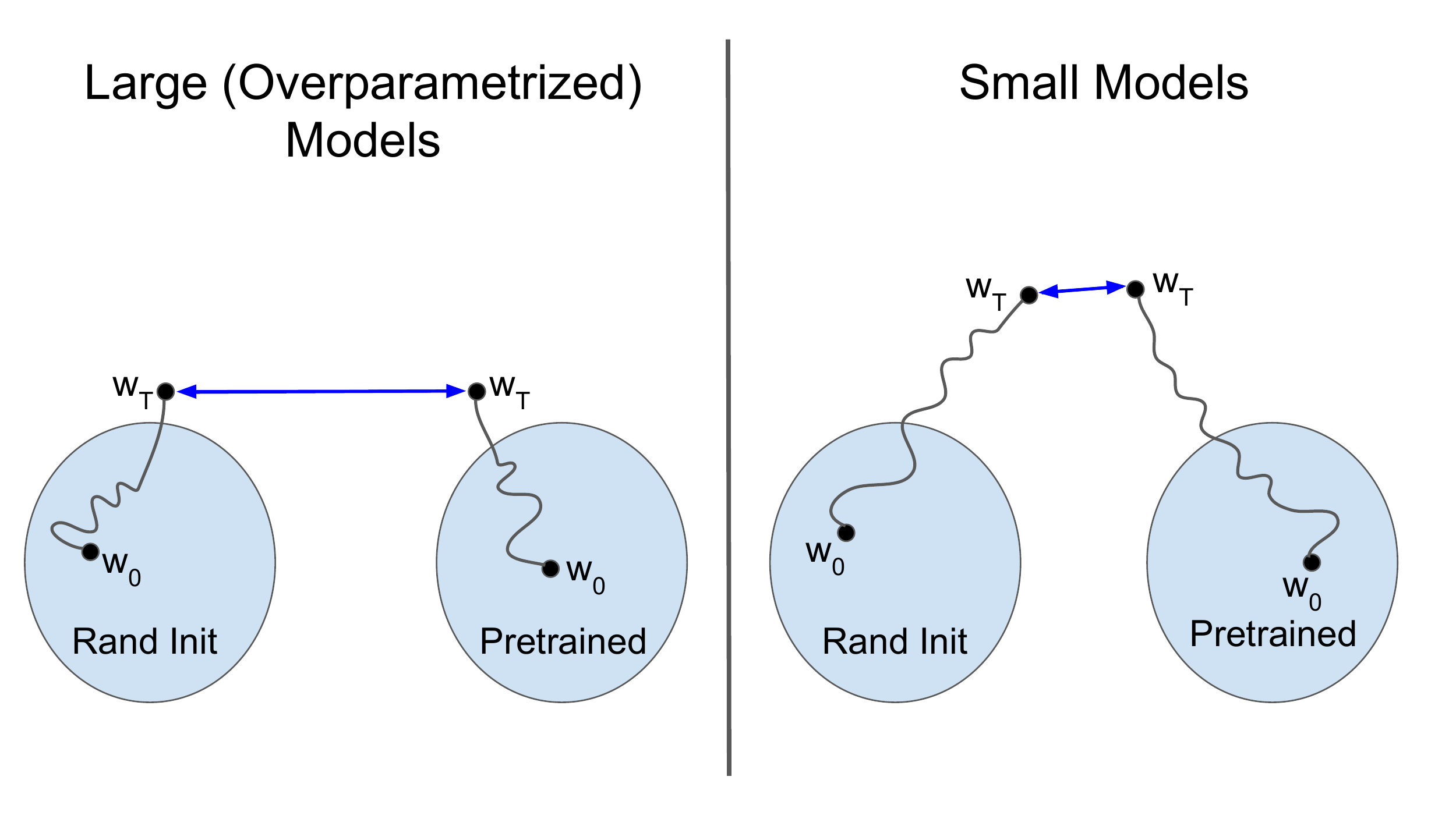}
\end{tabular}
\caption{\small \textbf{Larger models move less through training than smaller networks.} A schematic diagram of our intuition for optimization for larger and smaller models.}
\label{fig:modelsize-schematic}
\end{figure}

\section{The Fixed Feature Extraction Setting}
To complete the picture, we also study the fixed feature extractor setting. While the most popular methodology for transfer learning is to initialize from pretrained weights and fine-tune (train) the entire network, an alternative is to initialize all layers up to layer $L$ with pretrained weights. These are then treated as a fixed feature extractor, with only layers $L+1$ onwards, being trained. There are two variants of this fixed feature extractor experiment:
\textbf{[1]} Initialize all layers with pretrained weights and only train layer $L+1$ onwards. \textbf{[2]} Initialize only up to layer $L$ with pretrained weights, and layer $L+1$ onwards randomly; then train only layers $L+1$ onwards.

\begin{figure}
\centering
\begin{tabular}{ccc}
\hspace*{-17mm} \includegraphics[width=0.4\columnwidth]{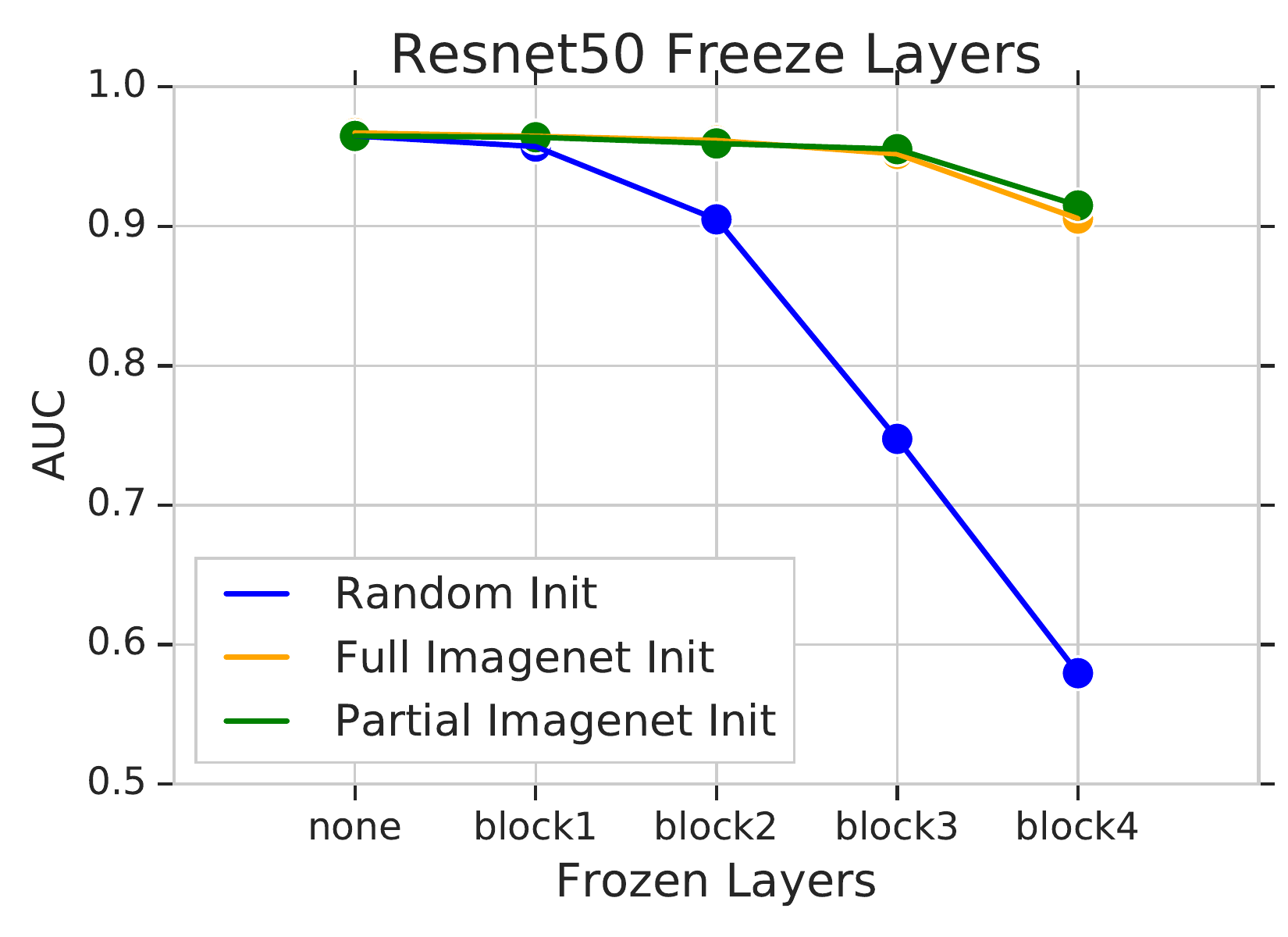} &
\hspace*{-5mm} \includegraphics[width=0.4\columnwidth]{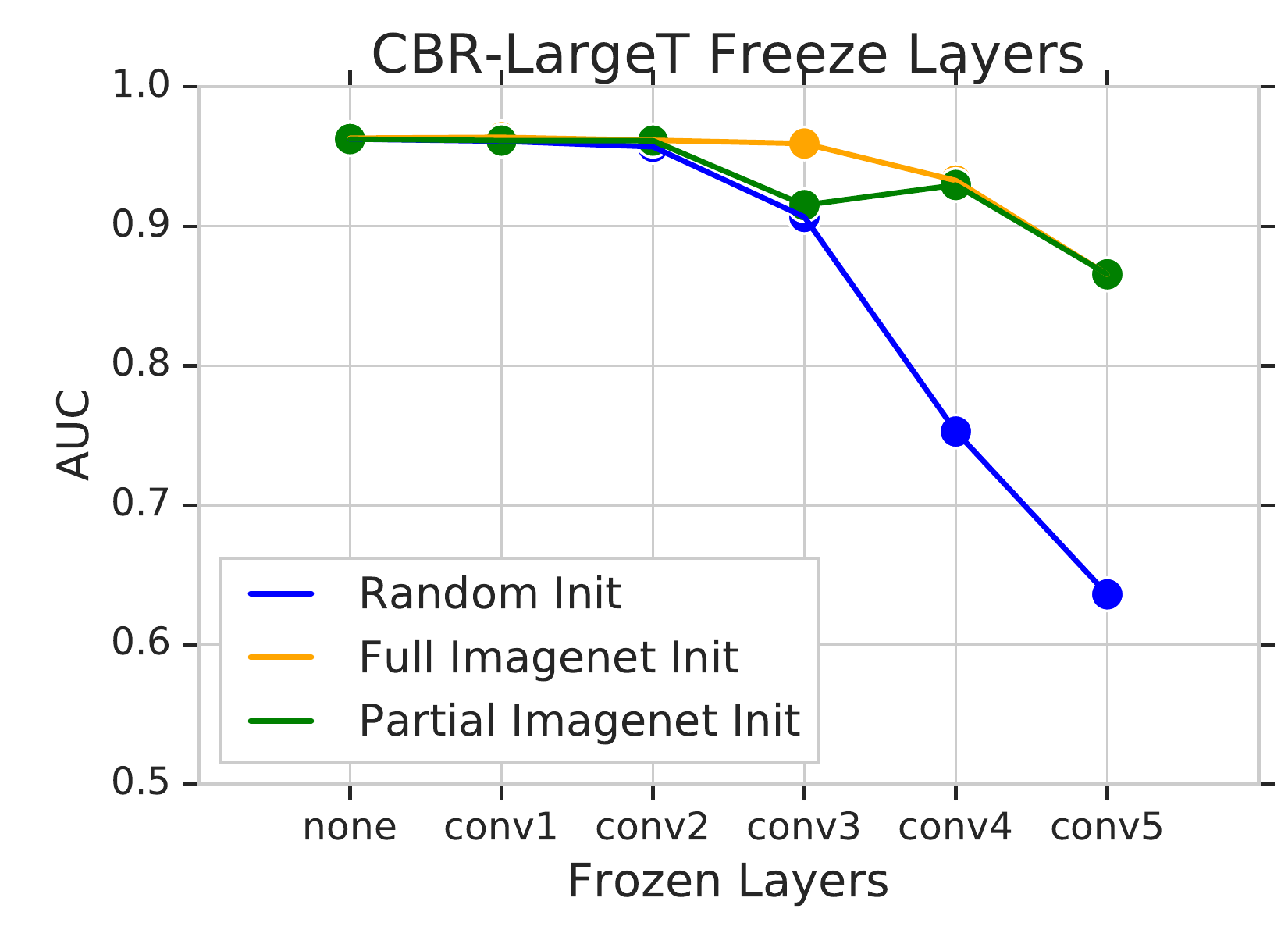} &
\hspace*{-5mm} \includegraphics[width=0.4\columnwidth]{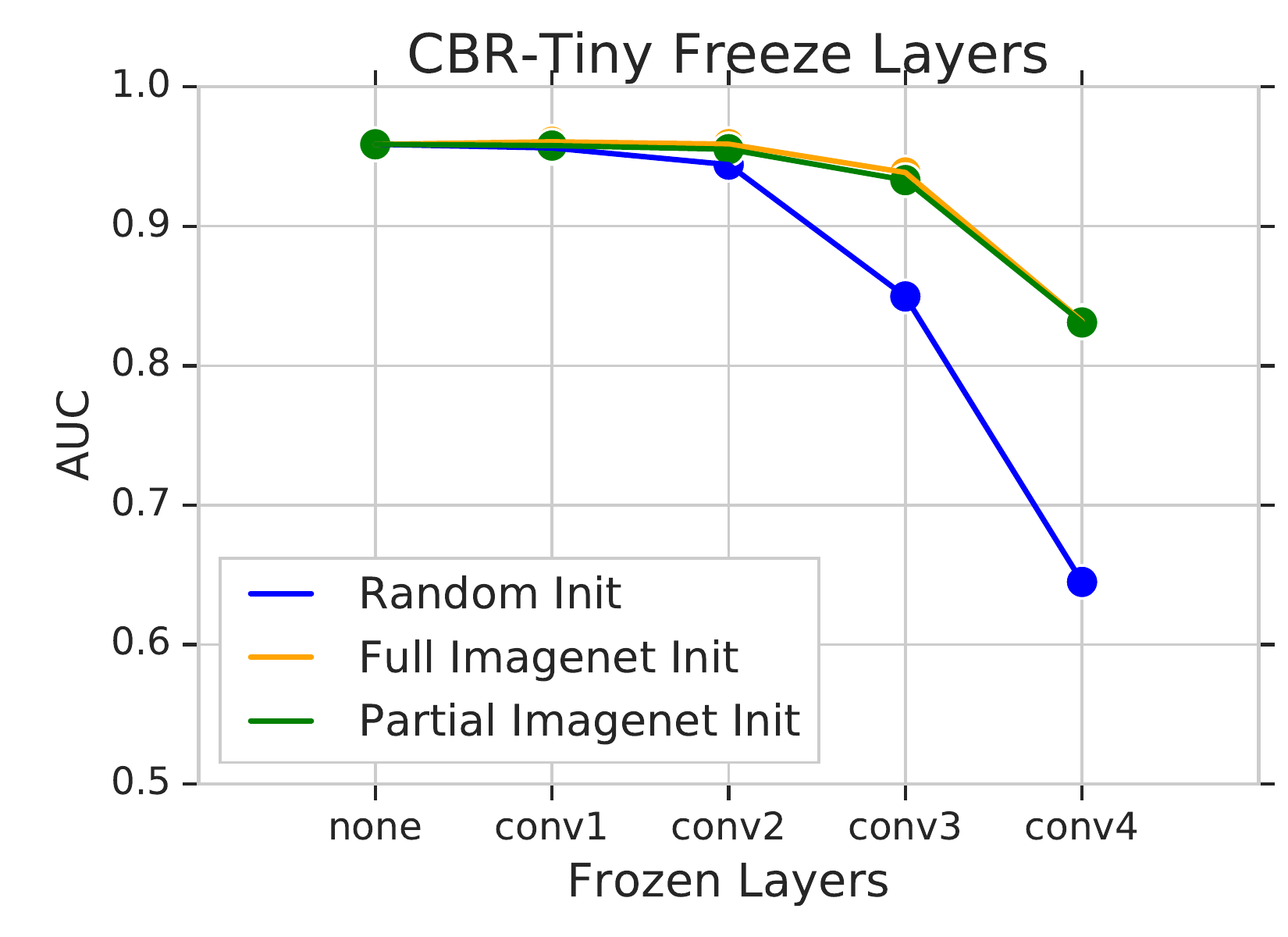}
\end{tabular}
\caption{\small \textbf{\imagenet features perform well as fixed feature extractors on the \retina task, and are robust to coadaptation performance drops.} We initialize (i) the full architecture with \imagenet weights (yellow) (ii) up to layer $L$ with \imagenet weights, and the rest randomly. In both, we keep up to layer $L$ fixed, and only train layers $L+1$ onwards. We compare to a random features baseline, initializing randomly and training layer $L+1$ onwards (blue). \imagenet features perform much better as fixed feature extractors than the random baseline (though this gap is much closer for the \chexpert dataset, Appendix Figure \ref{fig-freeze-layers-chexpert}.) Interestingly, there is no performance drop due to the \textit{coadaptation} issue \citep{yosinski2014transferable}, with partial \imagenet initialization performing equally to initialzing with all of the \imagenet weights.}
\label{fig:freezing-results}
\end{figure}

\begin{figure}
\begin{tabular}{ccc}
\includegraphics[width=.32\linewidth]{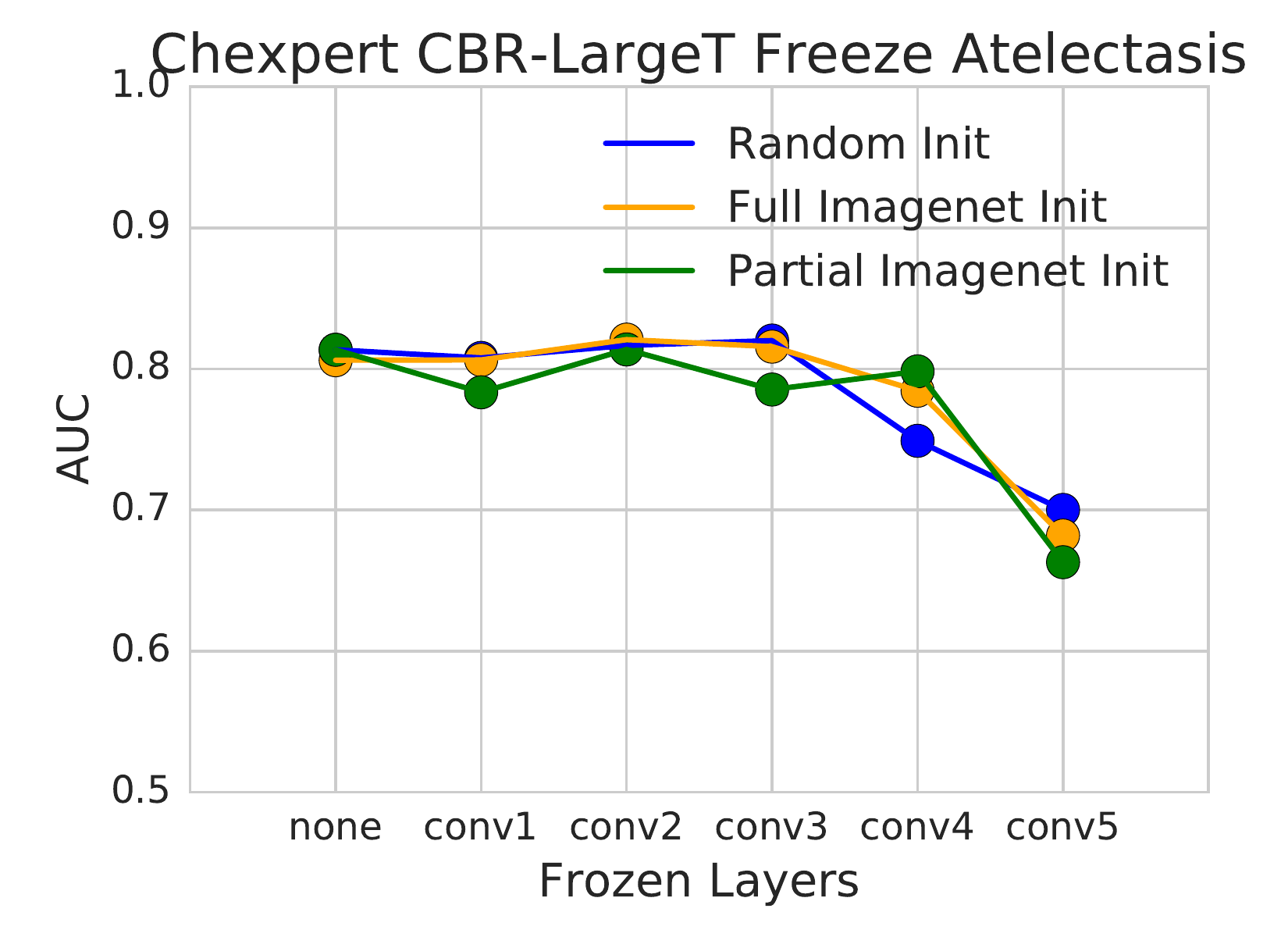}
&
\includegraphics[width=.32\linewidth]{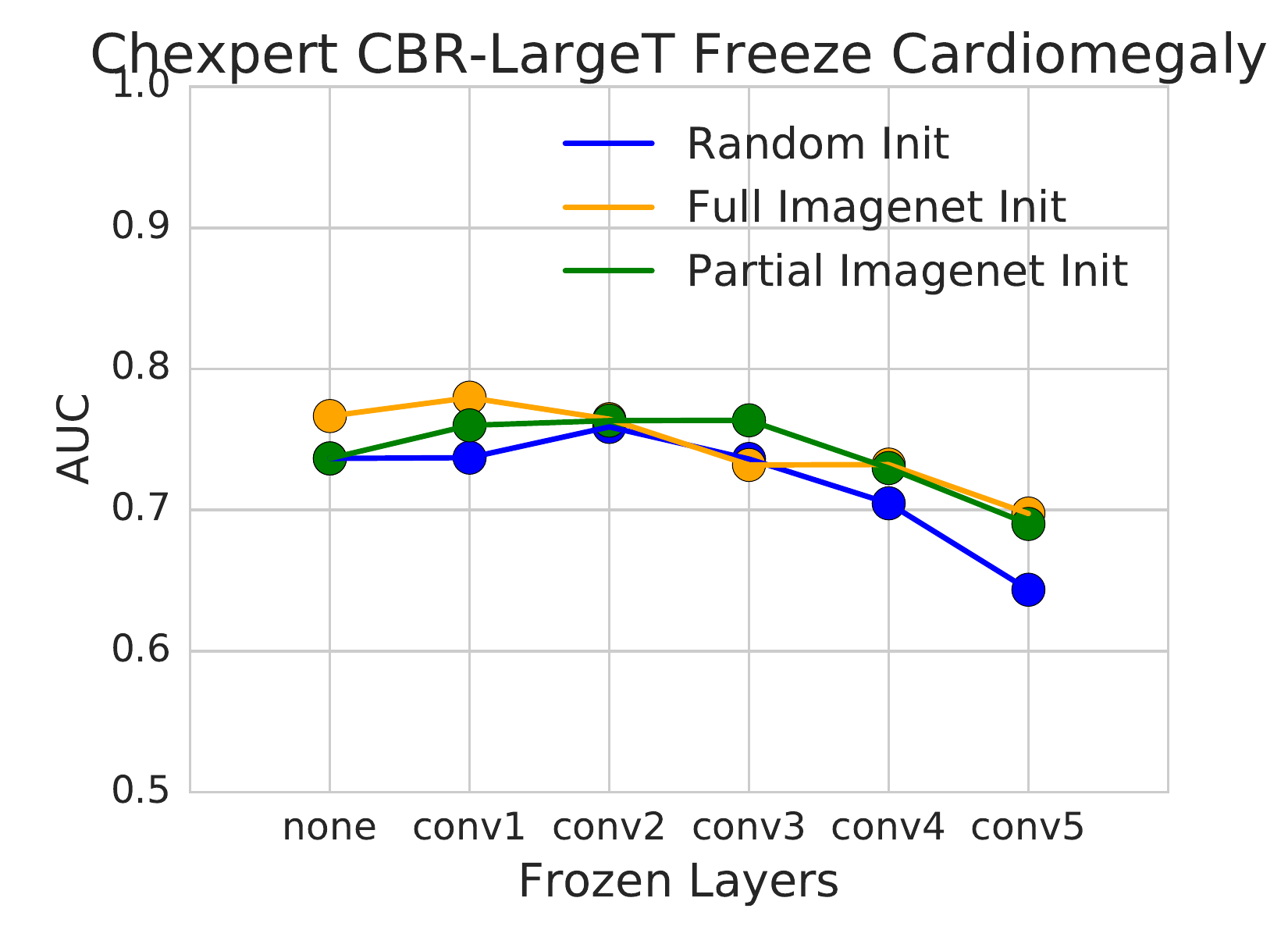}
&
\includegraphics[width=.32\linewidth]{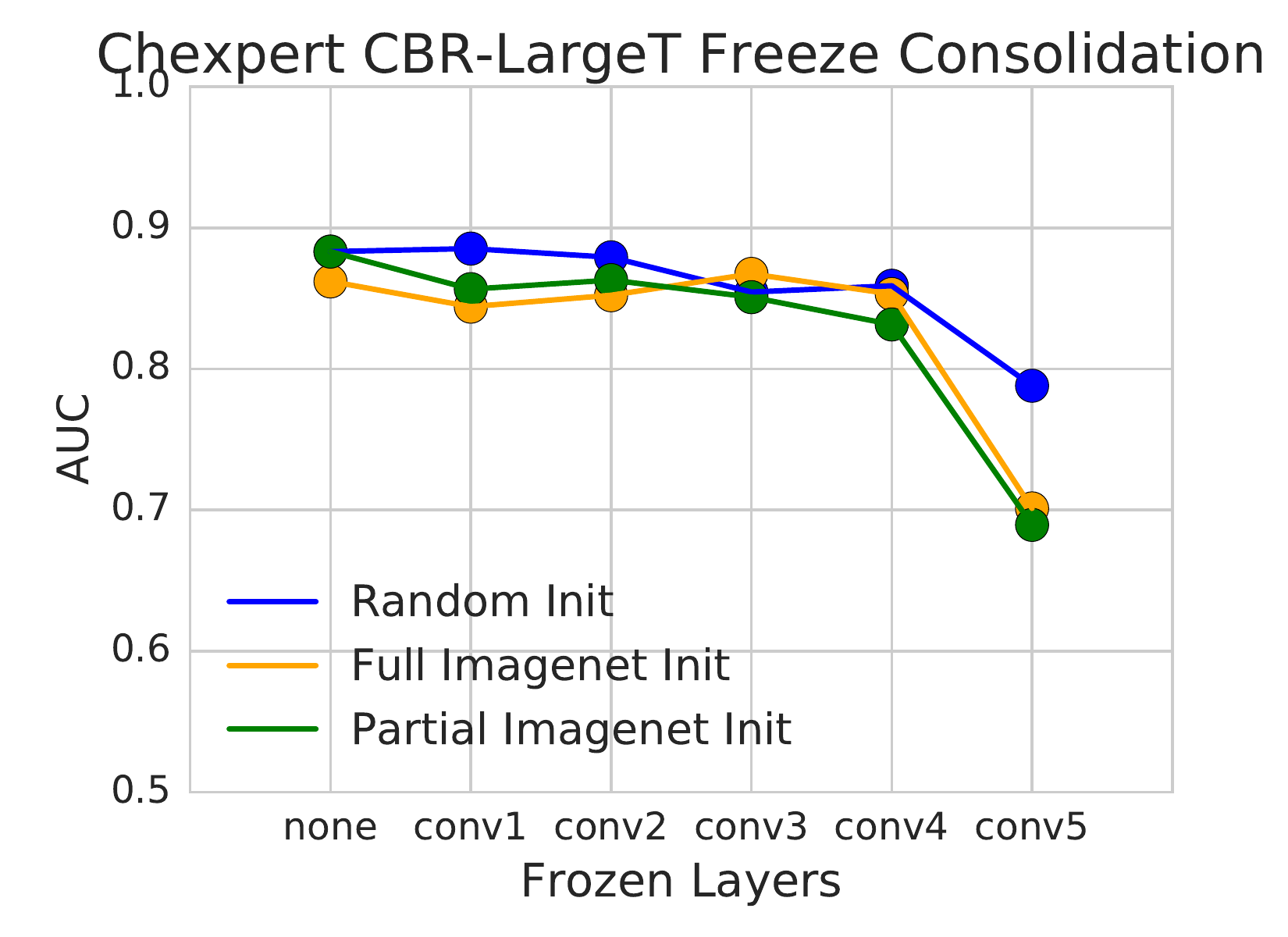} \\
\includegraphics[width=.32\linewidth]{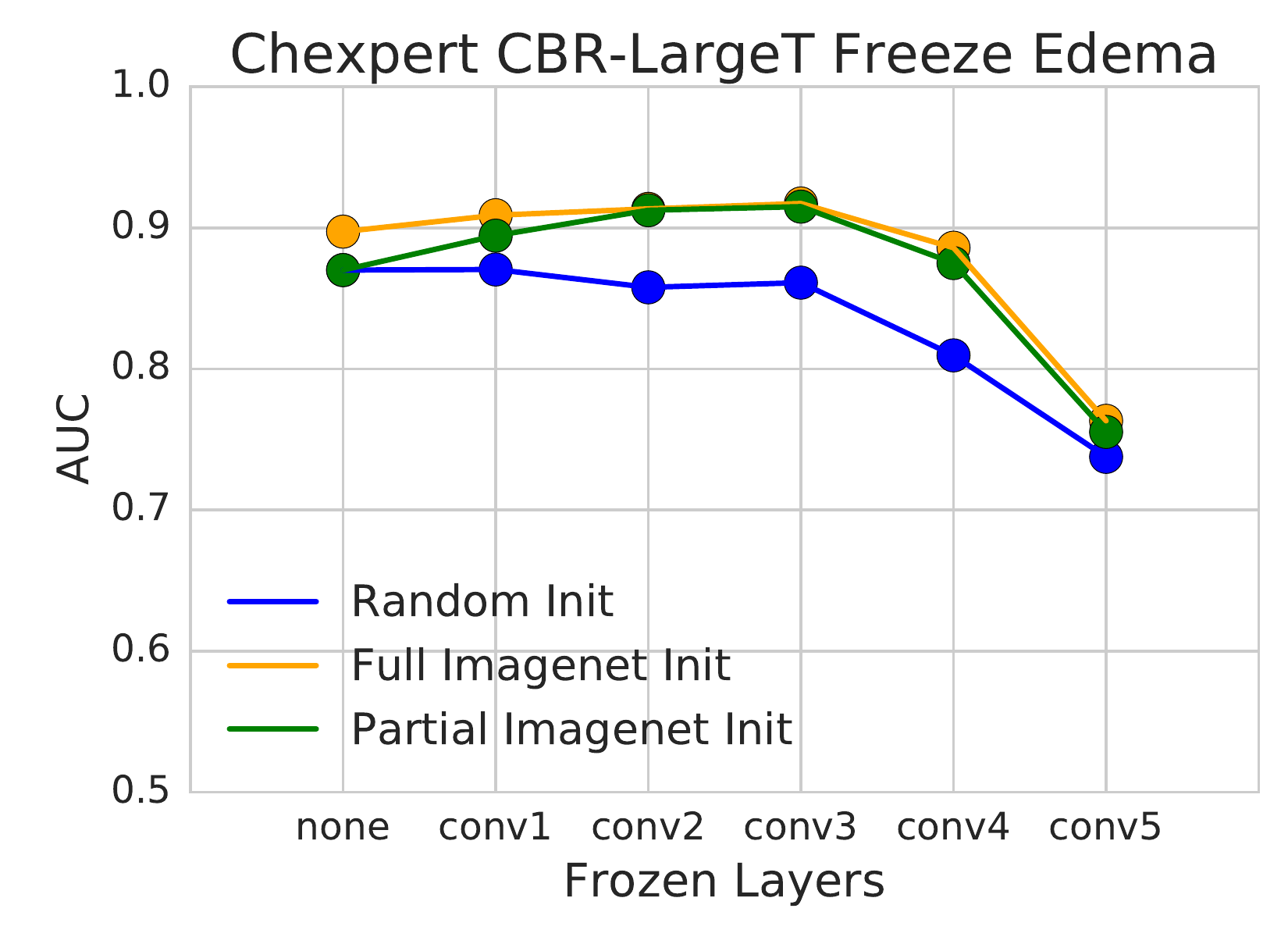}
&
\includegraphics[width=.32\linewidth]{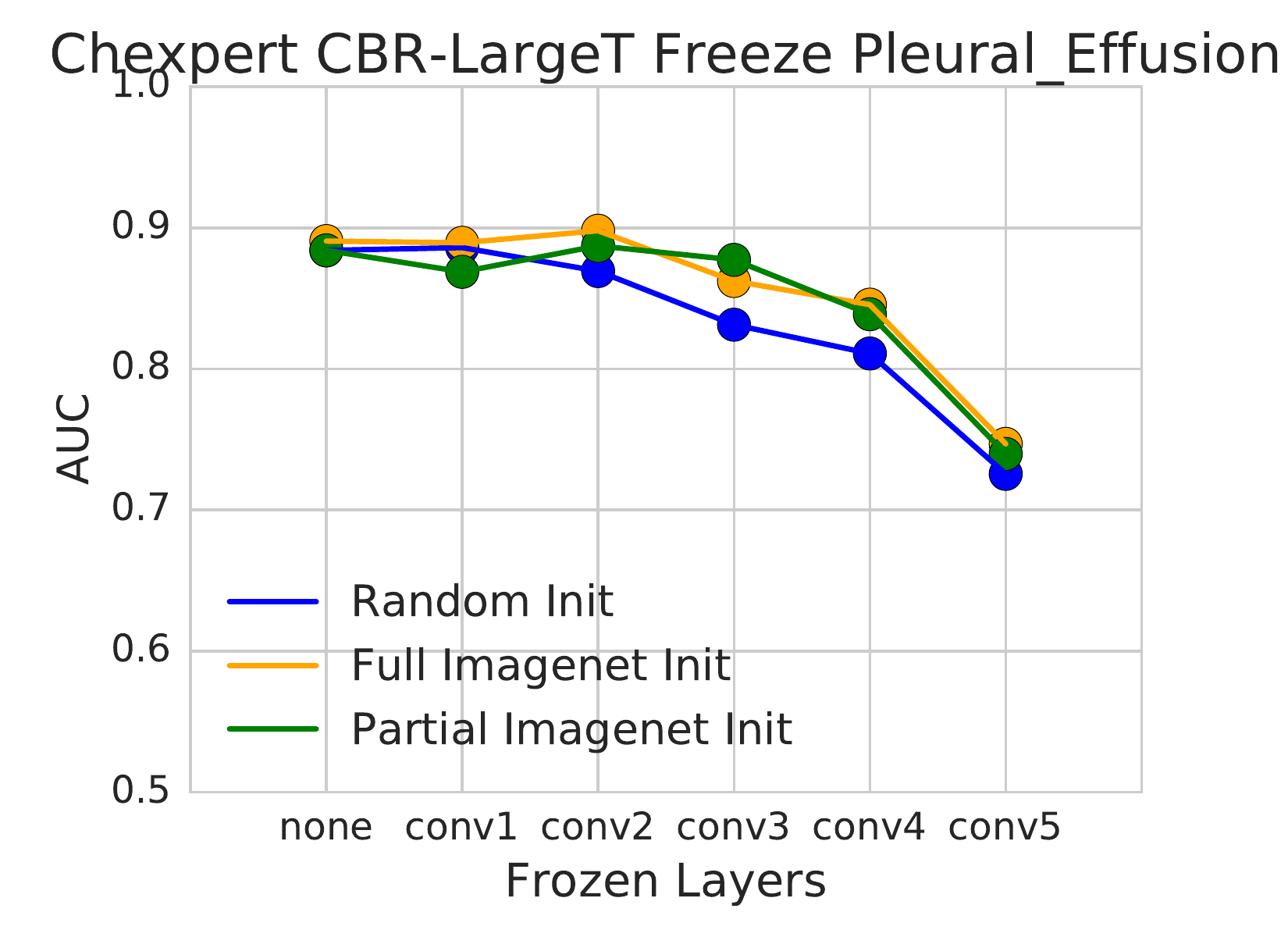}
&
\includegraphics[width=.32\linewidth]{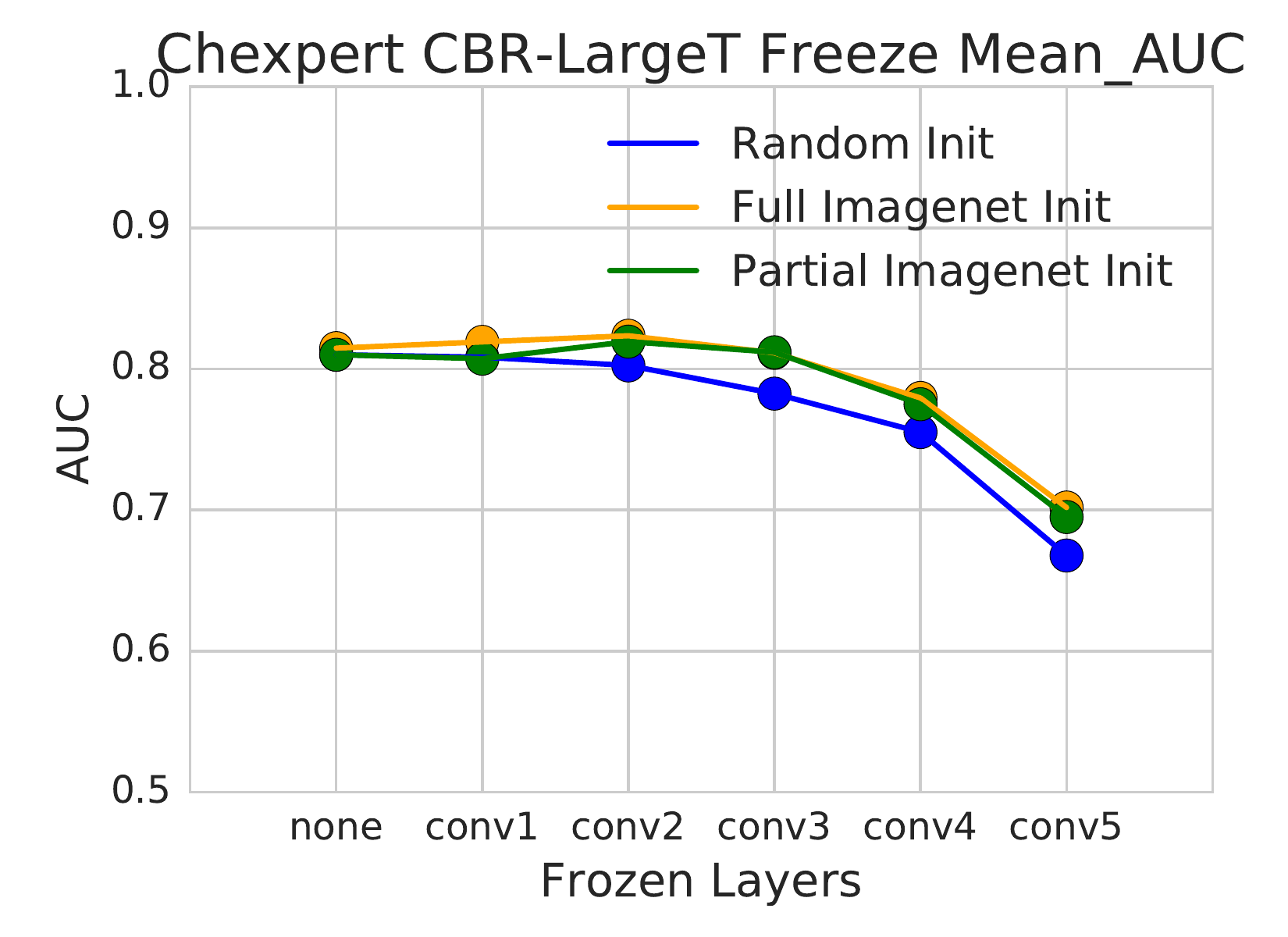} \\
\includegraphics[width=.32\linewidth]{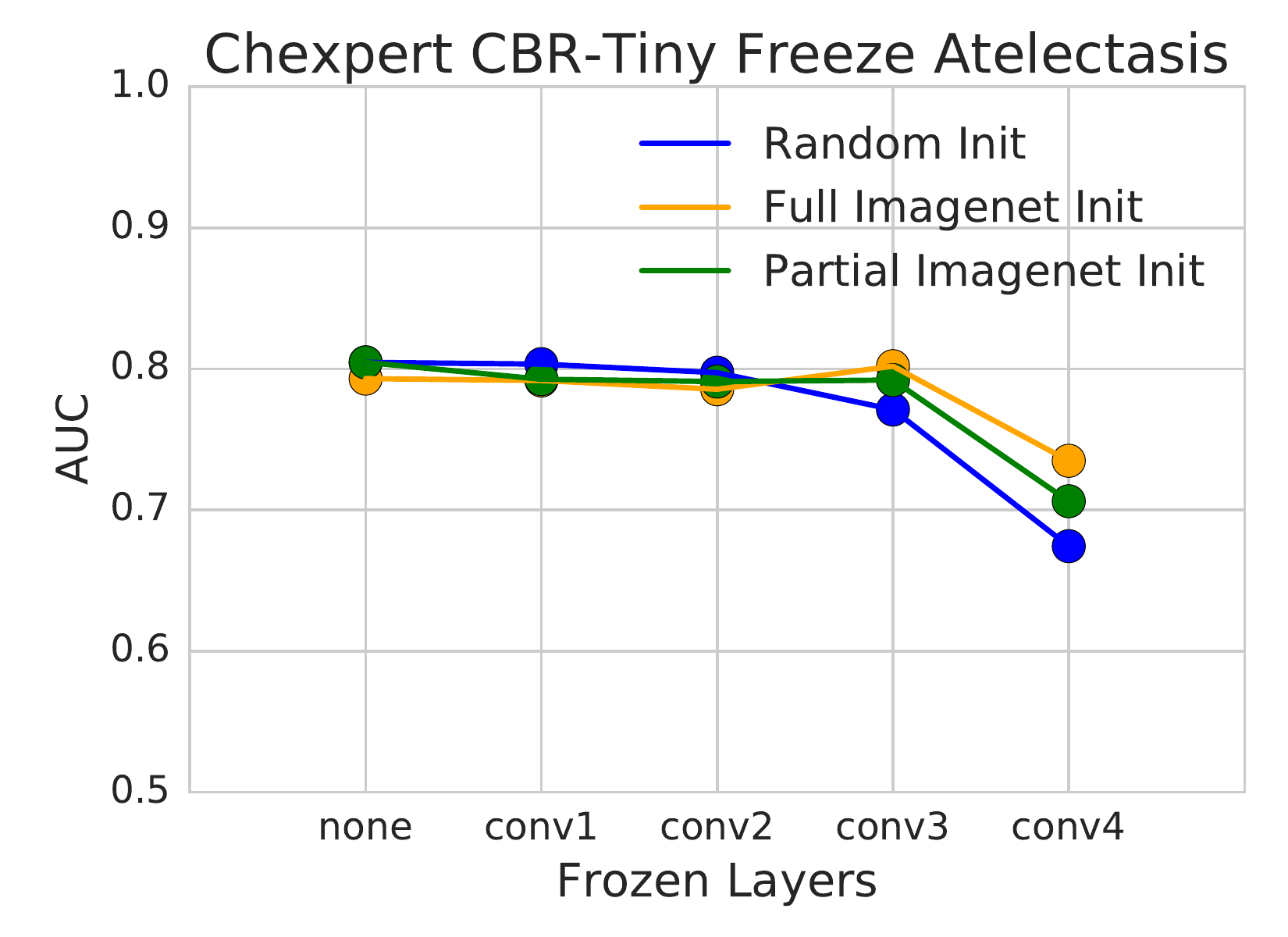}
&
\includegraphics[width=.32\linewidth]{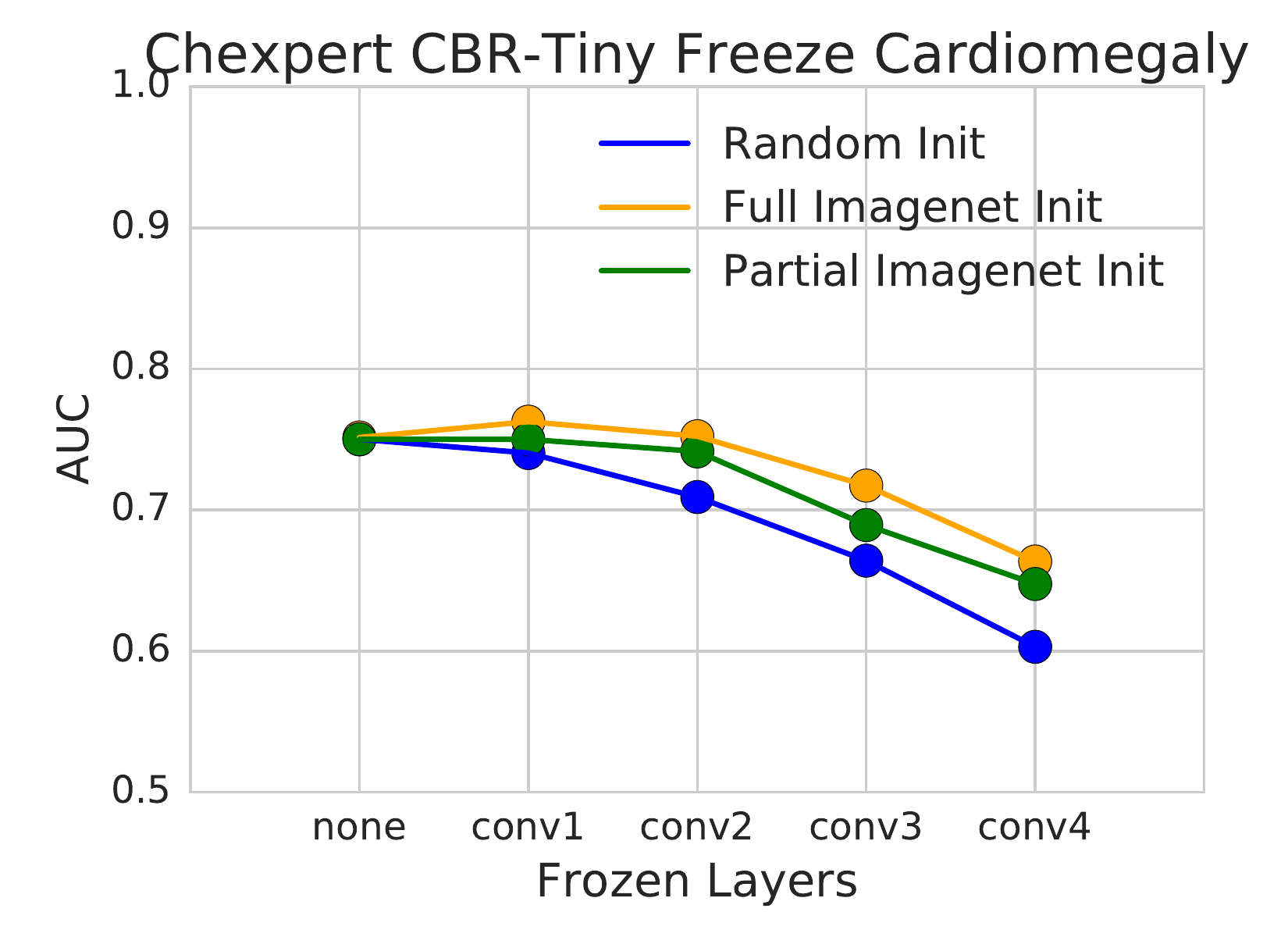}
&
\includegraphics[width=.32\linewidth]{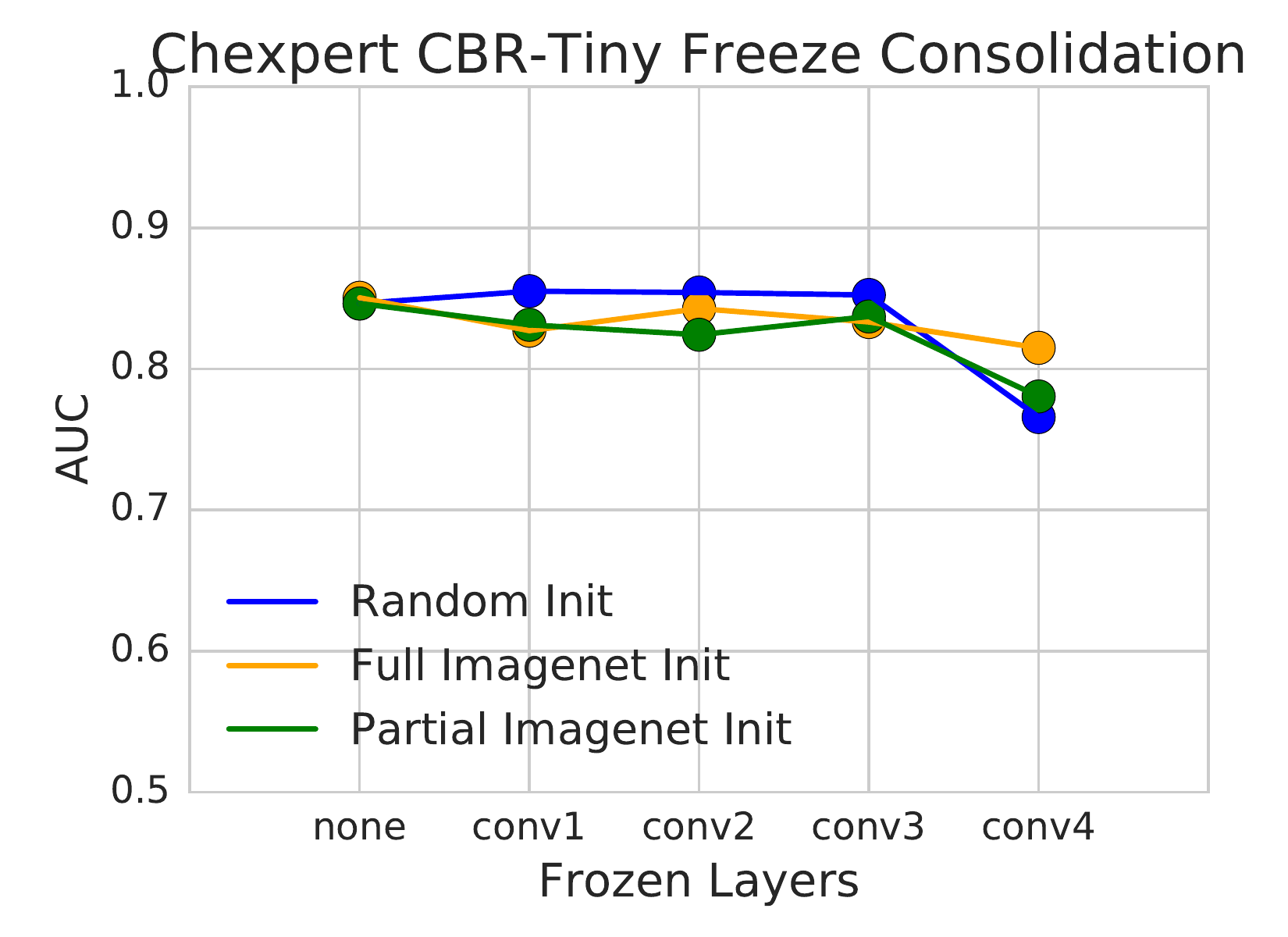} \\
\includegraphics[width=.32\linewidth]{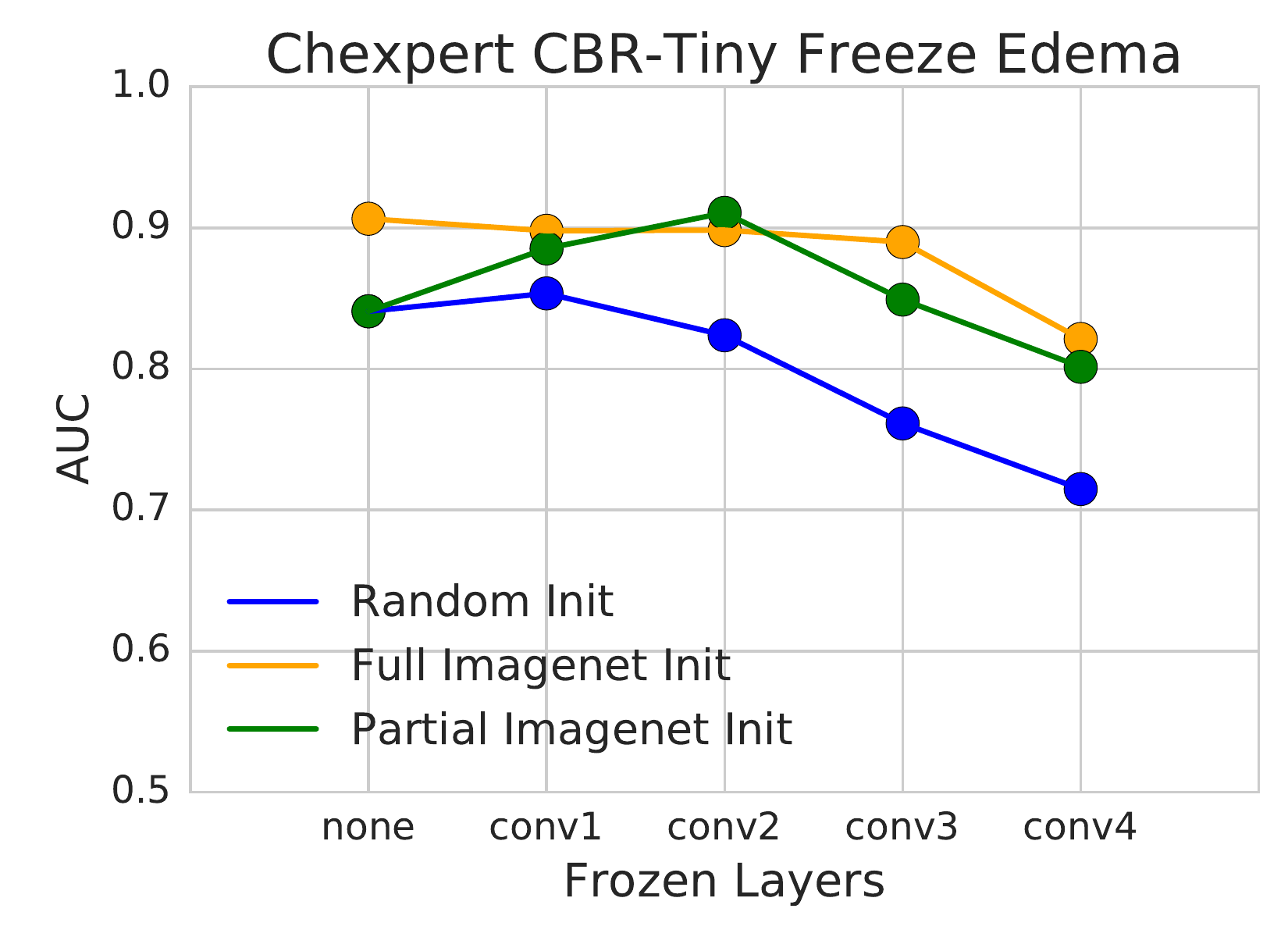}
&
\includegraphics[width=.32\linewidth]{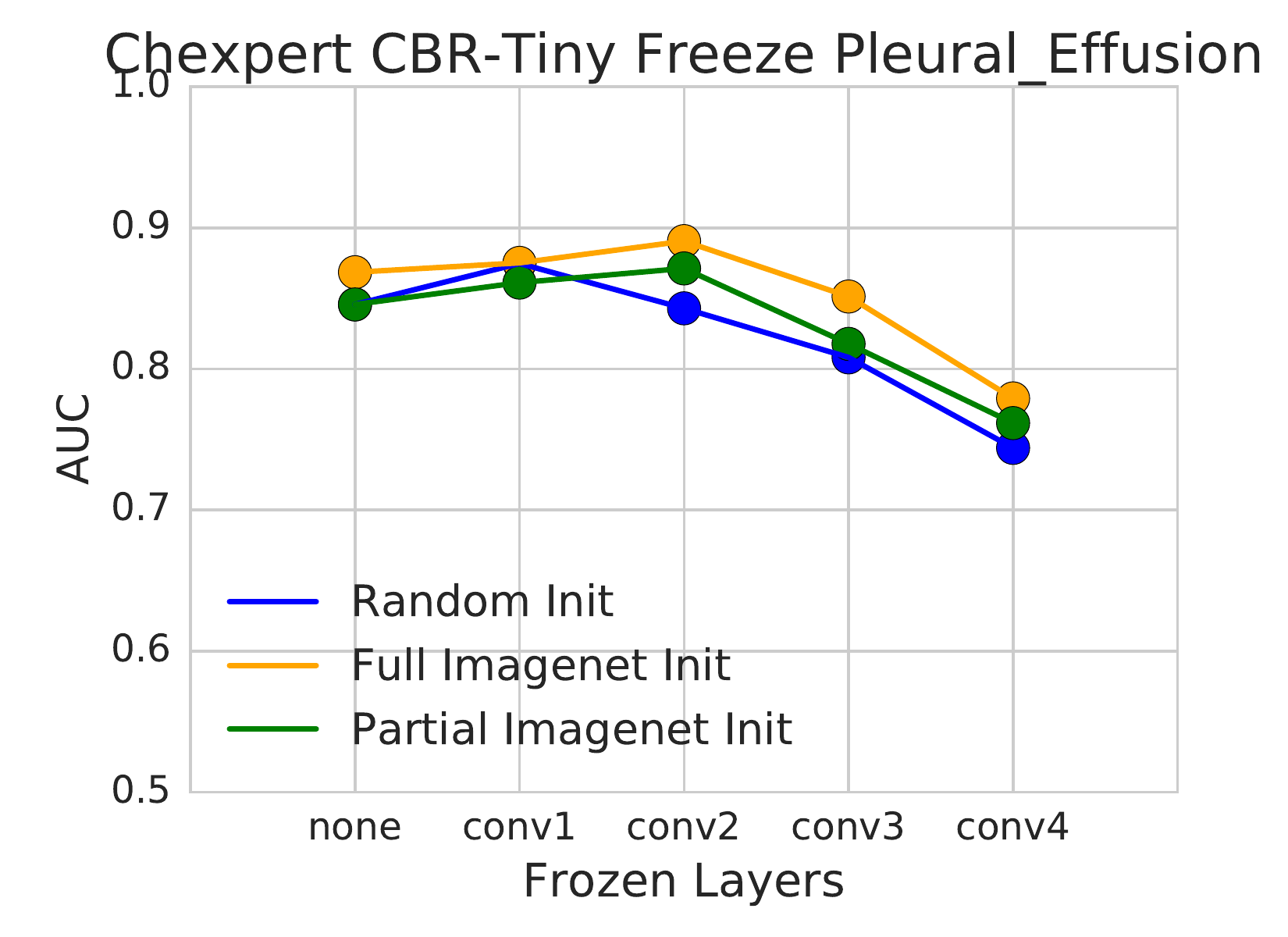}
&
\includegraphics[width=.32\linewidth]{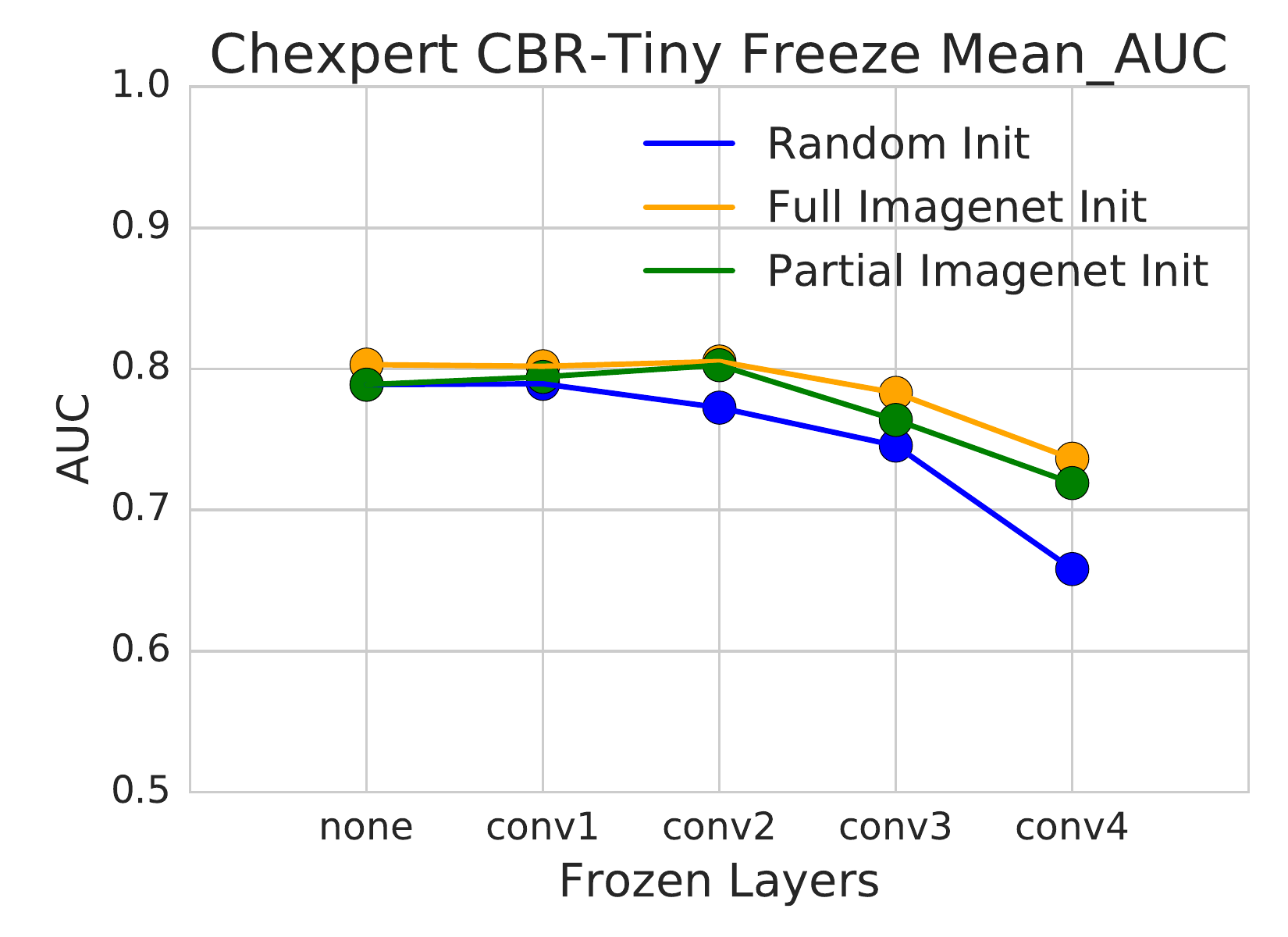} \\
\end{tabular}
\caption{\small \textbf{Experiments on freezing lower layers of CBR-LargeT and a CBR-Tiny model on the \chexpert data.} After random or transfer initialization, we keep up to layer $L$ fixed, and only train layers $L+1$ onwards. \imagenet features perform better as fixed feature extractors than the random baseline for most diseases, but the gap is much closer than for the \retina data, Figure \ref{fig:freezing-results}. We again see that there is no significant performance drop due to coadaptation challenges.}
\label{fig-freeze-layers-chexpert}
\end{figure}

We implement both of these versions across different models trained on the \retina task in Figure \ref{fig:freezing-results}, and \chexpert in Figure \ref{fig-freeze-layers-chexpert}, including a baseline of using random features -- initializing the network randomly, freezing up to layer $L$, and training layer $L+1$ onwards. For the \retina task, we see that the pretrained \imagenet features perform significantly better than the random features baseline, but this gap is significantly closer on the chest x-rays.

More surprisingly however, there is little difference in performance between initializing all layers with pretrained weights and only up to layer $L$ with pretrained weights. This latter experiment has also been studied in \citep{yosinski2014transferable}, where they found that re-initializing caused drops in performance due to \textit{co-adaptation}, where neurons in different layers have evolved together in a way that is not easily discoverable through retraining. This analysis was done for highly similar tasks (different subsets of \imagenet), and we hypothesise that in our setting, the significant changes of the higher layers (Figures \ref{fig:cca-before-after-finetuning}, \ref{fig:cca-start-end-conv1}) means that the correct adaptation is naturally learned through training.

\section{Additional Results on Feature Independent Benefits and Weight Transfusions}
\begin{figure}
\centering
\begin{tabular}{ccc}
\hspace*{-10mm} \includegraphics[width=0.34\columnwidth]{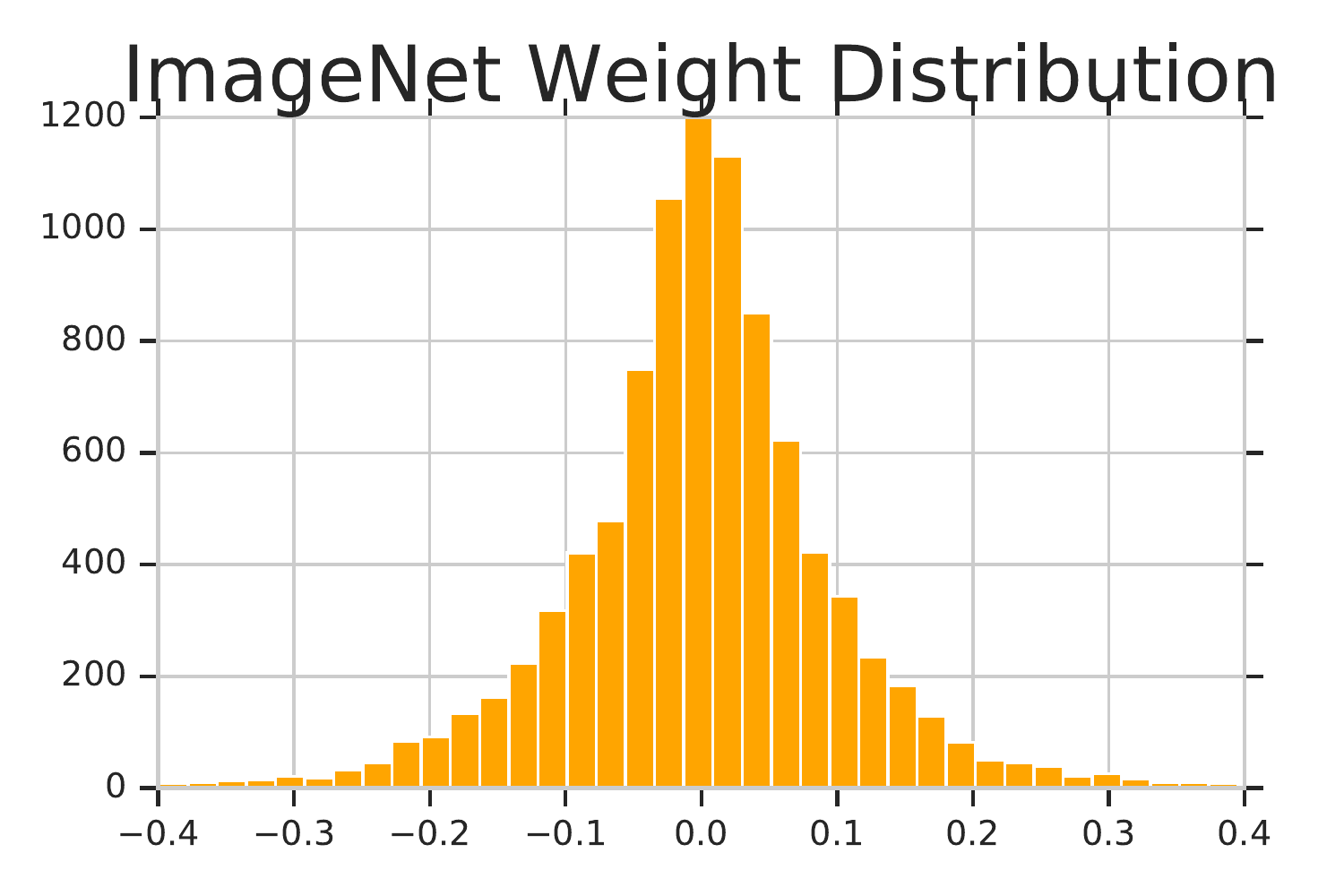} &
\hspace*{-5mm} \includegraphics[width=0.34\columnwidth]{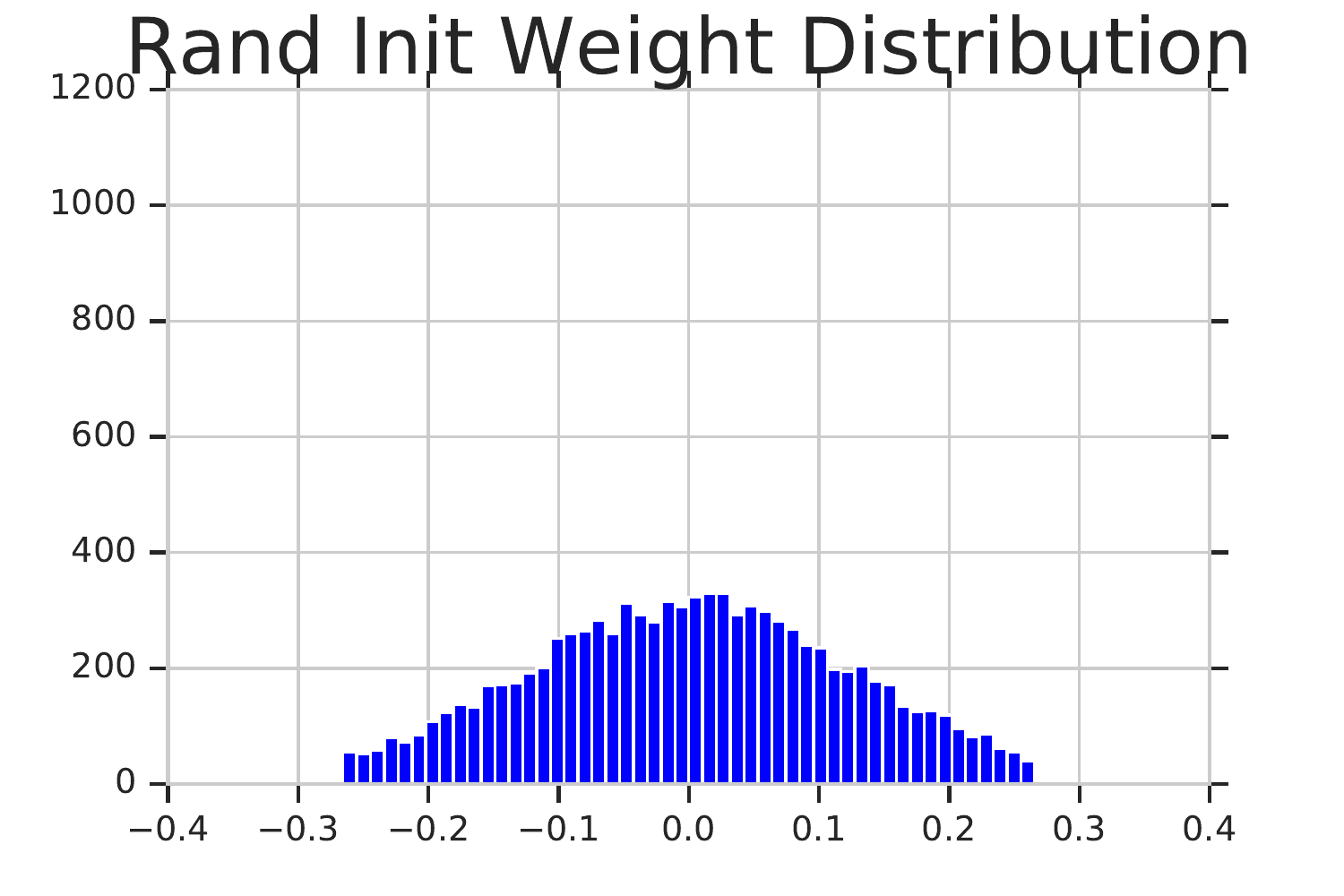} &
\hspace*{-5mm} \includegraphics[width=0.34\columnwidth]{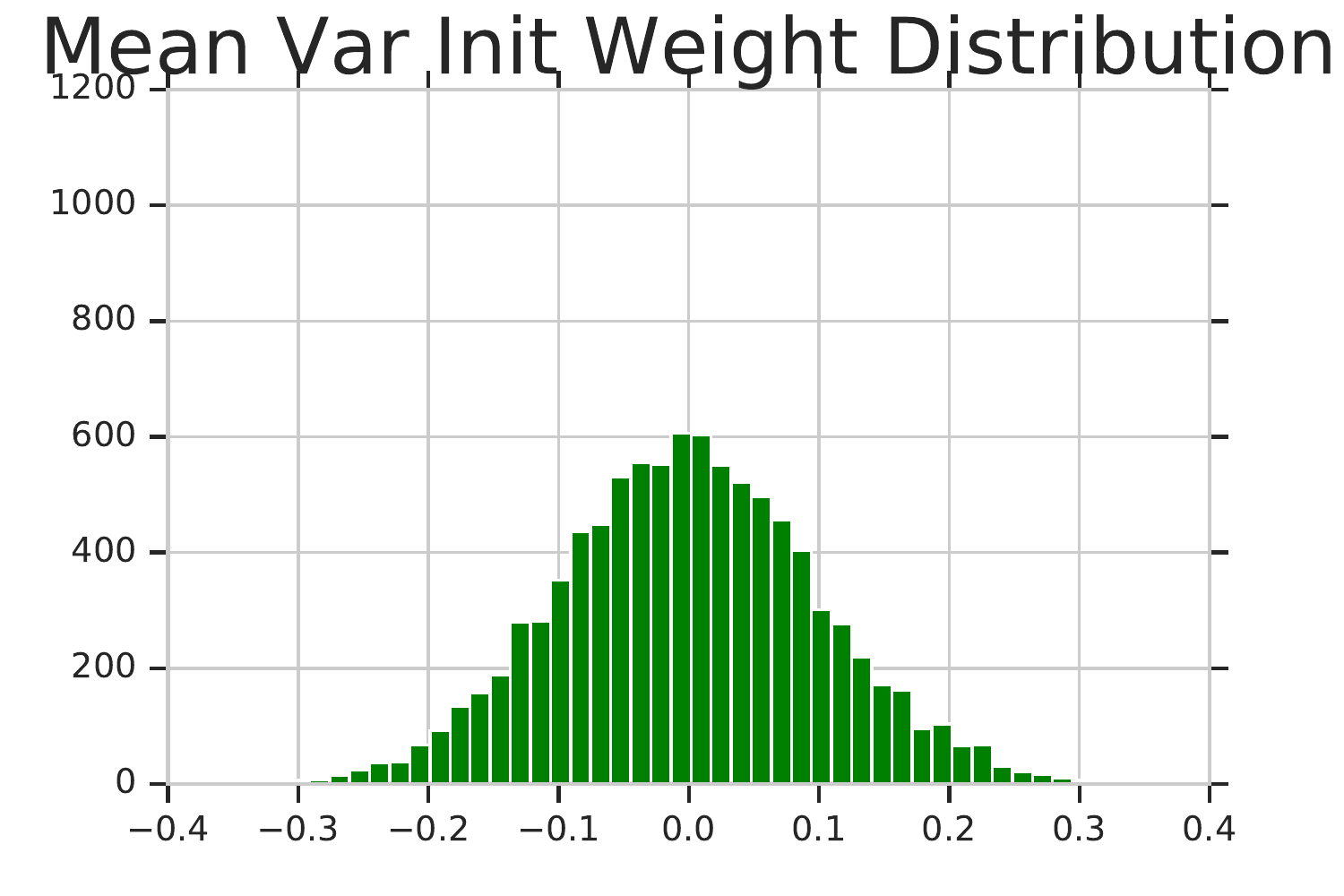} \\
\hspace*{-10mm} \includegraphics[width=0.34\columnwidth]{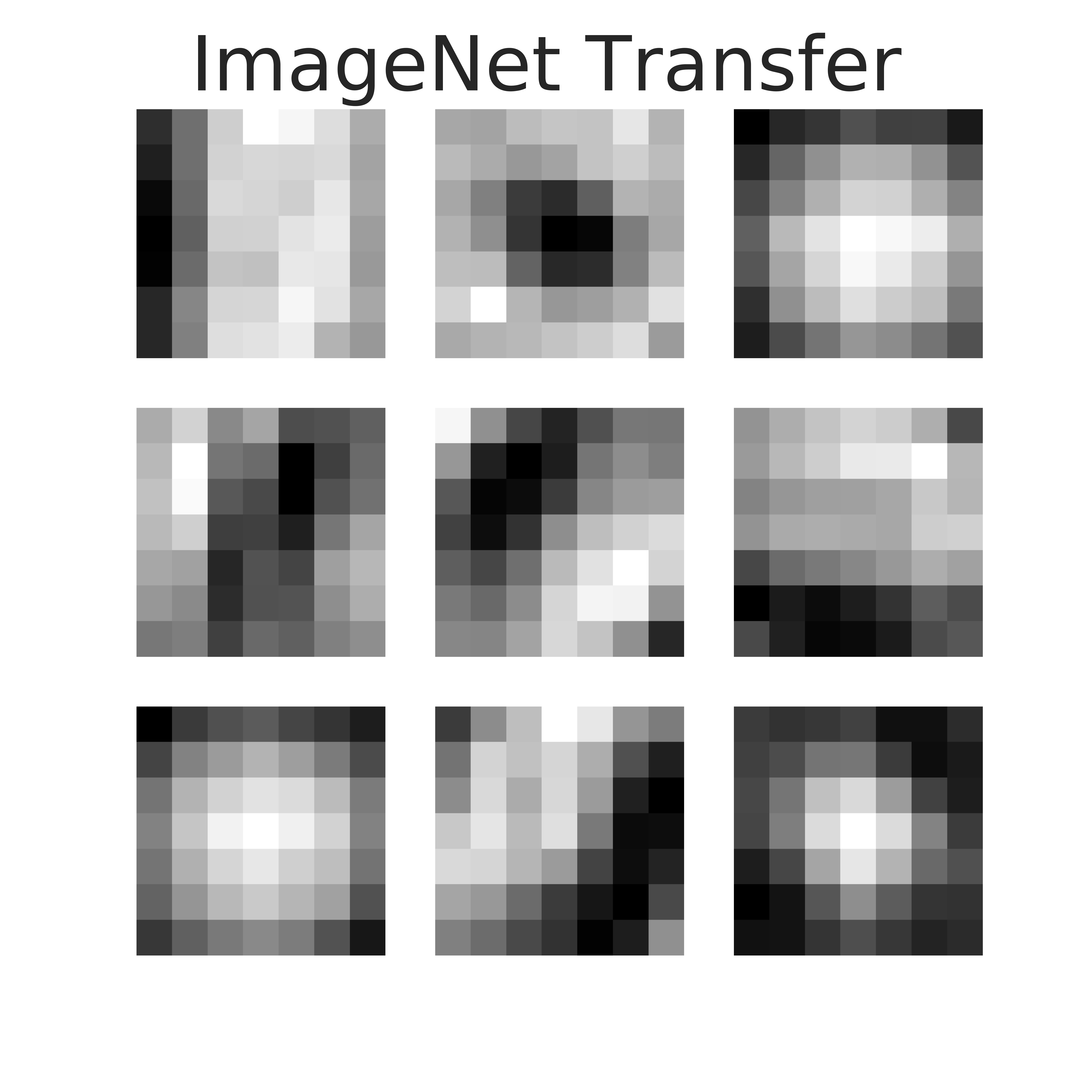} &
\hspace*{-5mm} \includegraphics[width=0.34\columnwidth]{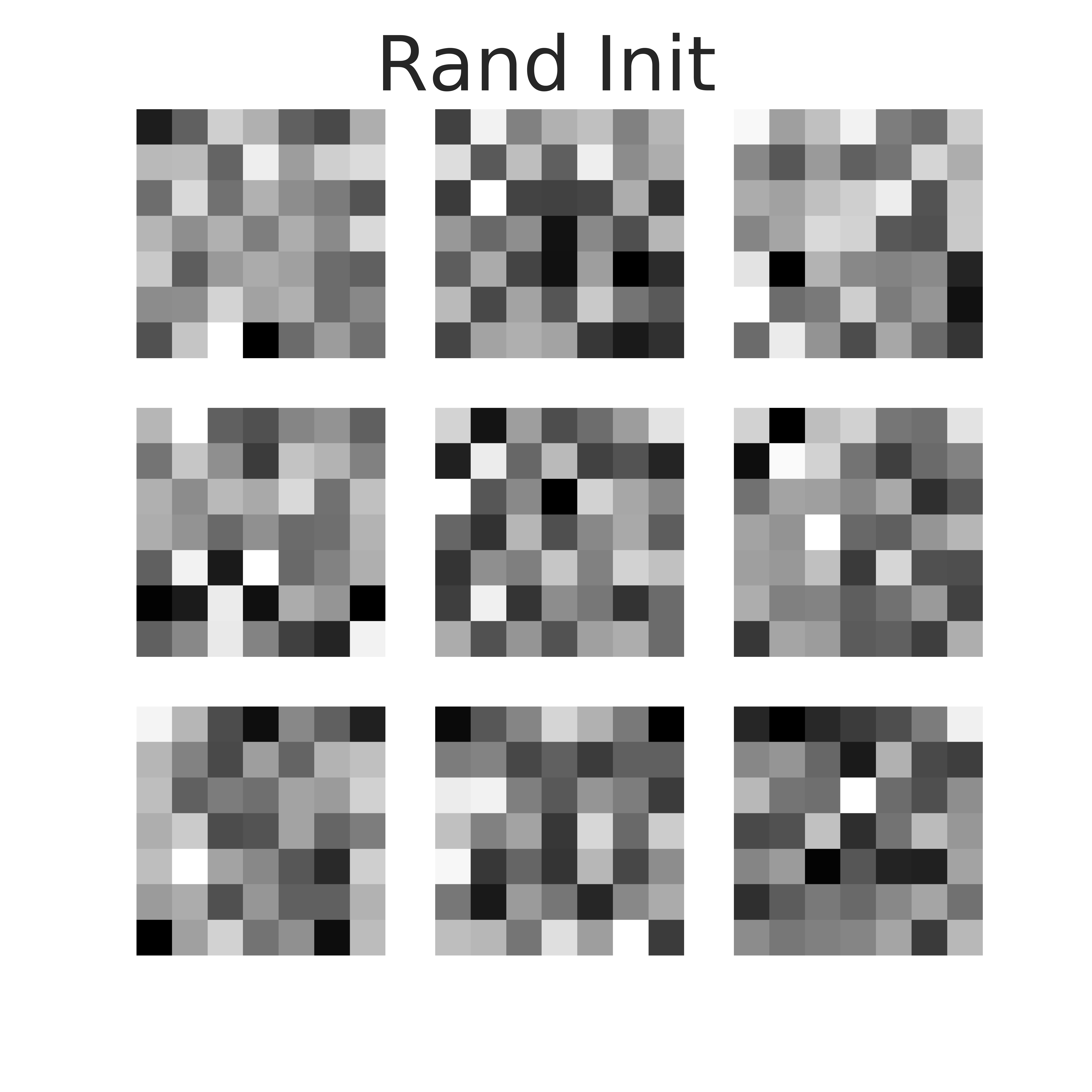} &
\hspace*{-5mm} \includegraphics[width=0.34\columnwidth]{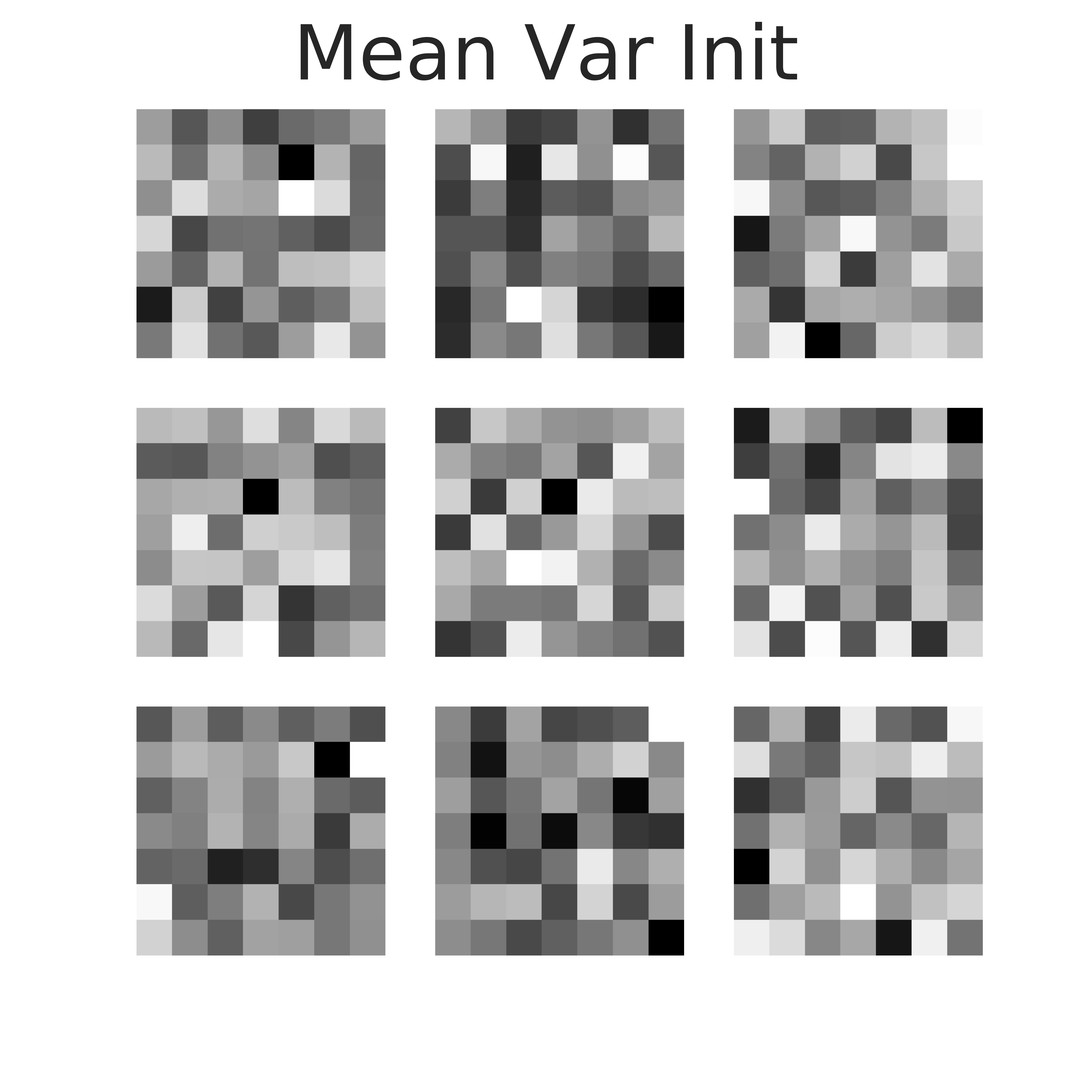}
\vspace{-3.5mm}
\end{tabular}
\caption{\small \textbf{Distribution and filter visualization of weights initialized according to pretrained \imagenet weights, Random Init, and Mean Var Init}. The top row is a histogram of the weight values of the the first layer of the network (Conv 1) when initialized with these three different schemes. The bottom row shows some of the filters corresponding to the different initializations. Only the \imagenet Init filters have pretrained (Gabor-like) structure, as Rand Init and Mean Var weights are iid.}
\label{fig:app-weight-dist}
\end{figure}

\begin{figure}
\centering
\hspace*{-5mm}\begin{tabular}{ccc}
\includegraphics[width=0.33\columnwidth,clip,trim={0 0 0 13mm}]{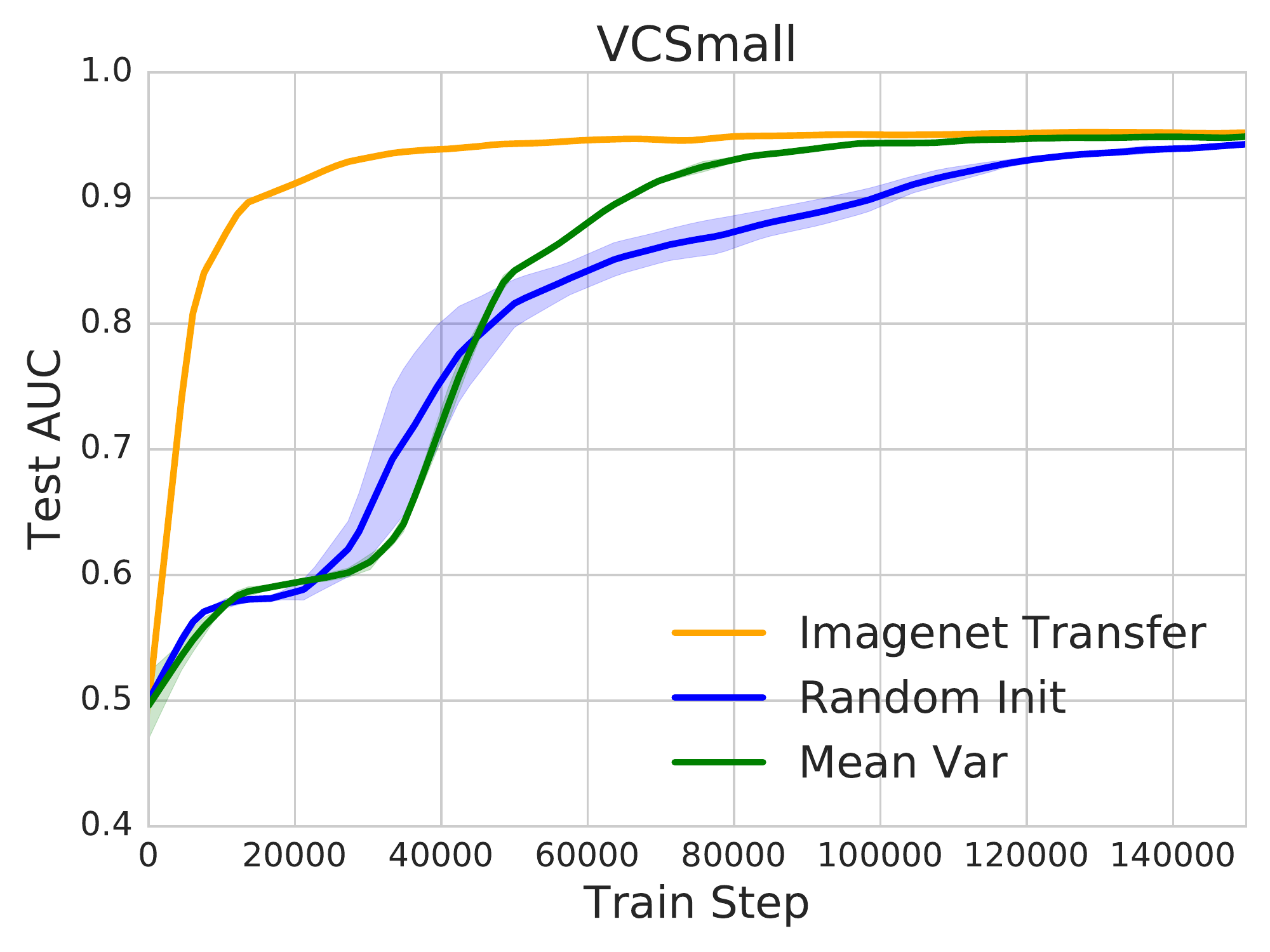} &
\includegraphics[width=0.33\columnwidth,clip,trim={0 0 0 12mm}]{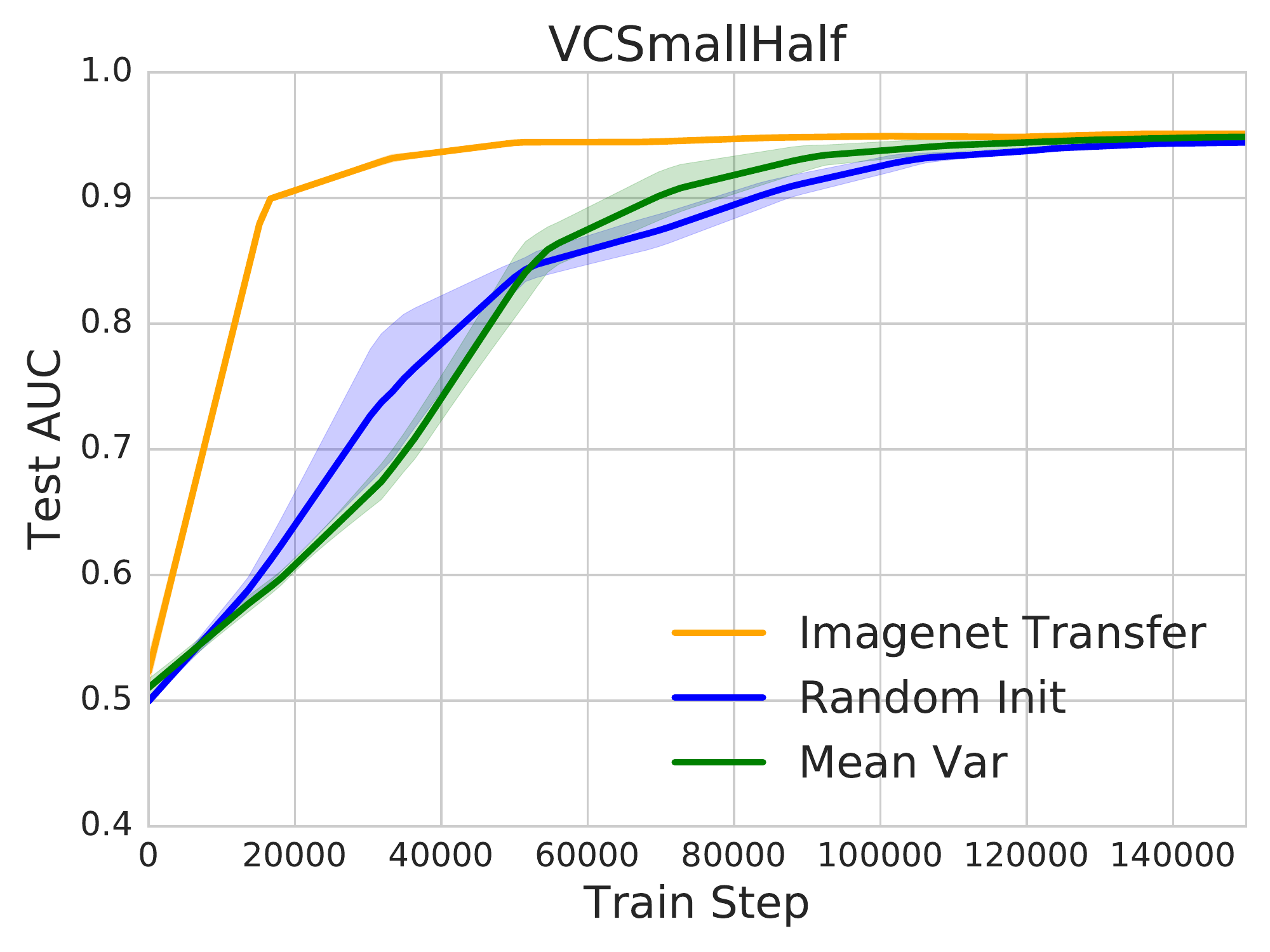}&
\includegraphics[width=0.33\columnwidth,clip,trim={0 0 0 13mm}]{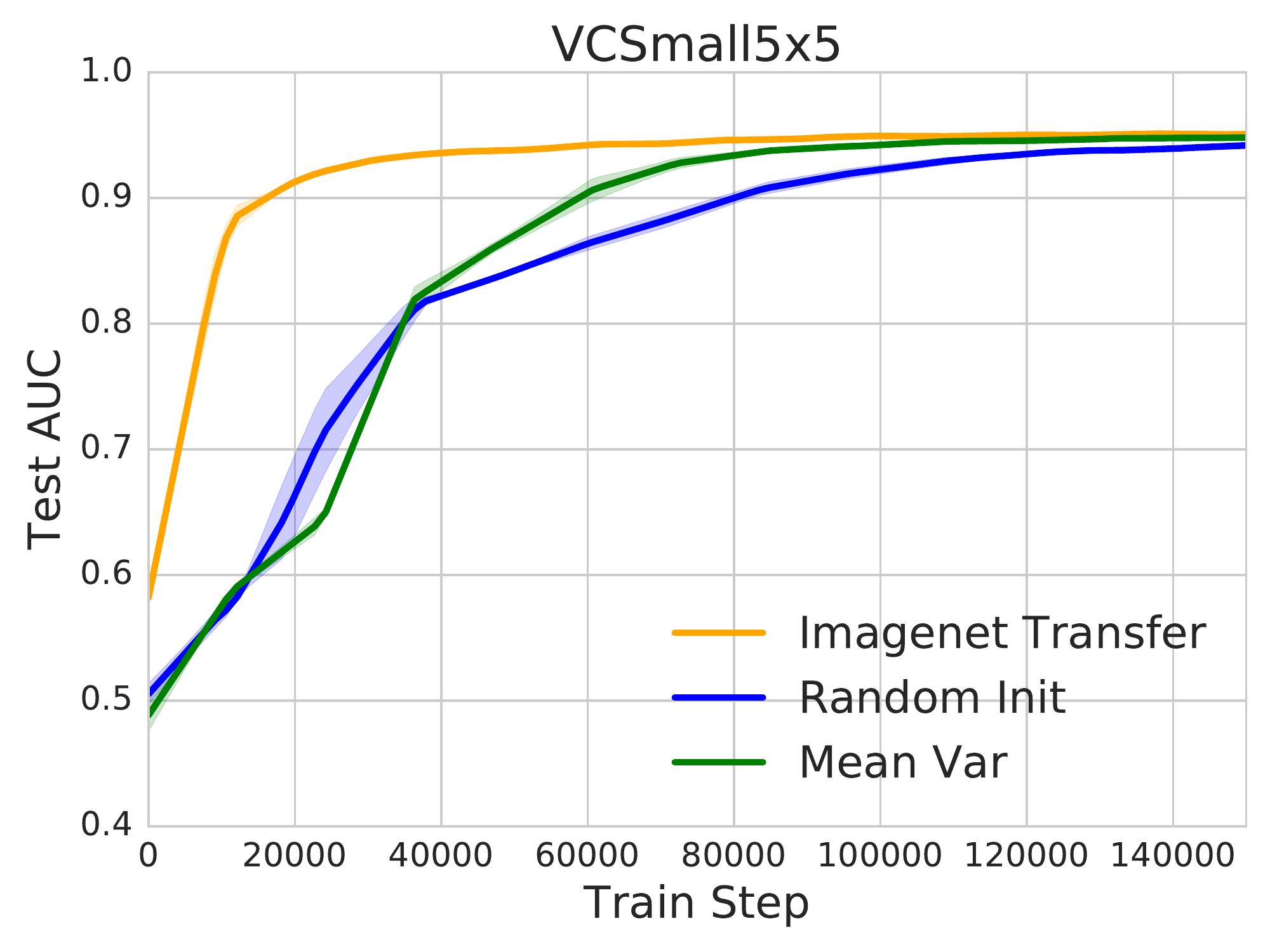} \\
\end{tabular}
\caption{\textbf{Comparison of convergence speed for different initialization schemes on \retina with various model architectures.} The three plots present the results for CBR-LargeW, CBR-Small and CBR-Tiny, respectively.}
\label{fig:app-fig-all-convergences}
\end{figure}

Figure~\ref{fig:app-weight-dist} visualizes the first layer filters from various initialization schemes. As shown in
the main text, the \emph{Mean Var} initialization could converge much faster than the baseline random initialization due to better parameter scaling transferred from the pre-trained weights. Figure~\ref{fig:app-fig-all-convergences} shows more results on \retina with various architectures. We find that on smaller models, the effectiveness of the \emph{Mean Var} initialization is less very pronounced, likely due to them being much shallower.

\begin{figure}
    \centering
    \begin{subfigure}{.32\linewidth}
    \includegraphics[width=\linewidth]{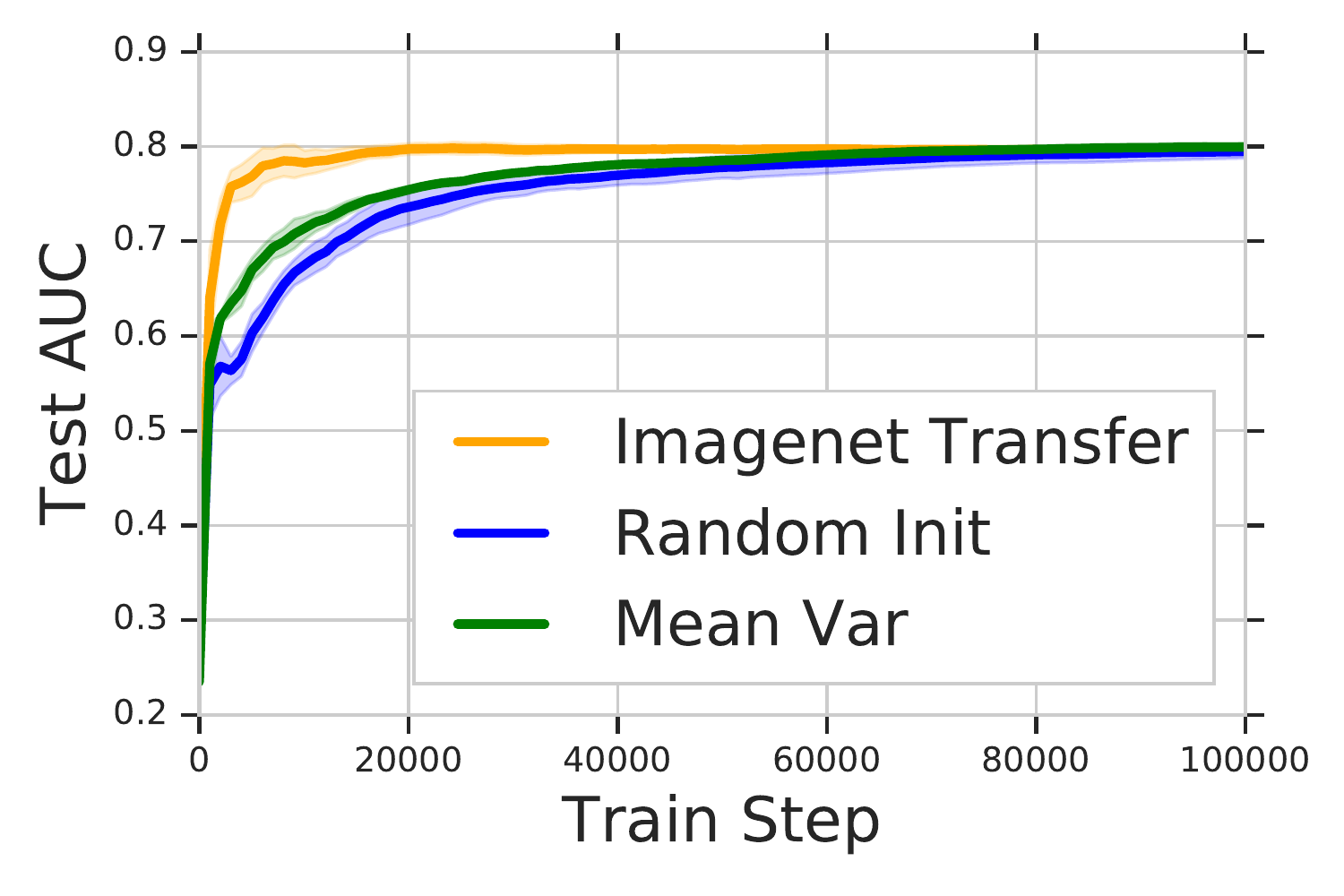}
    \caption{Atelectasis}
    \end{subfigure}
    \begin{subfigure}{.32\linewidth}
    \includegraphics[width=\linewidth]{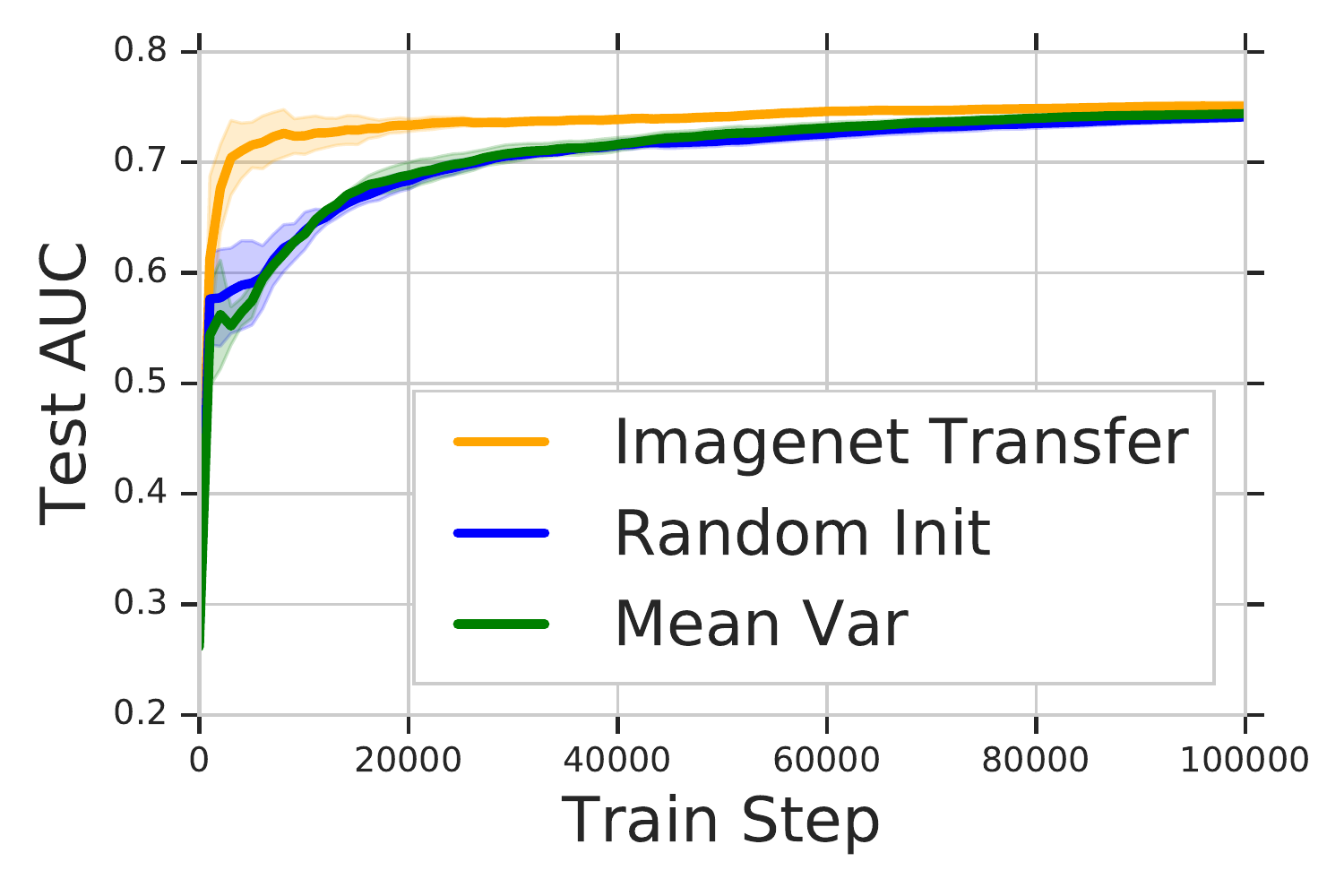}
    \caption{Cardiomegaly}
    \end{subfigure}
    \begin{subfigure}{.32\linewidth}
    \includegraphics[width=\linewidth]{convergence/chexpert/Consolidation}
    \caption{Consolidation}
    \end{subfigure}
    \begin{subfigure}{.49\linewidth}
    \includegraphics[width=\linewidth]{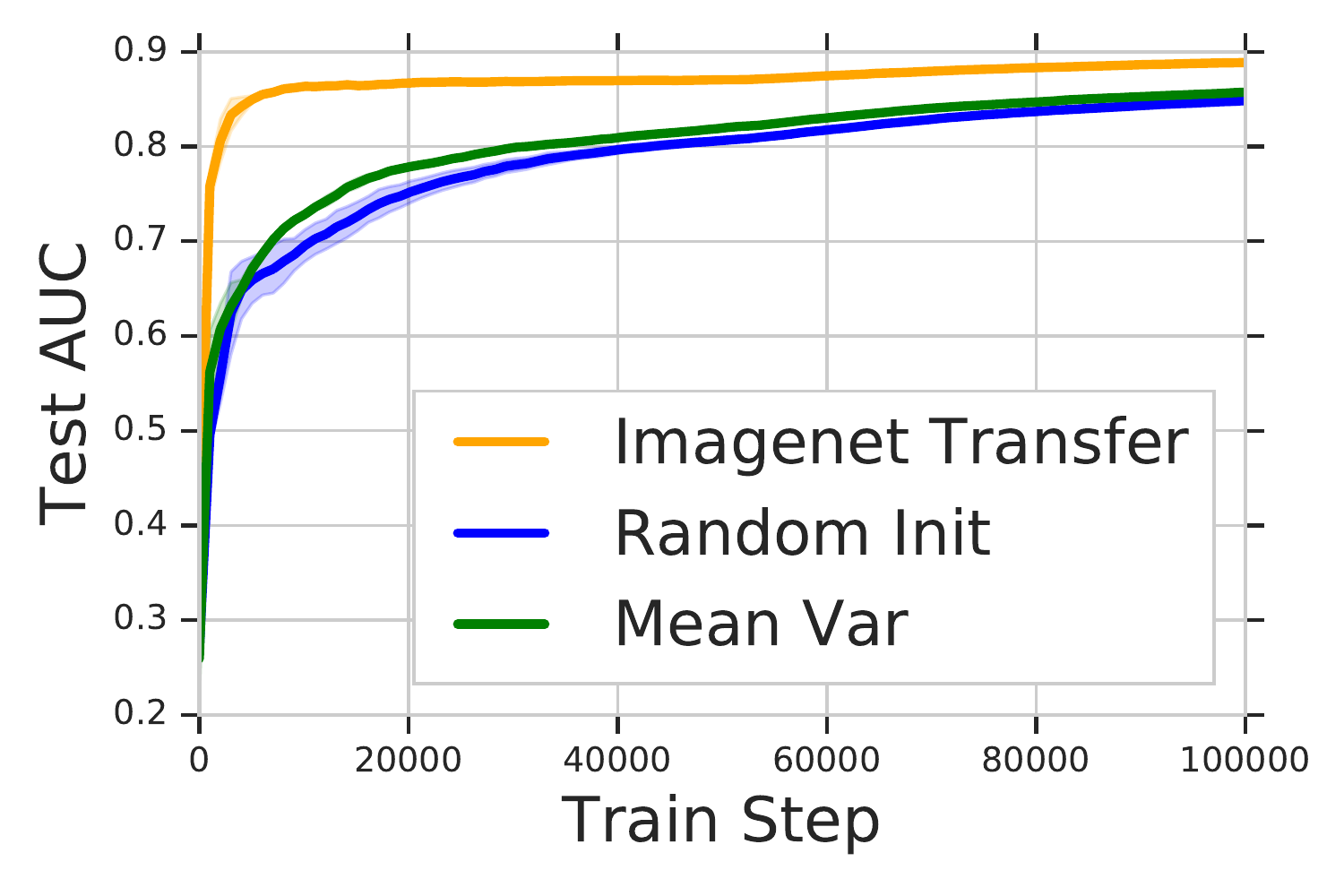}
    \caption{Edema}
    \end{subfigure}
    \begin{subfigure}{.49\linewidth}
    \includegraphics[width=\linewidth]{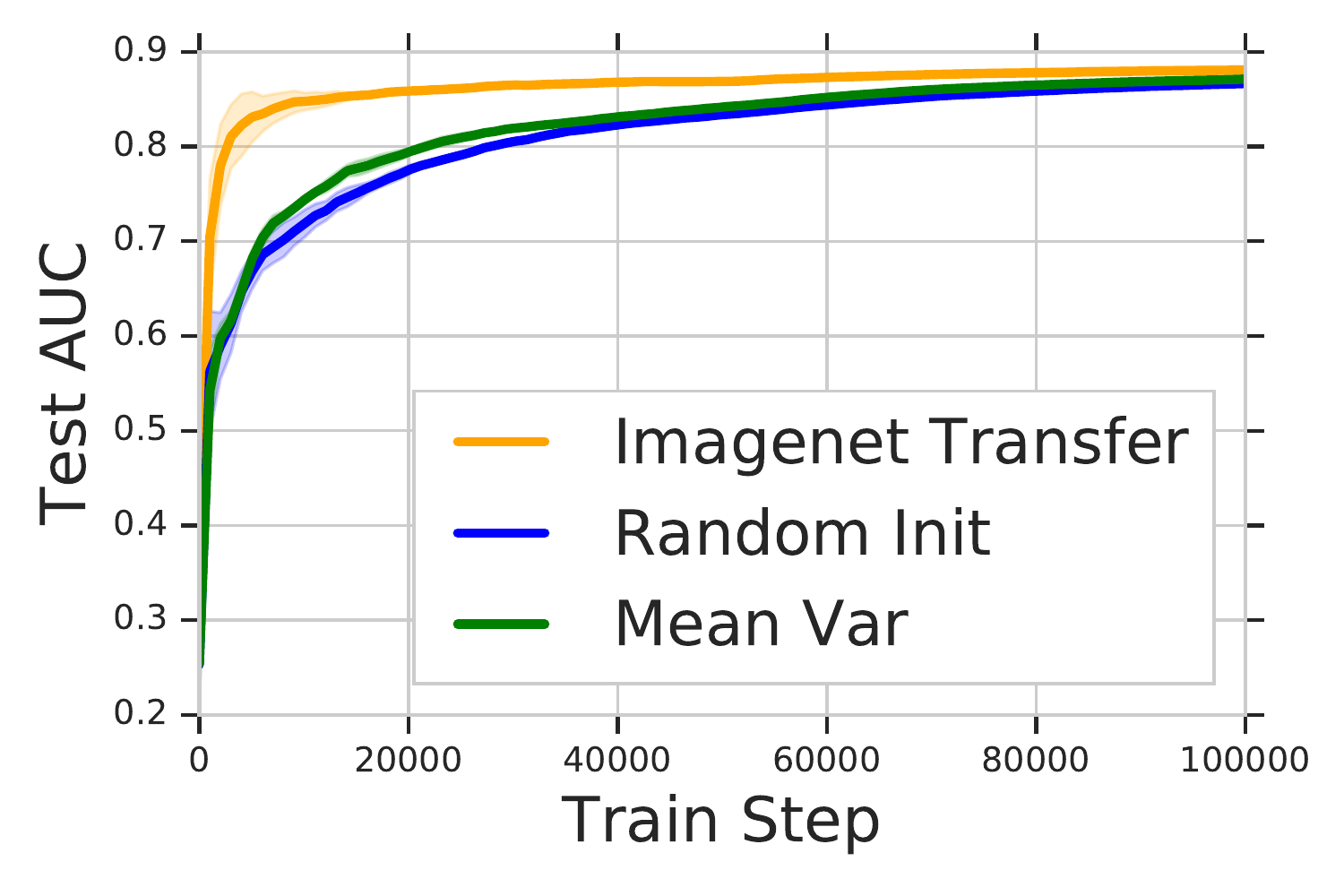}
    \caption{Pleural Effusion}
    \end{subfigure}
    \caption{\textbf{Comparison of convergence speed for different initialization schemes on the \chexpert data with Resnet-50.}}
    \label{fig:app-chexpert-meanvar}
\end{figure}

Figure~\ref{fig:app-chexpert-meanvar} shows all the five diseases on the \chexpert data for Resnet-50. Except for Cardiomegaly, we see benefits of the \emph{Mean Var} initialization scheme on convergence speed in all other diseases.

\subsection{Batch Normalization Layers}

\begin{SCfigure}\centering
\includegraphics[width=.6\linewidth]{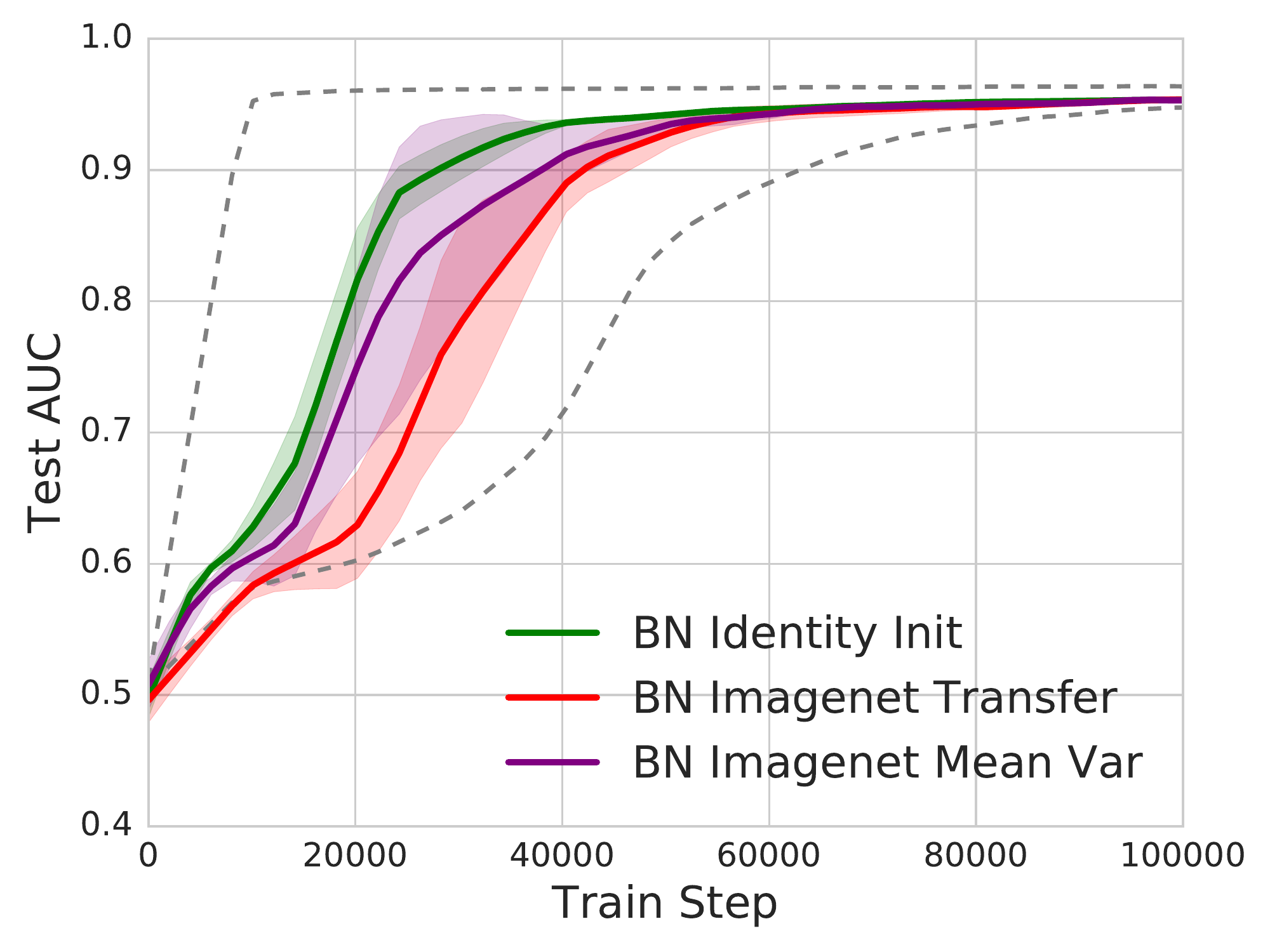}
\caption{\small\textbf{Comparing different ways of importing the weights and statistics for batch normalization layers.} The rest of the layers are initialized according to the Mean Var scheme. The two dashed lines show the convergence of the \imagenet init and the Random init for references. The lines are averaged over 5 runs.}
\label{fig:bn-zoomed}
\end{SCfigure}

Batch normalization layers \citet{ioffe2015batch} are an essential building block for most modern network architectures with visual inputs. However, these layers have a slightly different structure that requires more careful consideration when performing the Mean Var init. Letting $x$ be a batch of activations, batch norm computes
\[ \gamma \left( \frac{(x - \mu_B)}{\sigma_B + \epsilon} \right) + \beta \]
Here, $\gamma, \beta$ are learnable scale, shift parameters, and $\mu_B, \sigma_B$ are an accumulated running mean and variance over the train dataset. Thus, in transfer learning, $\mu_B, \sigma_B$ start off as the mean/variance of the \imagenet data activations, unlikely to match the medical image statistics. Therefore, for the Mean Var Init, we initialized all of the batch norm parameters to the identity: $\gamma, \sigma_B = 1$, $\beta, \mu_B$ = 0. We call this the \textit{BN Identity Init}. Two alternatives are \textit{BN \imagenet Mean Var}, resampling the values of all batch norm parameters according to the \imagenet means and variances, and \textit{BN \imagenet Transfer}, copying over the batch norm parameters from \imagenet. We compare these three methods in Figure \ref{fig:bn-zoomed}, with non-batchnorm layers initialized according to the Mean Var Init. Broadly, they perform similarly, with \textit{BN Identity Init} (used by default in other Mean Var related experiments) performing slightly better. We observe that \textit{BN \imagenet Transfer}, where the \imagenet batchnorm parameters are transferred directly to the medical images, performs the worst.

\subsection{Mean Var Init vs Using Knowledge of the Full Empirical \imagenet Weight Distribution}
\label{sec:full-empirical}
In Figure \ref{fig:app-weight-dist}, we see that while the Mean Var Init might have the same mean and variance as the \imagenet weight distribution, the two distributions themselves are quite different from each other. We examined the convergence speed of initializing with the Mean Var Init vs initializing using knowledge of the entire empirical distribution of the \imagenet weights.

\begin{SCfigure}\centering
\includegraphics[width=.6\linewidth]{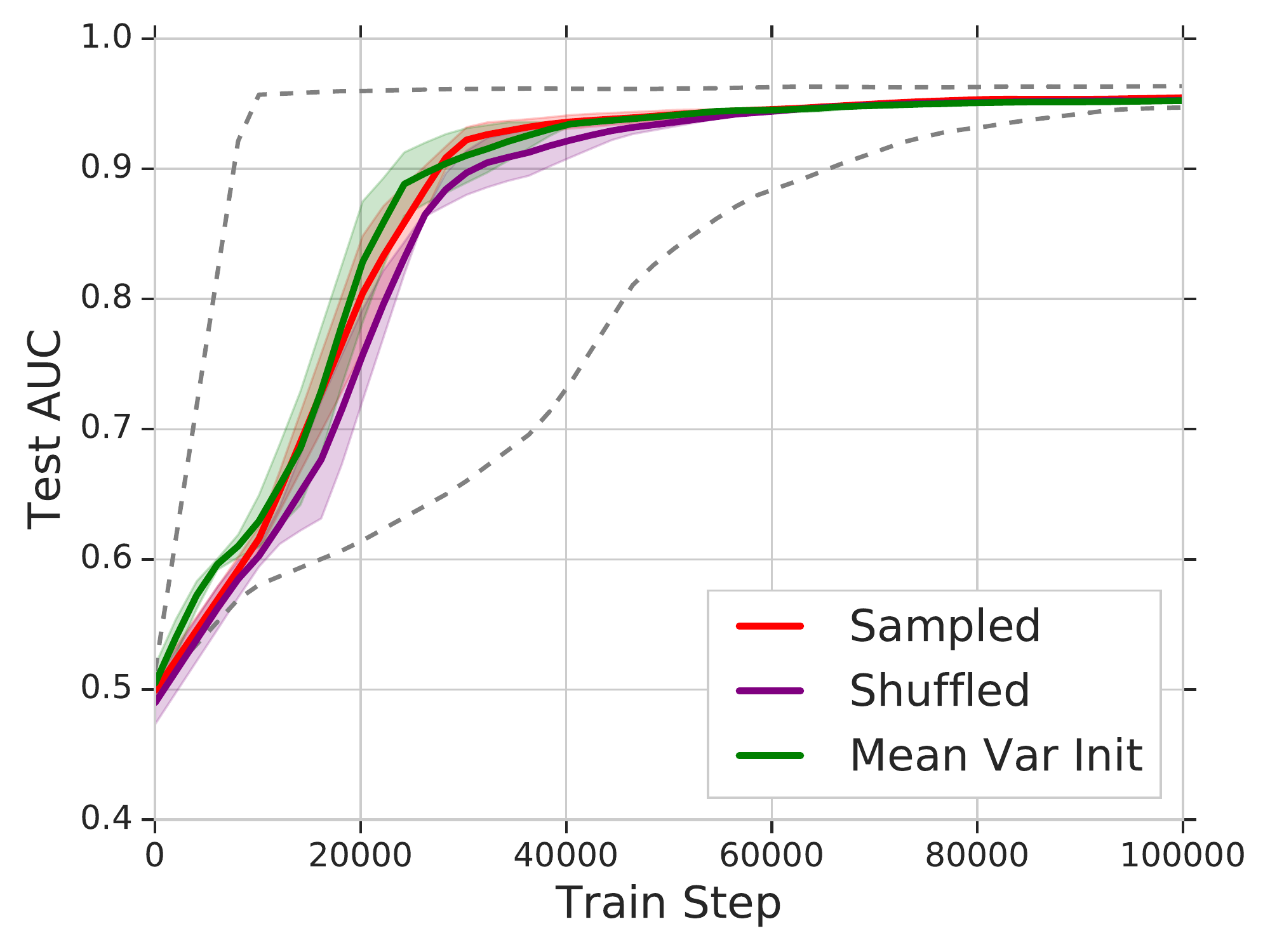}
\caption{\small\textbf{The Mean Var Init converges with a similar speed to using the full empirical distribution of the pretrained \imagenet weights.} The plots show the convergence speed of initializing by sampling from the empirical \imagenet weight distribution, and initializing by randomly shuffling the pretrained weights (i.e. sampling without replacement). We see that Mean Var converges at a similar speed to using the full empirical distribution. All lines are averaged over 3 runs, and the dashed lines show the convergence of the \imagenet init and the Random init as a reference.}
\label{fig:empirical-meanvar}
\end{SCfigure}

In particular, we looked at (1) \textit{Sampling Init:} each weight is drawn iid from the full empirical distribution of \imagenet weights (2) \textit{Shuffled Init:} random shuffle of the pretrained \imagenet weights to form a new initialization. (Note this is exactly sampling from the empirical distribution without replacement.) The results are illustrated in Figure \ref{fig:empirical-meanvar}. Interestingly, Mean Var is very similar in convergence speed to both of these alternatives. This would suggest that further improvements in convergence speed might have to come from also modelling correlations between weights.

\begin{figure}
\centering
\begin{tabular}{cc}
\hspace*{-10mm} \includegraphics[width=0.5\columnwidth]{./cameraready_transfusion/cameraready_resnet_transfusion.pdf} &
 \hspace*{-5mm} \includegraphics[width=0.5\columnwidth]{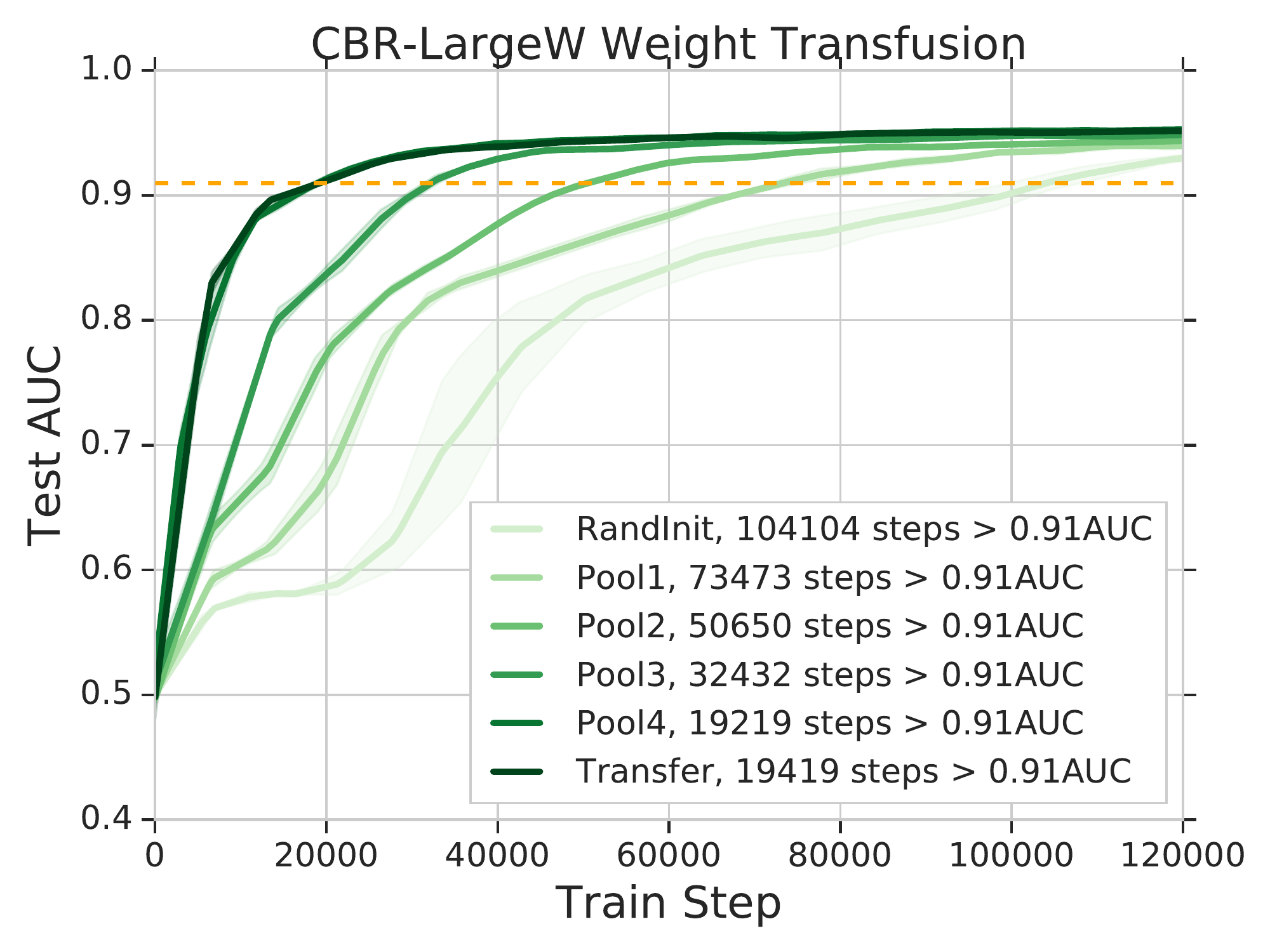}
 \\
 \hspace*{-10mm} \includegraphics[width=0.47\columnwidth]{./cameraready_transfusion/cameraready_resnet_transfusion_scatter.pdf} &
\hspace*{-5mm} \includegraphics[width=0.47\columnwidth]{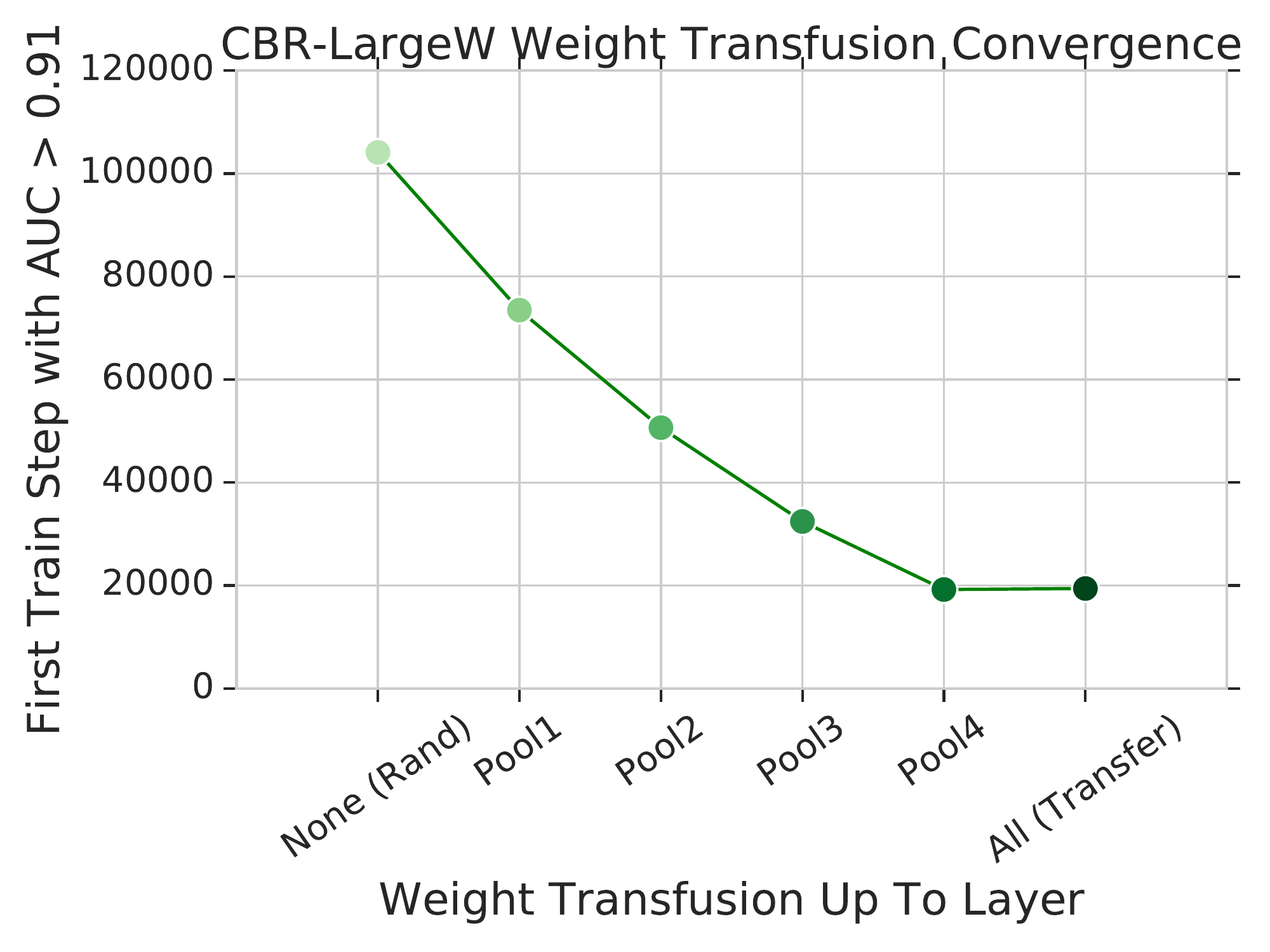}
\end{tabular}
\caption{\textbf{Weight transfusion results on Resnet50 (from main text) and CBR-LargeW}. These broadly show the same results --- reusing pretrained weights for lowest layers give significantly larger speedups. Because CBR-LargeW is a much smaller model, there is slightly more change when reusing pretrained weights in high layers, but we still see the same diminishing returns pattern.}
\label{fig:transfusion-new-app}
\end{figure}

\begin{figure}
\centering
\begin{tabular}{c}
\hspace*{-10mm} \includegraphics[width=0.8\columnwidth]{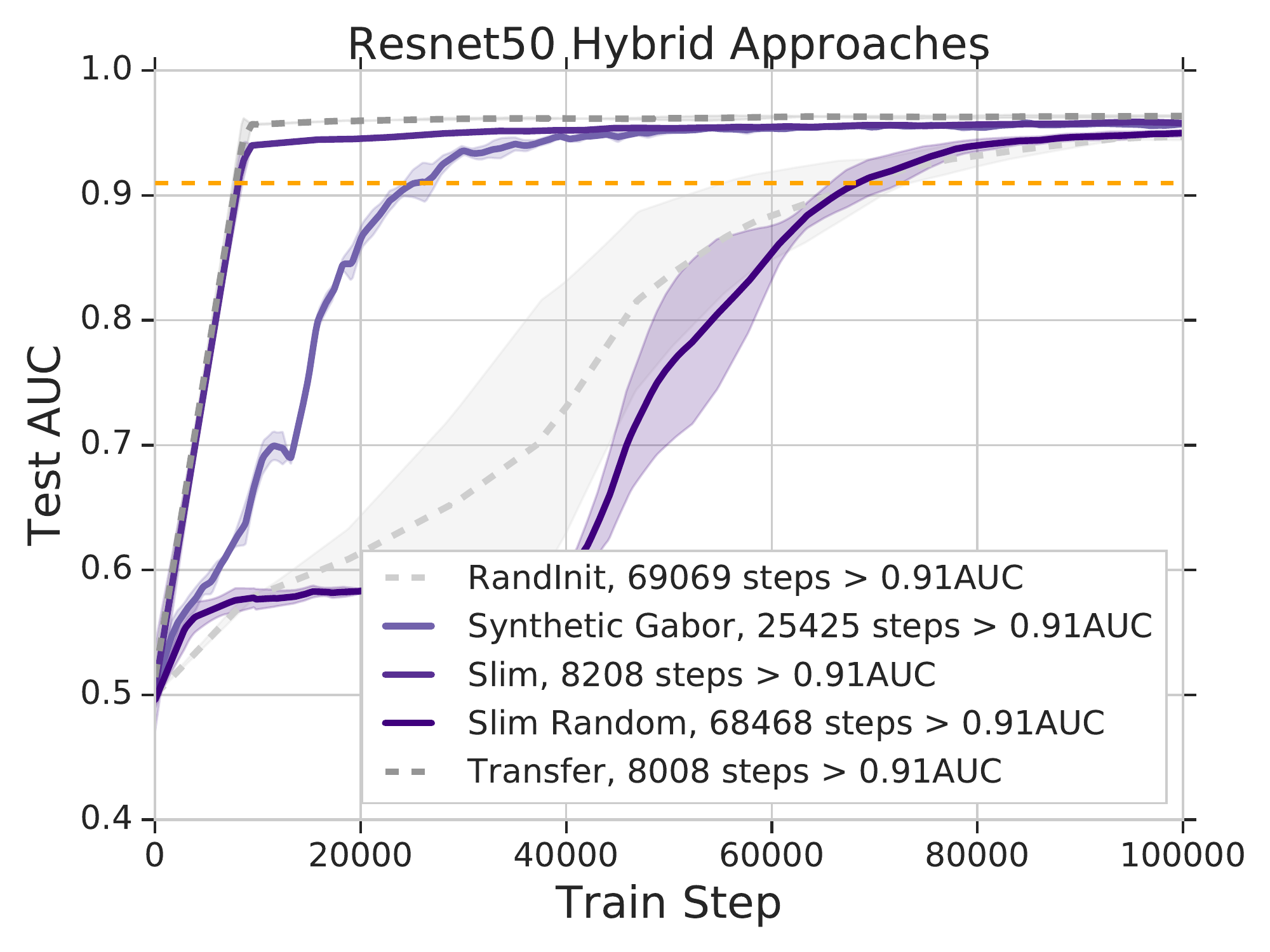}
\end{tabular}
\caption{\textbf{Convergence of Slim Resnet50 from random initialization}. We include the convergence of the slim Resnet50 --- where layers in Block3, Block4 have half the number of channels, and when we don't use any pretrained weights. We see that it is significantly slower than the hybrid approach in the main text.}
\label{fig:hybrid-app}
\end{figure}

\subsection{Synthetic Gabor Filters}
\label{sec:app-syn-gabor}

\begin{figure}
    \centering
    \includegraphics[width=.6\linewidth]{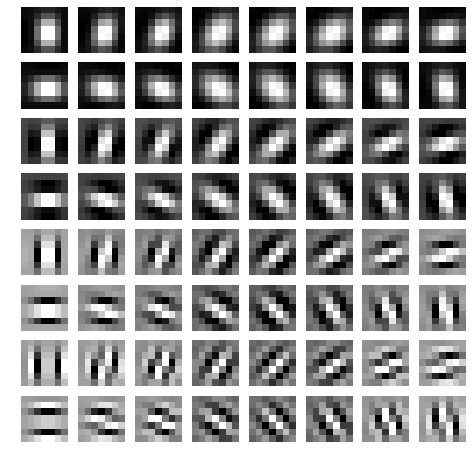}
    \caption{Synthetic Gabor filters used to initialize the first layer if neural networks in some of the experiments in this paper. The Gabor filters are generated as grayscale images and repeated across the RGB channels.}
    \label{fig:synthetic-gabors}
\end{figure}

We test mathematically synthetic Gabor filters in place of learned Gabor filters on \imagenet{} for its benefits in speeding up the convergence when used as initialization in the first layer of neural networks. The Gabor filters are generated with the \texttt{skimage} package, using the following code snippet.

\begin{verbatim}
from skimage.filters import gabor_kernel
from skimage.transform import resize
import numpy as np

def gen_gabors(n_angles=16, sigmas=[2], freqs = [0.08, 0.16, 0.25, 0.32],
               kernel_resize = 10, kernel_crop = 7):
  kernels = []
  for sigma in sigmas:
    for frequency in freqs:
      for theta in range(n_angles):
        theta = theta / n_angles * np.pi
        kernel = np.real(gabor_kernel(frequency, theta=theta,
                                      sigma_x=sigma, sigma_y=sigma))
        kernel_size = kernel.shape[0]
        if kernel_size > kernel_resize:
          kernel = resize(kernel, (kernel_resize, kernel_resize))
          kernel_size = kernel.shape[0]
        else:
          assert kernel_size >= kernel_crop
        # center crop
        size_delta = kernel_size - kernel_crop
        kernel = kernel[size_delta//2:-(size_delta-size_delta//2),
                        size_delta//2:-(size_delta-size_delta//2)]

        kernels.append(kernel)
  return kernels
\end{verbatim}

Figure~\ref{fig:synthetic-gabors} visualize the synthetic Gabor filters. To ensure fair comparison, the synthesized Gabor filters are scaled (globally across all the filters with a single scale parameter) to match the numerical magnitudes of the learned Gabor filters.

\end{document}